%% file: main.tex
\documentclass[twocolumn, 10pt]{article}
\usepackage[a4paper, margin=1.4cm]{geometry}
\input{prefix}

\title{Digitizing Paper ECGs at Scale: An Open-Source Algorithm for Clinical Research}

\author{
Elias Stenhede\textsuperscript{1,2}, 
Agnar Martin Bjørnstad\textsuperscript{1,2}, 
Arian Ranjbar\textsuperscript{1} \\\\[1ex]
\textsuperscript{1}Medical Technology \& E-health, Akershus University Hospital, 1478 Lørenskog, Norway\\
\textsuperscript{2}Faculty of Medicine, University of Oslo, 0317 Oslo, Norway
}
\date{}

\begin{document}

\maketitle
\textbf{
Millions of clinical ECGs exist only as paper scans, making them unusable for modern automated diagnostics. We introduce a fully automated, modular framework that converts scanned or photographed ECGs into digital signals, suitable for both clinical and research applications. The framework is validated on 37,191 ECG images with 1,596 collected at Akershus University Hospital, where the algorithm obtains a mean signal-to-noise ratio of 19.65\,dB on scanned papers with common artifacts. It is further evaluated on the Emory Paper Digitization ECG Dataset, comprising 35,595 images, including images with perspective distortion, wrinkles, and stains. The model improves on the state-of-the-art in all subcategories. The full software is released as open-source, promoting reproducibility and further development. We hope the software will contribute to unlocking retrospective ECG archives and democratize access to AI-driven diagnostics.}
\hspace{0.5cm}

Cardiovascular disease remains the leading cause of death globally, underscoring the importance of effective diagnostic tools~\cite{WHO_CVD_2022}. Among these, the ECG is the most widely used, primarily due to its ease of use, non-invasiveness, low operator variability, and cost-effectiveness. Traditionally, ECG recordings have been acquired using analog instruments that produce printed signals on paper strips, subsequently analyzed by clinicians and archived in patient medical records. Despite technological advances, this practice persists today due to the familiarity for clinicians, convenience, and ease of interpretation. With the widespread adoption of electronic health records, these paper ECGs are now often scanned to be stored as images within the patient database.

Advancements in AI, particularly with the advent of deep learning during the last decade, have significantly improved computer-assisted analysis of ECGs~\cite{liu_deep_2021, wu2025deep}, even surpassing human diagnostic performance~\cite{jiang_deep_2024, zhu_automatic_2020}. State-of-the-art algorithms primarily utilize ECG data in standardized time series formats to perform both diagnostic classification \cite{ribeiro2020automatic,al-zaiti_machine_2023, kwon_deep_2020}, and risk stratification~\cite{ouyang_electrocardiographic_2024}. Furthermore, beyond typical cardiovascular assessment, similar ECG-based AI models have shown potential in e.g., identifying demographic characteristics \cite{attia2019age}.

While the healthcare sector is gradually adopting digital storage of original ECG signals, many hospitals and healthcare providers continue to rely on scanned paper records. Although AI models can analyze scanned ECG images, their performance and reliability significantly improve when standardized digital time-series data are available, mitigating complications arising from variations in channel configurations and scanning artifacts. Developing a robust and efficient tool to digitize paper ECG records accurately offers two critical advantages. First, it democratizes access to AI-driven ECG interpretation tools, making advanced diagnostic capabilities available regardless of digital infrastructure. Second, digitizing historical ECG archives substantially expands the volume of available data for training AI models, potentially by several orders of magnitude. This enhancement is particularly valuable in studying rare cardiovascular conditions, where extensive historical datasets can significantly increase sample sizes.
\begin{figure}[t]
    \centering
    \resizebox{\columnwidth}{!}{\input{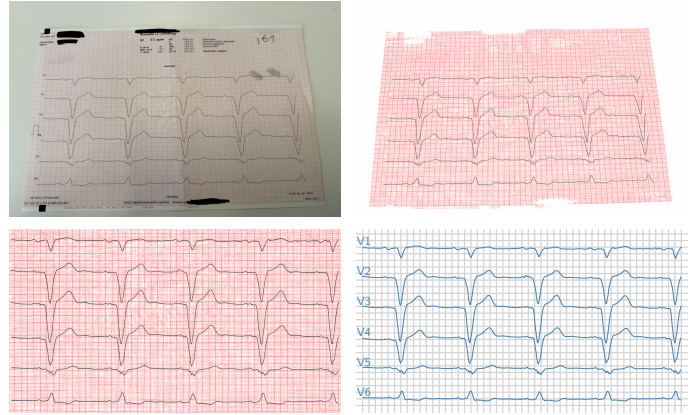}}
    \caption{Shows the digitization pipeline for a paper ECG captured using a mobile phone. \textbf{Top left:} Mobile phone photo of a printed ECG. \textbf{Top right:} Segmented image showing detected ECG waveform and grid structure. \textbf{Bottom left:} Perspective-corrected and cropped ECG aligned. \textbf{Bottom right:} Final extracted waveforms in standardized format for archiving and downstream processing.}
    \label{fig:visual-abstract}
\end{figure}

\subsection*{Related work}

Over the past two decades, several semi-automatic methods have been proposed to address the challenge of ECG digitization, including Badilini et al., Ravichandran et al., and Baydoun et al., who in the period between 2005 and 2019 introduced MATLAB-based solutions~\cite{badilini_ecgscan_2005,ravichandran_novel_2013,noauthor_high_2019}, all requiring some degree of user input. In 2021, Mishra et al. proposed an algorithm~\cite{mishra_ecg_2021} for crops of single leads. By 2021, Fortune et al. provided an openly available Python implementation of their semi-automatic method~\cite{fortune_digitizing_2021}.

In the last few years, efforts have been directed towards fully automated methods. In 2022, Wu et al. proposed a method including an accessible API for digitization tasks~\cite{wu_fully-automated_2022}. We were, however, not successful in digitizing any images from our database using the API. To accelerate the development of a general automated ECG digitization algorithm, the annual George B. Moody PhysioNet Challenge announced that the 2024 competition would be dedicated to the topic. This resulted in 17 distinct methods from the competing research groups, which were evaluated on a hidden test dataset consisting of scans and photographs of ECGs in varying conditions, including clean, strained, and significantly deteriorated samples. Performance was assessed quantitatively with the ground truth reference~\cite{a_reyna_digitization_2024}. Among the methods, three demonstrated superior performance compared to a naive baseline of assigning zero values to all signal points~\cite{h_krones_combining_2024, yoon_segmentation-based_2024, stenhede_ecg-dual_2024}. Nevertheless, despite surpassing the baseline, none showed robust results on photos, and the source code is not available at the time of writing. In 2025, Demolder et al. introduced an ECG digitization method and released an evaluation dataset comprising 6,000 ECG images~\cite{demolder_high_2025} derived from 100 ECGs. A commendable aspect of their work is the physical realism introduced by printing and photographing ECGs, mimicking real-world digitization scenarios. It does not, however, capture common degradations of handwritten notes by clinicians or variations introduced by printing on thermal paper, and the source code is not shared.

In summary, previous methods have been constrained by one or more of: (1) lack of automation or accuracy, (2) insufficient validation across realistic datasets, and (3) the absence of open-source code. To address these issues, we propose a new modular open-source framework for a general-purpose ECG digitization algorithm, capable of digitizing both scans and photos, validated on both publicly available datasets and real-world data. In short, our contributions include:
\begin{itemize}
    \item A fully automated modular framework for general-purpose ECG digitization consisting of five modules: Perspective correction and cropping, Segmentation, Layout identification, Grid size extraction/scaling, and 2D to 1D conversion; publicly released as open source.
    \item Methodological and algorithmic advances, including dewarping, alignment, and signal extraction methods. We also release a training dataset of 8,658 simulated high-resolution ECG images with labels for segmentation and ground-truth signals, enabling reproducibility and further development.
    \item A real-world clinical dataset collected at Akershus University Hospital, containing 1,596 samples of photographed and scanned paper ECGs, provided together with the raw time series data.
    \item Extensive validation of the proposed framework on real-world datasets.
\end{itemize}

\begin{figure*}[ht]
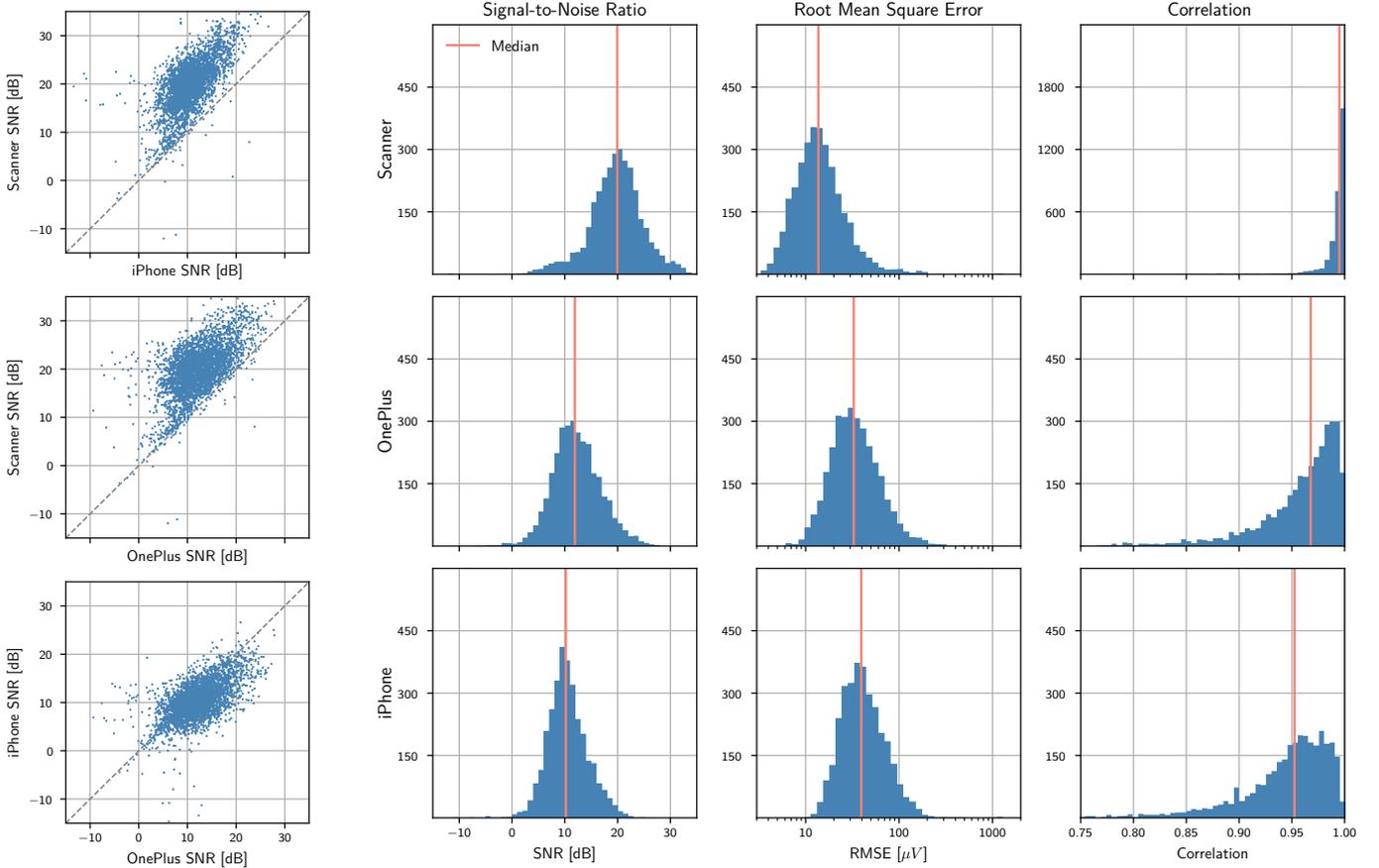

    \centering
    \begin{subfigure}[b]{0.245\textwidth}
        \centering
        \resizebox{\linewidth}{!}{\input{figures/results/pairwise_snr_scatter_plots.pgf}}
        \caption{Pairwise scatter plot of SNR per lead, for images captured with the three different devices.}
        \label{fig:pairwise-snr-scatter}
    \end{subfigure}
    \hfill
    \begin{subfigure}[b]{0.745\textwidth}
        \centering
        \resizebox{\linewidth}{!}{\input{figures/results/ecg_digitization_metrics_per_lead.pgf}}
        \caption{Distribution of signal quality metrics for ECG recordings captured using three devices (Scanner, iPhone, OnePlus). Using the scanner results in better average results across all three metrics. The two phones perform similarly, with OnePlus resulting in slightly better results.}
        \label{fig:digitization-histogram-per-lead}
    \end{subfigure}
    \caption{Digitization metrics across devices.}
    \label{fig:per-lead-results}
\end{figure*}

\section*{Results}
The full ECG digitization framework is released with all code, the synthetic training dataset, and the clinical dataset from the prospective validation, as open source. In this section*, we present the clinical data and results from the validation.

\subsection*{Validation data}
The proposed algorithm is evaluated using two main datasets, the Ahus Paper Digitization ECG Database and the Emory Paper Digitization ECG Database. The first contains paper ECGs from Akershus University Hospital in Norway, reflecting data quality in a real-world clinical setting. The second dataset was used as the hidden test set in the 2024 George Moody PhysioNet Challenge. At the time of writing, the full Emory dataset is hidden, representing a zero-shot evaluation. Following the publication of the current article, both the Ahus and Emory Paper ECG Databases will be jointly released as a unified dataset. Both datasets are described in further detail in \cite{reyna2024ecgimagedatabasedatasetecgimages}.



\subsection*{Validation metrics}
We use signal-to-noise ratio (SNR) as the main metric, and additionally report correlation and root mean squared error (RMSE). Metrics are calculated per lead and analyzed separately for the scanner and mobile phone cameras. The ground truth signals are sampled at \SI{1000}{\hertz}, and all reconstructed signals are resampled to match this rate. Denoting $y$ as the ground truth signal and $\hat y$ as the reconstruction, SNR is calculated as 
\begin{equation*}
    \mathrm{SNR} = 10 \log_{10}  \left(\frac{\sum_t y[t]^2}{\sum_t (y[t]-\hat y[t])^2}\right),
\end{equation*}
where the numerator is the power of the true signal and the denominator is the power of the noise signal. In evaluation, the ground truth and digitized signals are aligned using horizontal and vertical shifts. The horizontal alignment allows for a maximum shift of \SI{100}{\milli\second}, while the vertical alignment is performed by zero-centering both signals. Shifting is motivated by (a) there being no fixed baseline in the paper ECG and (b) translation having no clinical significance, and (c) previous work uses shifted metrics~\cite{h_krones_combining_2024, yoon_segmentation-based_2024, stenhede_ecg-dual_2024, demolder_high_2025}.

\subsection*{Ahus Paper Digitization ECG Database}
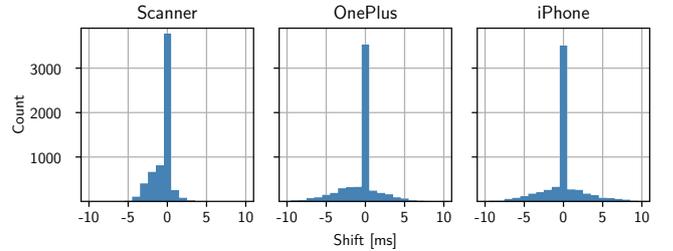
\begin{figure}[ht]
    \centering
    \resizebox{\columnwidth}{!}{\input{figures/results/shift_histograms.pgf}}
    \caption{Shows the necessary shift to align each lead with the ground truth. In general, smaller shifts are needed for images captured with a flatbed scanner, and for all three devices, no shift is the most common.}
    \label{fig:shift}
\end{figure}

Paper ECGs were collected from the Department of Cardiology at Akershus University Hospital over 3 months in
2025. ECGs were printed on thermal paper with 1 mm grid lines, 50\,mm/s, 10\,mm/mV, then photographed with two mobile phone camera setups (iPhone: $5712\times4284$ pixels, OnePlus: f/1.78 and $4624\times3468$ pixels, f/1.7) and scanned into flatbed scans ($7016\times4964$ pixels, 600 dpi). Each scan was then paired with the raw time series data extracted from the hospital's databases~\cite{ranjbar2023data}. As each 12-lead is printed on two papers, every ECG results in 6 images. The resulting dataset comprises 1,596 images derived from 266 unique ECGs and 266 unique patients, with 99 identified as female, 153 as male, and 14 with unknown gender due to no personal information being entered. Each image was de-identified and stored together with the raw time series in a DICOM file. The images and 266 corresponding de-identified DICOM files were saved for downstream processing and analysis~\cite{ranjbar2023enabling}.
Using the hardware described in \Cref{tab:env}, processing one image took on average \SI{5.88}{\second}, with a minimum processing time of \SI{5.43}{\second} and a maximum of \SI{7.45}{\second}, across the 1,596 images in the dataset. Using the same setup but on CPU only, processing one image took on average \SI{16.78}{\second} with the minimum and maximum values being \SI{15.53}{\second} and \SI{23.39}{\second}. On average, the signals were shifted by \SI{0.86}{\milli\second} (s.d. \SI{1.66}{\milli\second}) for scanner images, \SI{1.89}{\milli\second} (s.d. \SI{4.31}{\milli\second}) for iPhone images, and \SI{1.96}{\milli\second} (s.d. \SI{5.23}{\milli\second}) for OnePlus images. A histogram of time shifts is shown in \Cref{fig:shift}. The algorithm may assign an error code to individual elements (sets the value to NaN) in the reconstruction, for example, if a lead is not visible or if the algorithm partly fails.
Across all three devices, an average of \SI{0.21}{\percent} of samples were set to NaN. Specifically, \SI{0.15}{\percent} of samples were set to zero for the scanned paper, with corresponding numbers for OnePlus and iPhone being \SI{0.23}{\percent} and \SI{0.25}{\percent}.

We observed a mean SNR across all 12 leads of \SI{19.65}{\decibel} for scanned images, compared to \SI{12.19}{\decibel} for OnePlus photos and \SI{10.47}{\decibel} for iPhone photos. Corresponding RMSE and correlation values also followed this ranking, see \Cref{tab:metrics}. The same results are also displayed in \Cref{fig:digitization-histogram-per-lead}, allowing for assessment of the spread, and pairwise SNR comparisons across all leads and devices are plotted in \Cref{fig:pairwise-snr-scatter}.  

\begin{table}[ht]
    \centering
    \caption{Mean and standard deviation (s.d.) of evaluation metrics across the three devices tested.}
    \begin{tabular}{lcccccc}
    \toprule
     & \multicolumn{2}{c}{SNR [\SI{}{\decibel}]} 
     & \multicolumn{2}{c}{RMSE [$\mu$V]} 
     & \multicolumn{2}{c}{Correlation [-]} \\
    \cmidrule(lr){2-3}\cmidrule(lr){4-5}\cmidrule(lr){6-7}
     & Mean & s.d. & Mean & s.d. & Mean & s.d. \\
    \midrule
    Scanner & 19.65 & 5.20 & 19.3 & 35.4 & 0.987 & 0.038 \\
    OnePlus & 12.19 & 4.65 & 42.8 & 44.8 & 0.947 & 0.083 \\
    iPhone & 10.47 & 3.94 & 51.3 & 80.3 & 0.935 & 0.080 \\
    \bottomrule
    \end{tabular}
    \label{tab:metrics}
\end{table}

\begin{figure}[ht]
    \centering
    \resizebox{\columnwidth}{!}{\input{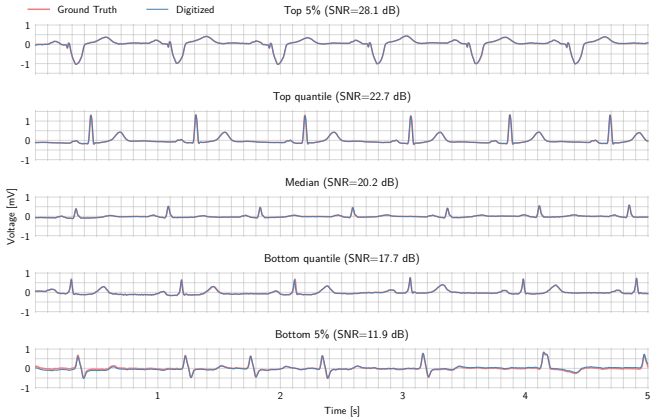}}
    \caption{Lead II as found in the original recording and after digitization of high-resolution \textbf{scanner} images using our proposed algorithm. Five different ECGs are selected based on their SNR. In the lowest-SNR examples (bottom two), a slight baseline drift appears in the scanned-digitized traces, possibly due to the error margins in detecting the rotation of the paper.}
    \label{fig:percentiles-scans}
\end{figure}

\begin{figure}[ht]
    \centering
    \resizebox{\columnwidth}{!}{\input{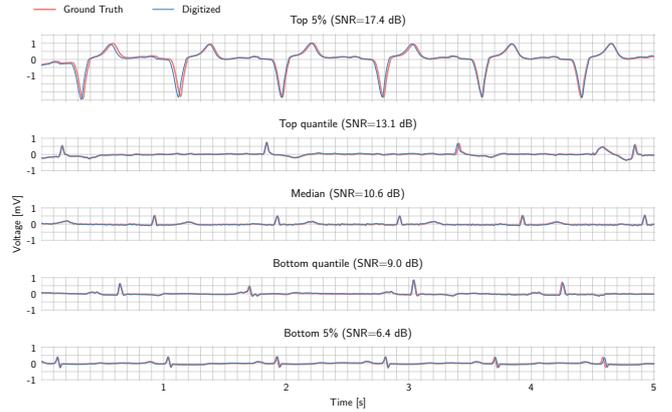}}
    \caption{Lead II as found in the original recording and after digitization of images taken with an \textbf{iPhone} Pro Max 16 using our proposed algorithm. Five different ECGs are selected based on their SNR. The signals are not always perfectly aligned in time, resulting in lower SNR as compared to scanner images.}
    \label{fig:percentiles-iphone}
\end{figure}

\begin{figure}[ht]
    \centering
    \resizebox{\columnwidth}{!}{\input{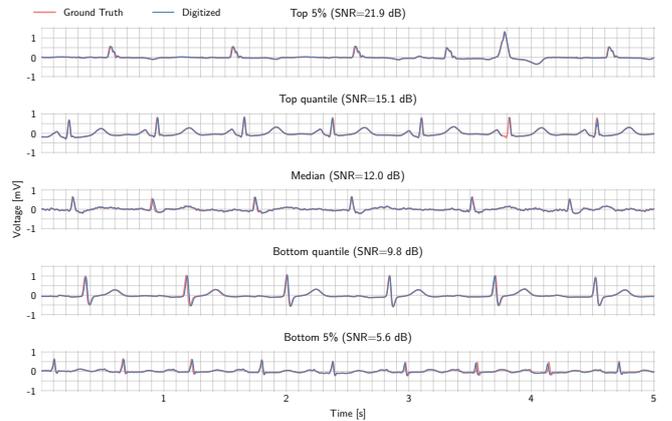}}
    \caption{Lead II as found in the original recording and after digitization of images taken with a \textbf{OnePlus} Nord CE 2 using our proposed algorithm. Five different ECGs are selected based on their SNR. The signals are not always aligned in time, resulting in lower SNR as compared to scanner images.}
    \label{fig:percentiles-oneplus}
\end{figure}

To illustrate representative reconstructions from our algorithm and highlight the limitations of SNR, we plot lead II of five signals from the scanner, iPhone, and OnePlus cameras. The signals are selected based on their respective SNR, and correspond to the 5th, 25th, 50th, 75th, and 95th percentiles. Signals reconstructed from the scanner are shown in \Cref{fig:percentiles-scans}, iPhone images in \Cref{fig:percentiles-iphone}, and OnePlus images in \Cref{fig:percentiles-oneplus}. 

\subsection*{Emory Paper Digitization ECG Database}
The dataset comprises 35,595 paper ECG images with print speed of, 25\,mm/s, and 10\,mm/mV. The images are captured under varying conditions, including color scans, black-and white scans, and mobile photos of papers that had been partially degraded, stained, or left to mold. This allows for systematic assessment of model performance across varying conditions~\cite{reyna2024ecgimagedatabasedatasetecgimages}.

As the Emory Paper Digitization ECG Database was used as the hidden test set for the 2024 George Moody PhysioNet Challenge, it is possible to directly compare our proposed algorithm with previously published methods, even as some of them have not publicly released their code. The main metric used in evaluation is shifted SNR, and results are presented in \Cref{tab:emory-results}. Our proposed algorithm outperforms previously published work across all types of images. Just as on the Ahus Paper Digitization ECG Dataset, the algorithm performs best on scans, achieving an SNR of \SI{7.34}{\decibel}. There is only a minor performance drop for black-and-white scans, with a corresponding SNR of \SI{7.24}{\decibel}. For mobile photos and scans of deteriorated papers, the algorithm achieves SNRs ranging from 1.05 to \SI{2.03}{\decibel}. Somewhat surprisingly, the algorithm performs the best on stained papers in the mobile photo category.

\begin{table*}[t]
\centering
\caption{SNR [\SI{}{\decibel}] by capture condition across the Emory Paper ECG Database. Negative scores are not shown, as an SNR of zero decibels can be achieved by setting the digitized signal to all zeros. Det. stands for deteriorated, the best values are bold, while the second best are underscored.}
\begin{tabular}{lcccccccc}
\toprule
& \multicolumn{4}{c}{Scans} & \multicolumn{3}{c}{Mobile photos} & Screenshots \\
\cmidrule(lr){2-5}\cmidrule(lr){6-8}\cmidrule(l){9-9}
& \multicolumn{2}{c}{Color} & \multicolumn{2}{c}{B\&W} &  & Color &  & Color \\
\cmidrule(lr){2-3}\cmidrule(lr){4-5}\cmidrule(lr){6-8}\cmidrule(lr){9-9}
 & Clean & Det. & Clean & Det. & Clean & Stained & Det. & Clean \\
\midrule
Open ECG Digitizer  & \textbf{7.34} & \textbf{1.82} & \textbf{7.24} & \textbf{1.73} & \textbf{1.05} & \textbf{2.03} & \textbf{ 1.59} & \textbf{2.11} \\
Yoon et al. \cite{yoon_segmentation-based_2024}       & \underline{6.22} & \underline{0.91} & 4.74 & 0.36 & -- & -- & \underline{0.81} & \underline{0.12} \\
Jammoul et al.~\cite{jammoul_layout-invariant_2024}    & 5.18 & --   & \underline{4.89} & --   & -- & -- & --   & -- \\
Krones et al. \cite{h_krones_combining_2024}  & 4.93 & 0.51 & 3.48 & \underline{0.51} & -- & -- & --   & -- \\
Stenhede et al. \cite{stenhede_ecg-dual_2024}    & 3.32 & 0.67 & 2.78 & --   & \underline{0.14} & -- & 0.79 & -- \\
\bottomrule
\end{tabular}
\label{tab:emory-results}
\end{table*}

\section*{Discussion}
The algorithm shows strong performance across all three scenarios, both quantitatively and qualitatively. The average processing time of \SI{5.88}{\second} is competitive compared to other methods, with the fastest competitor being~\cite{demolder_high_2025} reporting an average of \SI{4.86}{\second} with a minimum of \SI{3.88}{\second} and a maximum of \SI{6.33}{\second}. However, the authors do not specify the environment used in obtaining these measurements, nor share code allowing for replication. 


Our quantitative results on the Ahus Paper Digitization ECG Dataset showed an average SNR of \SI{19.65}{\decibel} on scanned images is the highest reported SNR across any previous work, even considering closed-source and data claims. The SNR is lower for mobile photos, reflecting the increased difficulty in handling paper that is often not flat, uneven lighting, and motion blur. However, we believe that the reconstruction quality is still sufficient in this case. 

Results on the Emory Paper Digitization ECG Dataset, while surpassing previously proposed methods, were comparatively lower. One likely explanation is the higher printing speed of \SI{25}{\milli\meter/\second}, which results in greater information loss during the original signal printing compared to \SI{50}{\milli\meter/\second} at Ahus. Notably, our algorithm is the first to achieve positive SNR across all categories, including mobile photos and degraded paper samples. The remaining performance gap might stem from curvature in deteriorated papers, misaligning digitized signals with their ground truth. With the public release of the Emory Paper Digitization ECG Dataset, identification of main bottlenecks will be possible, enabling further development of the methods.

\begin{figure*}[ht]
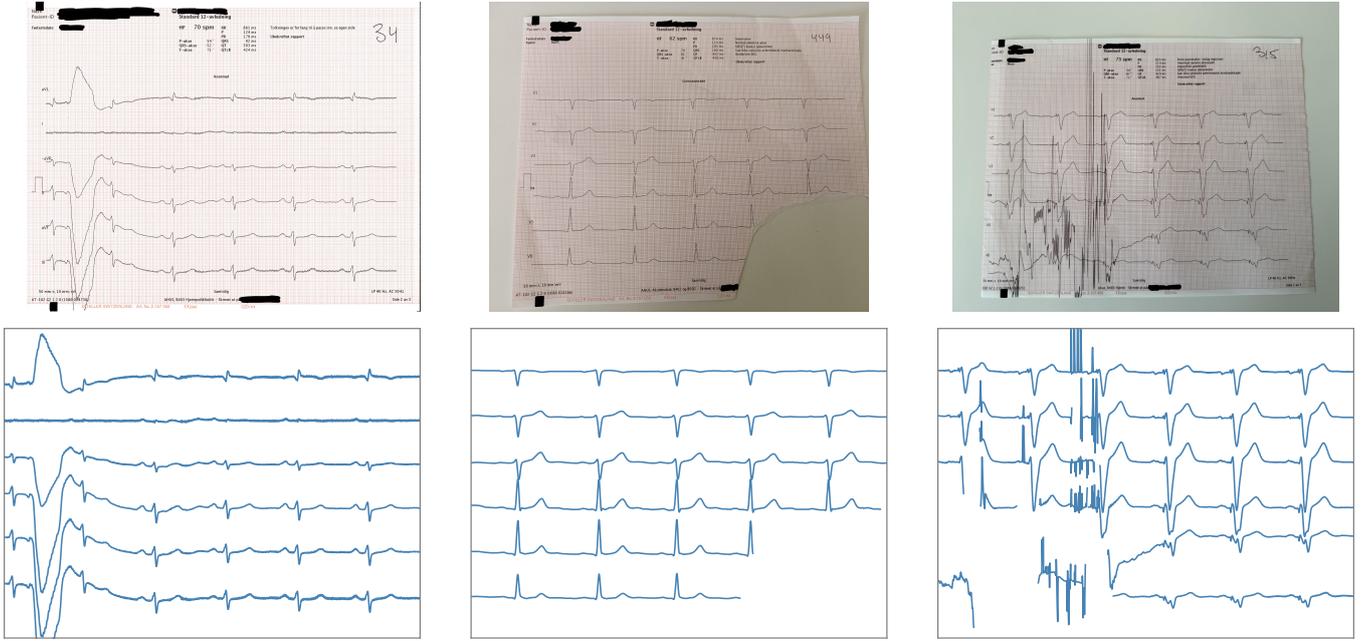

    \centering
    \begin{subfigure}[b]{0.325\textwidth}
        \centering
        \resizebox{\linewidth}{!}{\input{figures/results/example_limb_scanner_34.pgf}}
        \caption{Shows a scanner image together with its reconstruction, in which lead III has been partially set to NaN due to the signal being outside of the paper.}
        \label{fig:limb-scanner}
    \end{subfigure}
    \hfill
    \begin{subfigure}[b]{0.325\textwidth}
        \centering
        \resizebox{\linewidth}{!}{\input{figures/results/example_precordial_oneplus_449.pgf}}
        \caption{Shows a OnePlus image where the paper presumably got stuck in the ECG machine. The reconstruction puts NaNs in the part where the paper is missing.}
        \label{fig:precordial-oneplus}
    \end{subfigure}
    \hfill
    \begin{subfigure}[b]{0.325\textwidth}
        \centering
        \resizebox{\linewidth}{!}{\input{figures/results/example_precordial_iphone_315.pgf}}
        \caption{Shows an iPhone image with severe noise in lead V5, interfering with other leads. The reconstruction has several leads that have some parts with NaNs.}
        \label{fig:precordial-iphone}
    \end{subfigure}
    \caption{Three examples of signals with significant numbers of NaNs in them, across the three devices. Instead of rejecting the images, the leads are left blank, and the user can decide what to do with them.}
    \label{fig:edgecases}
\end{figure*}

The qualitative results show that in cases with lower SNR, misalignment is likely the main contributing cause of seemingly degraded performance, rather than obvious morphological inaccuracies. Unlike previously proposed ECG digitization solutions, our algorithm does not reject challenging images, but instead sets leads to NaN at points where it is not capable of digitizing. This comes with both advantages and disadvantages, which should be considered in clinical applications. One suggested approach could be to let the user manually determine what appropriate steps are for each challenging sample. Examples of cases with NaNs in the reconstructions are shown in \Cref{fig:edgecases}.

\subsection*{Limitations and future work}

All work is best understood in light of its limitations. We present an algorithm for digitizing paper ECGs at scale, suitable for use on a computer. To fully unlock the potential of this algorithm in clinical practice, its efficacy needs to be validated in further studies. Similar studies have been conducted, but not for open-source software~\cite{himmelreich_diagnostic_2023}. Another limitation is the use of standardized metrics such as SNR, correlation, and RMSE. These metrics might not always be completely aligned with the clinical utility of an algorithm. However, we believe that reporting quantitative data is suitable in the first stage of algorithm development, and that further validation should be performed to confirm the clinical utility. Lastly, it is always important to consider the effect of domain shift in evaluating deep learning models. We try to demonstrate the robustness of our algorithm by training it on data very unlike the testing datasets. The algorithm will likely need tweaking under certain circumstances. To aid in tweaking, we provide instructions for debugging the algorithm along with the source code.

Future directions also include improvements to the methodology. While we have aimed to keep the building blocks of our algorithm modular to enable further improvements and modifications, an end-to-end approach might prove superior, especially for the segmentation-to-trace conversion, as it is difficult to build rule-based solutions in cases of overlapping signals, see for example \Cref{fig:precordial-iphone} where noise in lead V5 interferes with the other leads. In addition, the limb leads are linearly dependent, a feature which we have not exploited in this work. By utilizing this, it is likely possible to get even more robust digitization results. For example, the missing part of lead III in \Cref{fig:limb-scanner} could be filled in with good precision even though it is missing.

\section*{Methods}
The following section contains the methodology for the digitization framework, training, and data collection for the clinical validation. All development was done on synthetic data, and clinical data were only used in the prospective testing. The study was approved by the Norwegian Directorate of Health, case no. 21/39600-3, dated 01 July 2022.

Our proposed digitization pipeline comprises five main modules, namely (1) semantic segmentation, (2) perspective correction and cropping, (3) layout identification, (4) grid size extraction, and (5) conversion to 1D time series. The first two steps are aimed at creating a high-quality 2D representation of the ECG signal: after semantic segmentation, details such as lighting and background should be removed. After perspective correction, the image coordinates should be aligned with the time-voltage axes. The last three steps involve taking the 2D representation of the segmented ECG signals and converting it to a 1D time series with physical units such as millivolt and seconds, as opposed to pixel coordinates. The pipeline is illustrated in \Cref{fig:pipeline}. The following section*s describe each module in detail.

\input{pipeline.tex}

\subsection*{Segmentation}
The initial module of our framework focuses on segmentation, tasked with identifying ECG signals, gridlines, text annotations, and background elements within the ECG images. This segmentation process facilitates the subsequent isolation and extraction of the ECG signals. We adopt a neural network in a residual U-Net configuration for the task \cite{ronneberger_u-net_2015}, due to its demonstrated performance across various semantic segmentation tasks \cite{azad2024medical}. Additional architectures were explored, but it was found that a diverse training set was more important than the choice of segmentation network, and thus, no formal ablations were deemed necessary.

The model architecture comprises eight encoder and \mbox{-decoder} blocks. The feature map widths in the encoder are: (32, 64, 128, 256, 320, 320, 320, 320). Each encoder block contains two \mbox{Conv - InstanceNorm - LeakyReLU} stages, followed by a $2\times2$ strided convolution for spatial downsampling. Bias terms are omitted in convolution layers since they are directly followed by normalization. In the decoder, each block upsamples via bilinear interpolation, concatenates skip connections, and applies a \mbox{Conv - InstanceNorm - LeakyReLU} block. A final $3\times3$ convolution projects to the four output channels.

\subsection*{Perspective correction and cropping}
The preprocessing required for an ECG image depends on whether it is scanned or photographed. For scanned paper ECGs, rotating the image is typically enough to align it with the image coordinates. However, for photographs taken with a camera (e.g., a phone), the image has some amount of perspective distortion, requiring additional transformations to properly align the ECG with the image coordinates. The same algorithm can handle both rotation and perspective adjustments, as perspective correction also corrects for rotation.

We propose a robust two-step method for correcting perspective in ECG images. The first step is to separate the grid lines from the rest of the image, a task that is handled by the segmentation network. The second step builds upon the insight that each set of parallel grid lines will correspond to a line in the angle-radius domain, even when the grid is viewed at an angle. By applying two subsequent transformations, the task of identifying the perspective reduces to finding local maxima in a grid.

The Hough transformation is a well-proven method for detecting straight line segments~\cite{duda_use_1972}. The transform builds on parameterizing lines using an angle $\theta$ and a radius from the image origin $\rho$ as
\begin{equation}
    y \sin \theta + x \cos \theta = \rho,
    \label{eq:rho-theta-line}
\end{equation}
and obtaining a dual domain that we denote the angle-radius domain, parameterized by $\rho$ and $\theta$ instead of $x$ and $y$. The maxima in the angle-radius domain correspond to lines in the image domain, but it is often hard to identify the correct maxima under moderately noisy circumstances. An illustration of the relationship between lines in the image domain and points in the angle-radius domain is shown in the two leftmost panes of \Cref{fig:angle_radius_domain}.

\begin{figure}[h!]
    \centering
    \resizebox{\columnwidth}{!}{\input{figures/perspective/angle_radius_domain.pgf}}
    \caption{The left image represents a paper with gridlines. Parallel lines viewed at an angle in the image domain end up, to a good approximation, on a line in the angle-radius domain. Sets of lines that are parallel in the image domain are mapped to vertical lines in the angle-radius domain. The green line in the angle-radius domain maps to the green star in the angle-angle domain. Note that the $\theta$-values where the line exits the plot correspond to the coordinates in the angle-angle domain.}
    \label{fig:angle_radius_domain}
\end{figure}

Now, we take advantage of the fact that we are not trying to identify a single line, but a set of nearly parallel line segments. By identifying sets of lines, the noise can be significantly suppressed. The full process, starting from segmentation, followed by angle-radius domain transformation, and angle-angle transformation, is shown in \Cref{fig:perspective-domains}. When calculating the discrete Hough transform, a set of angles $\theta$ has to be determined a priori, setting the resolution of the angle-radius domain. To ensure both good resolution and computational efficiency, we compute the transform twice, first with low resolution over the range $[-\pi/4, 3\pi/4]$, and then zoom with higher resolution. To go from the angle-radius domain to the angle-angle domain (depicted as the rightmost plot in \Cref{fig:perspective-domains}), we search for lines in the angle-radius domain, in essence, the value at $(\theta_i, \theta_j)$ is the variance of the pixels along the line going through the points $(\theta_i, \rho_\mathrm{min})$ and $(\theta_j, \rho_\mathrm{max})$ in the discretized angle-radius domain.

To clarify the relationship between the angle-radius and angle-angle domains, we map the green line in \Cref{fig:angle_radius_domain} to a green mark in the rightmost pane. Note that the values of $\theta$ where the green line exits the figure correspond to the coordinates of the green mark in the angle-angle domain. After locating the two most prominent maxima in the angle-angle domain, we essentially know the relationship between the camera and the gridded paper. All that remains is to resample and crop the image, while ensuring all of the signal is preserved. To accomplish this, the cropping is informed by the ECG signal channel of the segmentation network, as well as the two coordinates from the angle-angle domain. All the aforementioned steps, including segmentation, iterative Hough transforms, and transformation to angle-angle domain, have been implemented to run on the GPU (if available), ensuring the time spent at these steps in the process is negligible.

\begin{figure}[ht]
    \centering
    \resizebox{\columnwidth}{!}{\input{figures/perspective/hough_transform.pgf}}
    \caption{Steps to identify the perspective in a photograph of a paper ECG. The leftmost panel shows the gridlines as detected by the neural network. The middle panel shows the grid transformed into the angle-radius domain using the Hough transform. Finally, the rightmost panel illustrates the domain where the lines of maximum variance are identified. Note that the maxima are much more pronounced in this domain and correspond to the vertical and horizontal gridlines. We encourage the reader to zoom in on the middle panel at around $0$ and $\pi/2$}
    \label{fig:perspective-domains}
\end{figure}

\subsection*{Grid size extraction}
The standard units for ECG recordings are millimetres per second (usually 25 or \SI{50}{\milli\metre/\second}) along the horizontal axis and millimetres per millivolt (usually \SI{10}{\milli\metre/\milli\volt}) along the vertical axis. In the dewarped ECG image, both axes are represented in pixels. Therefore, it is necessary to estimate conversion factors between \SI{}{\milli\metre/\second}, \SI{}{\milli\metre/\milli\volt}, and pixels. To this end, we use an iterative grid search algorithm to determine the pixel spacing between grid lines in the dewarped image.

The algorithm operates independently on each axis. For clarity, its operation is described here for the horizontal axis. As input, the algorithm takes the dewarped feature map, where each pixel has a value in the range $(0,1)$, representing the probability of it being a grid line. The first step of the algorithm is to collapse the feature map from 2D to 1D by taking the sum of the columns and computing the autocorrelation. The discrete autocorrelation function $R_{xx}$ at index $m$ of a discrete signal $x$ is computed as
\begin{equation*}
    R_{xx}[m] = \sum_{n} x[n] \cdot x[n + m].
\end{equation*}
where $m$ is the lag and the sum is taken over all valid indices. We now find the correct conversion factor by matching the autocorrelation with a known template containing both major and minor gridlines, with the assumption that the major gridlines are spaced with $5d$ and the minor lines by $d$, with $d$ being the parameter of interest. The search is implemented as an adaptive grid search over $d$. \Cref{fig:autocorr-match} shows the autocorrelation and a matched template, where a suitable value for $d$ has been found.

\begin{figure}[ht]
    \centering
    \resizebox{\columnwidth}{!}{\input{figures/pipeline/autocorr_grid_search.pgf}}
    \caption{The left pane shows the autocorrelation for a column-sum signal for lags between 0 to 220 pixels. The left pane shows the template that is matched to the autocorrelation function.}
    \label{fig:autocorr-match}
\end{figure}

\subsection*{Layout identification}
ECG recordings appear in various layout formats, with $6\times2$, $12\times1$, and $3\times4$ being common in clinical practice. Accurately identifying the layout is a critical step for reliable ECG digitization.  To achieve this, we perform layout identification in two stages. First, a residual U-Net is used to segment lead text annotations (e.g., III or aVR) from the dewarped text predictions generated by the initial segmentation model.  Once the lead names are identified, their positions are estimated by computing the weighted centroids of the segmented text regions. This allows the system to infer the layout by matching detected leads to the best matching configuration, supporting both standard and arbitrary 12-lead ECG arrangements, including any subset of leads. The lead text segmentation network shares the same architectural principles as the first segmentation U-Net but is designed to be more lightweight, with feature map widths set to (32, 64, 128, 256, 256). It processes a single-channel probability map as input and outputs 13 channels, corresponding to the 12 standard leads and one background class. Notably, this algorithm allows the user to specify arbitrary layouts in a configuration file. The algorithm for finding the best match among the set of user-defined templates is described in \Cref{alg:match_layout}. The lead markers in the generated template layout are assigned 2D-positions in $[0,1]^2$, and missing lead markers are handled by assigning them a fixed distance corresponding to half the image width (parameterized by $\lambda$ in \Cref{alg:match_layout}). Thus, even if the U-Net fails to detect one or more lead markers, the lowest cost layout will still often be the correct one.

\begin{algorithm}[ht]
  \caption{Get best matching layout from lead markers.}
  \label{alg:match_layout}
  \begin{algorithmic}[1]
    \Require Set of candidate layouts $\mathcal{L} = \{L_j\}$,
      List of detected lead markers and their coordinates $\mathrm{P}$.
      
    \Ensure 
      Best matching layout $L^*$
    \State Initialize $L^* \gets \emptyset$
    \State Initialize $\lambda \gets 0.5$
    \State Initialize $\mathrm{lowest\_cost} \gets +\infty$
    \ForAll{$L_j \in \mathcal{L}$}
      \State $G \gets \mathrm{generate\_layout}(L_j)$
    \State $(M, M_{\mathrm{miss}}) \gets \mathrm{match}(\mathrm{P}, G)$
    \State $\mathrm{scale}, \mathrm{translate} \gets \mathrm{estimate\_transform}(M)$
    \State $M' \gets \mathrm{apply\_transform}(M, \text{scale}, \text{translate})$
    \State $\mathrm{cost} = \frac{1}{|G|} \left( \lambda |M_{\mathrm{miss}}| + \sum_{(p'_i, g_k) \in M'} \| p_i' - g_k \|\right)$
    \If{$\mathrm{cost} < \mathrm{lowest\_cost}$}
      \State $L^* \gets L_j$
      \State $\mathrm{lowest\_cost} \gets \textrm{cost}$
    \EndIf
    \EndFor
    \State \Return $L^*$
  \end{algorithmic}
\end{algorithm}

\subsection*{Segmentation-to-trace conversion}
The next step is to convert the dewarped segmentation output into distinct ECG lead traces. Segmented lines from the dewarped image are used as input to the lead detection process. The algorithm begins by estimating the initial lead locations. This is done by identifying connected components in the feature map. In cases where a connected component yields poor results, such as when overlapping signals cause multiple leads to merge into a single component, a snipping algorithm is applied to subdivide the component into smaller segments.

For a problematic component, the snipping algorithm attempts to divide it horizontally by drawing a path from the left to the right edge of the component, separating it into at least two disjoint parts. The path is identified via a greedy, one-step lookahead search that selects the path of least resistance, i.e., the path minimizing overlap with the segmented signal at each step. After snipping, the connected components are recalculated. This process, connected component detection followed by component snipping, is repeated iteratively until convergence or until a predefined maximum number of iterations is reached. As a result of this stage, each lead is divided into multiple disjoint connected components from the original dewarped segmentation output. The disjoint components need to be re-merged to match the number of leads in the image. This is formulated as a minimum-weight matching problem in a bipartite graph, where the leftmost endpoint of each component should be connected with the rightmost part of some other lead segment, i.e., the linear sum assignment problem. It is solved with a modified Jonker-Volgenant algorithm~\cite{crouse_implementing_2016}. To ensure a correct matching, wrap-around is needed for some components; see \Cref{fig:hungarian} for an illustration of how the component merging process looks visually.

\begin{figure}[ht]
    \centering
    \resizebox{0.9\columnwidth}{!}{\input{figures/pipeline/linear_sum_assignment_example.pgf}}
    \caption{Example of a solution to a linear sum assignment problem. Note that three of the leftmost endpoints and three of the rightmost endpoints are unmatched, illustrating that wrapped matches are dropped. Further, note that the connected component in the top right corner is matched with itself. Self-connections are later rejected as noise.}
    \label{fig:hungarian}
\end{figure}

The cost function for matching is based on a weighted Manhattan distance between component endpoints. Matches that imply a backward connection in time (i.e., most likely incorrect matches) are penalized by multiplying the cost by 2.

\section*{Training}
    \label{sec:training}

    The initial segmentation model is trained using synthetic ECG images generated using real ECGs from the CODE15\% ECG dataset~\cite{ribeiro_code-15_2021}, and a modified version of ECG-Image-Kit~\cite{shivashankara_ecg-image-kit_2024}. Each training image is procedurally augmented with randomized parameters for the width and color of the ECG line, text overlays, and perspective transformations using real photo backgrounds to simulate varied imaging conditions. The generated images follow a $3\times4$ with lead II as a rhythm lead layout. By using a dataset collected in Brazil (i.e., a big demographic shift with regards to the test datasets), random coloring, and a layout not present in the test set, we aim to verify that our algorithm can withstand significant domain shifts without degraded performance.

    Training samples are obtained by randomly cropping $1024\times1024$ pixel regions from the generated images. The loss function is a sum of soft Dice loss and focal loss, which has shown increased robustness in multi-class segmentation tasks~\cite{ma_segmentation_2020}. Both terms are suitable for unbalanced segmentation problems, including the present task, where the ECG signal is of high importance but takes up a small minority of the total number of pixels.

    Model weights are updated using the Muon optimizer~\cite{noauthor_muon_nodate} for convolutional layers, while AdamW~\cite{kingma_adam_2017, loshchilov_decoupled_2019} is applied to the affine parameters in the instance normalization layers. Using the Muon optimizer results in significantly faster convergence and improved final accuracy compared to using AdamW alone. Automatic mixed precision is employed to accelerate training. Full training details, including hardware and software, are provided in \Cref{tab:env}, and hyperparameters are provided in \Cref{tab:training}.

    \begin{table}[!htbp]
        \caption{Hardware and software specifications used for model development and training.}
        \label{tab:env}
        \centering
        \begin{tabular}{@{}ll@{}}
            \toprule
            Component                   & Specification \\
            \midrule
            System                      & Debian 12 \\
            CPU                         & Intel i9-14900KF \\
            RAM                         & 2$\times$48\,GB; 4800\,MT/s \\
            GPU                         & NVIDIA RTX 5090 (32\,GB) \\
            CUDA version                & 12.8 \\
            Programming language        & Python 3.12 \\
            Deep learning framework     & PyTorch 2.7.0 \\
            \bottomrule
        \end{tabular}
    \end{table}

    \begin{table}[!htbp]
        \caption{Training protocol used for the main model. FLOPs are calculated by assuming a 1:2 ratio of computation between the forward and backward passes, and correspond to a full training run.}
        \label{tab:training}
        \centering
        \begin{tabular}{@{}ll@{}}
            \toprule
            Parameter                        & Value \\
            \midrule
            Batch size                       & 12 \\
            Patch size                       & $1024\times1024$ \\
            Loss function                    & Soft Dice + Focal \\
            Optimizer                        & Muon \& AdamW \\
            \quad Muon momentum                    & 0.95 \\
            \quad AdamW betas                      & 0.9, 0.999 \\
            \quad Weight decay                     & 0.001\\
            \quad Initial learning rate            & 0.0037 \\
            \quad Learning rate scheduler          & Cosine to constant\\
            \quad Batches until final learning rate& 20,000\\
            \quad Final learning rate              & 0.00037 \\
            Training time                    & 5 hours \\
            Total epochs                     & 45 \\
            Model parameters                 & 22.6 M \\
            FLOPs                            & 260 PFLOP \\
            \bottomrule
        \end{tabular}
    \end{table}

\section*{Code Availability}
The full software, including instructions, can be found on GitHub: \href{https://github.com/Ahus-AIM/Open-ECG-Digitizer}{github.com/Ahus-AIM/Open-ECG-Digitizer}.
\section*{Acknowledgements}
We thank the staff at the Department of Cardiology at Akershus University Hospital for helping us in data collection, and James Weigle and Matthew Reyna for kindly evaluating our framework on the Emory Paper Digitization ECG Dataset.
\section*{Author Contributions}
E.S. and A.B. designed the software as well as designed and performed experiments for evaluation, with supervision from A.R.. E.S. and A.B. wrote the article with input from A.R.
\section*{Competing interests}
The authors declare no competing interests.
\printbibliography
\end{document}

%% file: prefix.tex
\usepackage[T1]{fontenc}

\usepackage{amsmath}%
\usepackage{amssymb}%
\usepackage{lmodern}
\usepackage{textcomp}
\usepackage[table]{xcolor}%
\usepackage{siunitx}
\usepackage{pgf}
\usepackage{float}
\usepackage{booktabs}
\usepackage{tabularx}
\usepackage{amssymb}
\usepackage{graphicx}
\usepackage{float}
\usepackage{xcolor}%
\usepackage{mleftright}%
\usepackage{xspace}%
\usepackage{graphicx}
\usepackage{hyperref}%
\usepackage{csquotes}
\usepackage{cleveref}
\usepackage{algorithm}
\usepackage{algpseudocode}
\usepackage{subcaption}
\usepackage{pgfplots}
\usepackage{subcaption}
\usepackage[english]{babel}
\usepackage[backend=biber,style=ieee]{biblatex}
\usepackage{tikz}
\usetikzlibrary{arrows.meta, positioning, shapes.geometric}
\usetikzlibrary{positioning}
\usepackage{todonotes}
\setuptodonotes{inline}

\usepackage{tcolorbox}
\usepackage{xcolor}
\usepackage{etoolbox}

\definecolor{LightRed}{RGB}{255,235,235}   
\definecolor{DarkRed}{RGB}{139,0,0}        

\newtcolorbox[auto counter, number within=section]{incontext}[1][]{
  colback=LightRed!100!white,
  colframe=LightRed, 
  coltitle=DarkRed,
  fonttitle=\bfseries\color{DarkRed},
  title={#1},
  breakable,
  before upper={%
    \pretocmd{\section}{\color{DarkRed}}{}{}%
    \pretocmd{\subsection}{\color{DarkRed}}{}{}%
    \pretocmd{\subsubsection}{\color{DarkRed}}{}{}%
    \pretocmd{\paragraph}{\color{DarkRed}}{}{}%
    \pretocmd{\subparagraph}{\color{DarkRed}}{}{}%
  }
}

\usepackage{titlesec}%
\titlelabel{\thetitle. }%
\AtEveryBibitem{
  \clearfield{url}
  \clearfield{urldate}
  \clearfield{pages}
  \clearfield{issn}
}
\renewbibmacro*{doi+eprint+url}{
  \printfield{doi}
  \newunit\newblock
  \iftoggle{bbx:eprint}{\usebibmacro{eprint}}{}
}
\pgfplotsset{compat=1.18}

%% file: figures/results/shift_histograms.pgf
\begingroup%
\makeatletter%
\begin{pgfpicture}%
\pgfpathrectangle{\pgfpointorigin}{\pgfqpoint{6.000000in}{2.500000in}}%
\pgfusepath{use as bounding box, clip}%
\begin{pgfscope}%
\pgfsetbuttcap%
\pgfsetmiterjoin%
\definecolor{currentfill}{rgb}{1.000000,1.000000,1.000000}%
\pgfsetfillcolor{currentfill}%
\pgfsetlinewidth{0.000000pt}%
\definecolor{currentstroke}{rgb}{1.000000,1.000000,1.000000}%
\pgfsetstrokecolor{currentstroke}%
\pgfsetdash{}{0pt}%
\pgfpathmoveto{\pgfqpoint{0.000000in}{0.000000in}}%
\pgfpathlineto{\pgfqpoint{6.000000in}{0.000000in}}%
\pgfpathlineto{\pgfqpoint{6.000000in}{2.500000in}}%
\pgfpathlineto{\pgfqpoint{0.000000in}{2.500000in}}%
\pgfpathlineto{\pgfqpoint{0.000000in}{0.000000in}}%
\pgfpathclose%
\pgfusepath{fill}%
\end{pgfscope}%
\begin{pgfscope}%
\pgfsetbuttcap%
\pgfsetmiterjoin%
\definecolor{currentfill}{rgb}{1.000000,1.000000,1.000000}%
\pgfsetfillcolor{currentfill}%
\pgfsetlinewidth{0.000000pt}%
\definecolor{currentstroke}{rgb}{0.000000,0.000000,0.000000}%
\pgfsetstrokecolor{currentstroke}%
\pgfsetstrokeopacity{0.000000}%
\pgfsetdash{}{0pt}%
\pgfpathmoveto{\pgfqpoint{0.794028in}{0.582778in}}%
\pgfpathlineto{\pgfqpoint{2.319836in}{0.582778in}}%
\pgfpathlineto{\pgfqpoint{2.319836in}{2.126667in}}%
\pgfpathlineto{\pgfqpoint{0.794028in}{2.126667in}}%
\pgfpathlineto{\pgfqpoint{0.794028in}{0.582778in}}%
\pgfpathclose%
\pgfusepath{fill}%
\end{pgfscope}%
\begin{pgfscope}%
\pgfpathrectangle{\pgfqpoint{0.794028in}{0.582778in}}{\pgfqpoint{1.525809in}{1.543889in}}%
\pgfusepath{clip}%
\pgfsetrectcap%
\pgfsetroundjoin%
\pgfsetlinewidth{0.803000pt}%
\definecolor{currentstroke}{rgb}{0.690196,0.690196,0.690196}%
\pgfsetstrokecolor{currentstroke}%
\pgfsetdash{}{0pt}%
\pgfpathmoveto{\pgfqpoint{0.863383in}{0.582778in}}%
\pgfpathlineto{\pgfqpoint{0.863383in}{2.126667in}}%
\pgfusepath{stroke}%
\end{pgfscope}%
\begin{pgfscope}%
\pgfsetbuttcap%
\pgfsetroundjoin%
\definecolor{currentfill}{rgb}{0.000000,0.000000,0.000000}%
\pgfsetfillcolor{currentfill}%
\pgfsetlinewidth{0.803000pt}%
\definecolor{currentstroke}{rgb}{0.000000,0.000000,0.000000}%
\pgfsetstrokecolor{currentstroke}%
\pgfsetdash{}{0pt}%
\pgfsys@defobject{currentmarker}{\pgfqpoint{0.000000in}{-0.048611in}}{\pgfqpoint{0.000000in}{0.000000in}}{%
\pgfpathmoveto{\pgfqpoint{0.000000in}{0.000000in}}%
\pgfpathlineto{\pgfqpoint{0.000000in}{-0.048611in}}%
\pgfusepath{stroke,fill}%
}%
\begin{pgfscope}%
\pgfsys@transformshift{0.863383in}{0.582778in}%
\pgfsys@useobject{currentmarker}{}%
\end{pgfscope}%
\end{pgfscope}%
\begin{pgfscope}%
\definecolor{textcolor}{rgb}{0.000000,0.000000,0.000000}%
\pgfsetstrokecolor{textcolor}%
\pgfsetfillcolor{textcolor}%
\pgftext[x=0.863383in,y=0.485556in,,top]{\color{textcolor}{\sffamily\fontsize{10.000000}{12.000000}\selectfont\catcode`\^=\active\def^{\ifmmode\sp\else\^{}\fi}\catcode`\%=\active\def
\end{pgfscope}%
\begin{pgfscope}%
\pgfpathrectangle{\pgfqpoint{0.794028in}{0.582778in}}{\pgfqpoint{1.525809in}{1.543889in}}%
\pgfusepath{clip}%
\pgfsetrectcap%
\pgfsetroundjoin%
\pgfsetlinewidth{0.803000pt}%
\definecolor{currentstroke}{rgb}{0.690196,0.690196,0.690196}%
\pgfsetstrokecolor{currentstroke}%
\pgfsetdash{}{0pt}%
\pgfpathmoveto{\pgfqpoint{1.210157in}{0.582778in}}%
\pgfpathlineto{\pgfqpoint{1.210157in}{2.126667in}}%
\pgfusepath{stroke}%
\end{pgfscope}%
\begin{pgfscope}%
\pgfsetbuttcap%
\pgfsetroundjoin%
\definecolor{currentfill}{rgb}{0.000000,0.000000,0.000000}%
\pgfsetfillcolor{currentfill}%
\pgfsetlinewidth{0.803000pt}%
\definecolor{currentstroke}{rgb}{0.000000,0.000000,0.000000}%
\pgfsetstrokecolor{currentstroke}%
\pgfsetdash{}{0pt}%
\pgfsys@defobject{currentmarker}{\pgfqpoint{0.000000in}{-0.048611in}}{\pgfqpoint{0.000000in}{0.000000in}}{%
\pgfpathmoveto{\pgfqpoint{0.000000in}{0.000000in}}%
\pgfpathlineto{\pgfqpoint{0.000000in}{-0.048611in}}%
\pgfusepath{stroke,fill}%
}%
\begin{pgfscope}%
\pgfsys@transformshift{1.210157in}{0.582778in}%
\pgfsys@useobject{currentmarker}{}%
\end{pgfscope}%
\end{pgfscope}%
\begin{pgfscope}%
\definecolor{textcolor}{rgb}{0.000000,0.000000,0.000000}%
\pgfsetstrokecolor{textcolor}%
\pgfsetfillcolor{textcolor}%
\pgftext[x=1.210157in,y=0.485556in,,top]{\color{textcolor}{\sffamily\fontsize{10.000000}{12.000000}\selectfont\catcode`\^=\active\def^{\ifmmode\sp\else\^{}\fi}\catcode`\%=\active\def
\end{pgfscope}%
\begin{pgfscope}%
\pgfpathrectangle{\pgfqpoint{0.794028in}{0.582778in}}{\pgfqpoint{1.525809in}{1.543889in}}%
\pgfusepath{clip}%
\pgfsetrectcap%
\pgfsetroundjoin%
\pgfsetlinewidth{0.803000pt}%
\definecolor{currentstroke}{rgb}{0.690196,0.690196,0.690196}%
\pgfsetstrokecolor{currentstroke}%
\pgfsetdash{}{0pt}%
\pgfpathmoveto{\pgfqpoint{1.556932in}{0.582778in}}%
\pgfpathlineto{\pgfqpoint{1.556932in}{2.126667in}}%
\pgfusepath{stroke}%
\end{pgfscope}%
\begin{pgfscope}%
\pgfsetbuttcap%
\pgfsetroundjoin%
\definecolor{currentfill}{rgb}{0.000000,0.000000,0.000000}%
\pgfsetfillcolor{currentfill}%
\pgfsetlinewidth{0.803000pt}%
\definecolor{currentstroke}{rgb}{0.000000,0.000000,0.000000}%
\pgfsetstrokecolor{currentstroke}%
\pgfsetdash{}{0pt}%
\pgfsys@defobject{currentmarker}{\pgfqpoint{0.000000in}{-0.048611in}}{\pgfqpoint{0.000000in}{0.000000in}}{%
\pgfpathmoveto{\pgfqpoint{0.000000in}{0.000000in}}%
\pgfpathlineto{\pgfqpoint{0.000000in}{-0.048611in}}%
\pgfusepath{stroke,fill}%
}%
\begin{pgfscope}%
\pgfsys@transformshift{1.556932in}{0.582778in}%
\pgfsys@useobject{currentmarker}{}%
\end{pgfscope}%
\end{pgfscope}%
\begin{pgfscope}%
\definecolor{textcolor}{rgb}{0.000000,0.000000,0.000000}%
\pgfsetstrokecolor{textcolor}%
\pgfsetfillcolor{textcolor}%
\pgftext[x=1.556932in,y=0.485556in,,top]{\color{textcolor}{\sffamily\fontsize{10.000000}{12.000000}\selectfont\catcode`\^=\active\def^{\ifmmode\sp\else\^{}\fi}\catcode`\%=\active\def
\end{pgfscope}%
\begin{pgfscope}%
\pgfpathrectangle{\pgfqpoint{0.794028in}{0.582778in}}{\pgfqpoint{1.525809in}{1.543889in}}%
\pgfusepath{clip}%
\pgfsetrectcap%
\pgfsetroundjoin%
\pgfsetlinewidth{0.803000pt}%
\definecolor{currentstroke}{rgb}{0.690196,0.690196,0.690196}%
\pgfsetstrokecolor{currentstroke}%
\pgfsetdash{}{0pt}%
\pgfpathmoveto{\pgfqpoint{1.903707in}{0.582778in}}%
\pgfpathlineto{\pgfqpoint{1.903707in}{2.126667in}}%
\pgfusepath{stroke}%
\end{pgfscope}%
\begin{pgfscope}%
\pgfsetbuttcap%
\pgfsetroundjoin%
\definecolor{currentfill}{rgb}{0.000000,0.000000,0.000000}%
\pgfsetfillcolor{currentfill}%
\pgfsetlinewidth{0.803000pt}%
\definecolor{currentstroke}{rgb}{0.000000,0.000000,0.000000}%
\pgfsetstrokecolor{currentstroke}%
\pgfsetdash{}{0pt}%
\pgfsys@defobject{currentmarker}{\pgfqpoint{0.000000in}{-0.048611in}}{\pgfqpoint{0.000000in}{0.000000in}}{%
\pgfpathmoveto{\pgfqpoint{0.000000in}{0.000000in}}%
\pgfpathlineto{\pgfqpoint{0.000000in}{-0.048611in}}%
\pgfusepath{stroke,fill}%
}%
\begin{pgfscope}%
\pgfsys@transformshift{1.903707in}{0.582778in}%
\pgfsys@useobject{currentmarker}{}%
\end{pgfscope}%
\end{pgfscope}%
\begin{pgfscope}%
\definecolor{textcolor}{rgb}{0.000000,0.000000,0.000000}%
\pgfsetstrokecolor{textcolor}%
\pgfsetfillcolor{textcolor}%
\pgftext[x=1.903707in,y=0.485556in,,top]{\color{textcolor}{\sffamily\fontsize{10.000000}{12.000000}\selectfont\catcode`\^=\active\def^{\ifmmode\sp\else\^{}\fi}\catcode`\%=\active\def
\end{pgfscope}%
\begin{pgfscope}%
\pgfpathrectangle{\pgfqpoint{0.794028in}{0.582778in}}{\pgfqpoint{1.525809in}{1.543889in}}%
\pgfusepath{clip}%
\pgfsetrectcap%
\pgfsetroundjoin%
\pgfsetlinewidth{0.803000pt}%
\definecolor{currentstroke}{rgb}{0.690196,0.690196,0.690196}%
\pgfsetstrokecolor{currentstroke}%
\pgfsetdash{}{0pt}%
\pgfpathmoveto{\pgfqpoint{2.250482in}{0.582778in}}%
\pgfpathlineto{\pgfqpoint{2.250482in}{2.126667in}}%
\pgfusepath{stroke}%
\end{pgfscope}%
\begin{pgfscope}%
\pgfsetbuttcap%
\pgfsetroundjoin%
\definecolor{currentfill}{rgb}{0.000000,0.000000,0.000000}%
\pgfsetfillcolor{currentfill}%
\pgfsetlinewidth{0.803000pt}%
\definecolor{currentstroke}{rgb}{0.000000,0.000000,0.000000}%
\pgfsetstrokecolor{currentstroke}%
\pgfsetdash{}{0pt}%
\pgfsys@defobject{currentmarker}{\pgfqpoint{0.000000in}{-0.048611in}}{\pgfqpoint{0.000000in}{0.000000in}}{%
\pgfpathmoveto{\pgfqpoint{0.000000in}{0.000000in}}%
\pgfpathlineto{\pgfqpoint{0.000000in}{-0.048611in}}%
\pgfusepath{stroke,fill}%
}%
\begin{pgfscope}%
\pgfsys@transformshift{2.250482in}{0.582778in}%
\pgfsys@useobject{currentmarker}{}%
\end{pgfscope}%
\end{pgfscope}%
\begin{pgfscope}%
\definecolor{textcolor}{rgb}{0.000000,0.000000,0.000000}%
\pgfsetstrokecolor{textcolor}%
\pgfsetfillcolor{textcolor}%
\pgftext[x=2.250482in,y=0.485556in,,top]{\color{textcolor}{\sffamily\fontsize{10.000000}{12.000000}\selectfont\catcode`\^=\active\def^{\ifmmode\sp\else\^{}\fi}\catcode`\%=\active\def
\end{pgfscope}%
\begin{pgfscope}%
\pgfpathrectangle{\pgfqpoint{0.794028in}{0.582778in}}{\pgfqpoint{1.525809in}{1.543889in}}%
\pgfusepath{clip}%
\pgfsetrectcap%
\pgfsetroundjoin%
\pgfsetlinewidth{0.803000pt}%
\definecolor{currentstroke}{rgb}{0.690196,0.690196,0.690196}%
\pgfsetstrokecolor{currentstroke}%
\pgfsetdash{}{0pt}%
\pgfpathmoveto{\pgfqpoint{0.794028in}{0.978352in}}%
\pgfpathlineto{\pgfqpoint{2.319836in}{0.978352in}}%
\pgfusepath{stroke}%
\end{pgfscope}%
\begin{pgfscope}%
\pgfsetbuttcap%
\pgfsetroundjoin%
\definecolor{currentfill}{rgb}{0.000000,0.000000,0.000000}%
\pgfsetfillcolor{currentfill}%
\pgfsetlinewidth{0.803000pt}%
\definecolor{currentstroke}{rgb}{0.000000,0.000000,0.000000}%
\pgfsetstrokecolor{currentstroke}%
\pgfsetdash{}{0pt}%
\pgfsys@defobject{currentmarker}{\pgfqpoint{-0.048611in}{0.000000in}}{\pgfqpoint{-0.000000in}{0.000000in}}{%
\pgfpathmoveto{\pgfqpoint{-0.000000in}{0.000000in}}%
\pgfpathlineto{\pgfqpoint{-0.048611in}{0.000000in}}%
\pgfusepath{stroke,fill}%
}%
\begin{pgfscope}%
\pgfsys@transformshift{0.794028in}{0.978352in}%
\pgfsys@useobject{currentmarker}{}%
\end{pgfscope}%
\end{pgfscope}%
\begin{pgfscope}%
\definecolor{textcolor}{rgb}{0.000000,0.000000,0.000000}%
\pgfsetstrokecolor{textcolor}%
\pgfsetfillcolor{textcolor}%
\pgftext[x=0.343344in, y=0.925591in, left, base]{\color{textcolor}{\sffamily\fontsize{10.000000}{12.000000}\selectfont\catcode`\^=\active\def^{\ifmmode\sp\else\^{}\fi}\catcode`\%=\active\def
\end{pgfscope}%
\begin{pgfscope}%
\pgfpathrectangle{\pgfqpoint{0.794028in}{0.582778in}}{\pgfqpoint{1.525809in}{1.543889in}}%
\pgfusepath{clip}%
\pgfsetrectcap%
\pgfsetroundjoin%
\pgfsetlinewidth{0.803000pt}%
\definecolor{currentstroke}{rgb}{0.690196,0.690196,0.690196}%
\pgfsetstrokecolor{currentstroke}%
\pgfsetdash{}{0pt}%
\pgfpathmoveto{\pgfqpoint{0.794028in}{1.374323in}}%
\pgfpathlineto{\pgfqpoint{2.319836in}{1.374323in}}%
\pgfusepath{stroke}%
\end{pgfscope}%
\begin{pgfscope}%
\pgfsetbuttcap%
\pgfsetroundjoin%
\definecolor{currentfill}{rgb}{0.000000,0.000000,0.000000}%
\pgfsetfillcolor{currentfill}%
\pgfsetlinewidth{0.803000pt}%
\definecolor{currentstroke}{rgb}{0.000000,0.000000,0.000000}%
\pgfsetstrokecolor{currentstroke}%
\pgfsetdash{}{0pt}%
\pgfsys@defobject{currentmarker}{\pgfqpoint{-0.048611in}{0.000000in}}{\pgfqpoint{-0.000000in}{0.000000in}}{%
\pgfpathmoveto{\pgfqpoint{-0.000000in}{0.000000in}}%
\pgfpathlineto{\pgfqpoint{-0.048611in}{0.000000in}}%
\pgfusepath{stroke,fill}%
}%
\begin{pgfscope}%
\pgfsys@transformshift{0.794028in}{1.374323in}%
\pgfsys@useobject{currentmarker}{}%
\end{pgfscope}%
\end{pgfscope}%
\begin{pgfscope}%
\definecolor{textcolor}{rgb}{0.000000,0.000000,0.000000}%
\pgfsetstrokecolor{textcolor}%
\pgfsetfillcolor{textcolor}%
\pgftext[x=0.343344in, y=1.321561in, left, base]{\color{textcolor}{\sffamily\fontsize{10.000000}{12.000000}\selectfont\catcode`\^=\active\def^{\ifmmode\sp\else\^{}\fi}\catcode`\%=\active\def
\end{pgfscope}%
\begin{pgfscope}%
\pgfpathrectangle{\pgfqpoint{0.794028in}{0.582778in}}{\pgfqpoint{1.525809in}{1.543889in}}%
\pgfusepath{clip}%
\pgfsetrectcap%
\pgfsetroundjoin%
\pgfsetlinewidth{0.803000pt}%
\definecolor{currentstroke}{rgb}{0.690196,0.690196,0.690196}%
\pgfsetstrokecolor{currentstroke}%
\pgfsetdash{}{0pt}%
\pgfpathmoveto{\pgfqpoint{0.794028in}{1.770293in}}%
\pgfpathlineto{\pgfqpoint{2.319836in}{1.770293in}}%
\pgfusepath{stroke}%
\end{pgfscope}%
\begin{pgfscope}%
\pgfsetbuttcap%
\pgfsetroundjoin%
\definecolor{currentfill}{rgb}{0.000000,0.000000,0.000000}%
\pgfsetfillcolor{currentfill}%
\pgfsetlinewidth{0.803000pt}%
\definecolor{currentstroke}{rgb}{0.000000,0.000000,0.000000}%
\pgfsetstrokecolor{currentstroke}%
\pgfsetdash{}{0pt}%
\pgfsys@defobject{currentmarker}{\pgfqpoint{-0.048611in}{0.000000in}}{\pgfqpoint{-0.000000in}{0.000000in}}{%
\pgfpathmoveto{\pgfqpoint{-0.000000in}{0.000000in}}%
\pgfpathlineto{\pgfqpoint{-0.048611in}{0.000000in}}%
\pgfusepath{stroke,fill}%
}%
\begin{pgfscope}%
\pgfsys@transformshift{0.794028in}{1.770293in}%
\pgfsys@useobject{currentmarker}{}%
\end{pgfscope}%
\end{pgfscope}%
\begin{pgfscope}%
\definecolor{textcolor}{rgb}{0.000000,0.000000,0.000000}%
\pgfsetstrokecolor{textcolor}%
\pgfsetfillcolor{textcolor}%
\pgftext[x=0.343344in, y=1.717532in, left, base]{\color{textcolor}{\sffamily\fontsize{10.000000}{12.000000}\selectfont\catcode`\^=\active\def^{\ifmmode\sp\else\^{}\fi}\catcode`\%=\active\def
\end{pgfscope}%
\begin{pgfscope}%
\definecolor{textcolor}{rgb}{0.000000,0.000000,0.000000}%
\pgfsetstrokecolor{textcolor}%
\pgfsetfillcolor{textcolor}%
\pgftext[x=0.287789in,y=1.354722in,,bottom,rotate=90.000000]{\color{textcolor}{\sffamily\fontsize{10.000000}{12.000000}\selectfont\catcode`\^=\active\def^{\ifmmode\sp\else\^{}\fi}\catcode`\%=\active\def
\end{pgfscope}%
\begin{pgfscope}%
\pgfsys@transformshift{0.896667in}{0.583333in}%
\pgftext[left,bottom]{\includegraphics[interpolate=true,width=1.043333in,height=1.493333in]{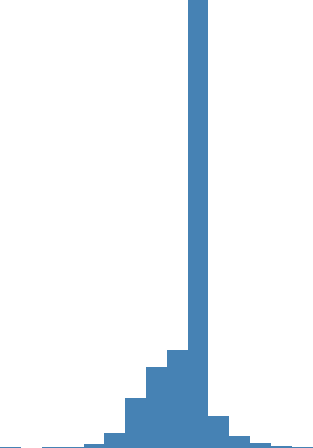}}%
\end{pgfscope}%
\begin{pgfscope}%
\pgfsetrectcap%
\pgfsetmiterjoin%
\pgfsetlinewidth{0.803000pt}%
\definecolor{currentstroke}{rgb}{0.000000,0.000000,0.000000}%
\pgfsetstrokecolor{currentstroke}%
\pgfsetdash{}{0pt}%
\pgfpathmoveto{\pgfqpoint{0.794028in}{0.582778in}}%
\pgfpathlineto{\pgfqpoint{0.794028in}{2.126667in}}%
\pgfusepath{stroke}%
\end{pgfscope}%
\begin{pgfscope}%
\pgfsetrectcap%
\pgfsetmiterjoin%
\pgfsetlinewidth{0.803000pt}%
\definecolor{currentstroke}{rgb}{0.000000,0.000000,0.000000}%
\pgfsetstrokecolor{currentstroke}%
\pgfsetdash{}{0pt}%
\pgfpathmoveto{\pgfqpoint{2.319836in}{0.582778in}}%
\pgfpathlineto{\pgfqpoint{2.319836in}{2.126667in}}%
\pgfusepath{stroke}%
\end{pgfscope}%
\begin{pgfscope}%
\pgfsetrectcap%
\pgfsetmiterjoin%
\pgfsetlinewidth{0.803000pt}%
\definecolor{currentstroke}{rgb}{0.000000,0.000000,0.000000}%
\pgfsetstrokecolor{currentstroke}%
\pgfsetdash{}{0pt}%
\pgfpathmoveto{\pgfqpoint{0.794028in}{0.582778in}}%
\pgfpathlineto{\pgfqpoint{2.319836in}{0.582778in}}%
\pgfusepath{stroke}%
\end{pgfscope}%
\begin{pgfscope}%
\pgfsetrectcap%
\pgfsetmiterjoin%
\pgfsetlinewidth{0.803000pt}%
\definecolor{currentstroke}{rgb}{0.000000,0.000000,0.000000}%
\pgfsetstrokecolor{currentstroke}%
\pgfsetdash{}{0pt}%
\pgfpathmoveto{\pgfqpoint{0.794028in}{2.126667in}}%
\pgfpathlineto{\pgfqpoint{2.319836in}{2.126667in}}%
\pgfusepath{stroke}%
\end{pgfscope}%
\begin{pgfscope}%
\definecolor{textcolor}{rgb}{0.000000,0.000000,0.000000}%
\pgfsetstrokecolor{textcolor}%
\pgfsetfillcolor{textcolor}%
\pgftext[x=1.556932in,y=2.210000in,,base]{\color{textcolor}{\sffamily\fontsize{12.000000}{14.400000}\selectfont\catcode`\^=\active\def^{\ifmmode\sp\else\^{}\fi}\catcode`\%=\active\def
\end{pgfscope}%
\begin{pgfscope}%
\pgfsetbuttcap%
\pgfsetmiterjoin%
\definecolor{currentfill}{rgb}{1.000000,1.000000,1.000000}%
\pgfsetfillcolor{currentfill}%
\pgfsetlinewidth{0.000000pt}%
\definecolor{currentstroke}{rgb}{0.000000,0.000000,0.000000}%
\pgfsetstrokecolor{currentstroke}%
\pgfsetstrokeopacity{0.000000}%
\pgfsetdash{}{0pt}%
\pgfpathmoveto{\pgfqpoint{2.546130in}{0.582778in}}%
\pgfpathlineto{\pgfqpoint{4.071939in}{0.582778in}}%
\pgfpathlineto{\pgfqpoint{4.071939in}{2.126667in}}%
\pgfpathlineto{\pgfqpoint{2.546130in}{2.126667in}}%
\pgfpathlineto{\pgfqpoint{2.546130in}{0.582778in}}%
\pgfpathclose%
\pgfusepath{fill}%
\end{pgfscope}%
\begin{pgfscope}%
\pgfpathrectangle{\pgfqpoint{2.546130in}{0.582778in}}{\pgfqpoint{1.525809in}{1.543889in}}%
\pgfusepath{clip}%
\pgfsetrectcap%
\pgfsetroundjoin%
\pgfsetlinewidth{0.803000pt}%
\definecolor{currentstroke}{rgb}{0.690196,0.690196,0.690196}%
\pgfsetstrokecolor{currentstroke}%
\pgfsetdash{}{0pt}%
\pgfpathmoveto{\pgfqpoint{2.615485in}{0.582778in}}%
\pgfpathlineto{\pgfqpoint{2.615485in}{2.126667in}}%
\pgfusepath{stroke}%
\end{pgfscope}%
\begin{pgfscope}%
\pgfsetbuttcap%
\pgfsetroundjoin%
\definecolor{currentfill}{rgb}{0.000000,0.000000,0.000000}%
\pgfsetfillcolor{currentfill}%
\pgfsetlinewidth{0.803000pt}%
\definecolor{currentstroke}{rgb}{0.000000,0.000000,0.000000}%
\pgfsetstrokecolor{currentstroke}%
\pgfsetdash{}{0pt}%
\pgfsys@defobject{currentmarker}{\pgfqpoint{0.000000in}{-0.048611in}}{\pgfqpoint{0.000000in}{0.000000in}}{%
\pgfpathmoveto{\pgfqpoint{0.000000in}{0.000000in}}%
\pgfpathlineto{\pgfqpoint{0.000000in}{-0.048611in}}%
\pgfusepath{stroke,fill}%
}%
\begin{pgfscope}%
\pgfsys@transformshift{2.615485in}{0.582778in}%
\pgfsys@useobject{currentmarker}{}%
\end{pgfscope}%
\end{pgfscope}%
\begin{pgfscope}%
\definecolor{textcolor}{rgb}{0.000000,0.000000,0.000000}%
\pgfsetstrokecolor{textcolor}%
\pgfsetfillcolor{textcolor}%
\pgftext[x=2.615485in,y=0.485556in,,top]{\color{textcolor}{\sffamily\fontsize{10.000000}{12.000000}\selectfont\catcode`\^=\active\def^{\ifmmode\sp\else\^{}\fi}\catcode`\%=\active\def
\end{pgfscope}%
\begin{pgfscope}%
\pgfpathrectangle{\pgfqpoint{2.546130in}{0.582778in}}{\pgfqpoint{1.525809in}{1.543889in}}%
\pgfusepath{clip}%
\pgfsetrectcap%
\pgfsetroundjoin%
\pgfsetlinewidth{0.803000pt}%
\definecolor{currentstroke}{rgb}{0.690196,0.690196,0.690196}%
\pgfsetstrokecolor{currentstroke}%
\pgfsetdash{}{0pt}%
\pgfpathmoveto{\pgfqpoint{2.962260in}{0.582778in}}%
\pgfpathlineto{\pgfqpoint{2.962260in}{2.126667in}}%
\pgfusepath{stroke}%
\end{pgfscope}%
\begin{pgfscope}%
\pgfsetbuttcap%
\pgfsetroundjoin%
\definecolor{currentfill}{rgb}{0.000000,0.000000,0.000000}%
\pgfsetfillcolor{currentfill}%
\pgfsetlinewidth{0.803000pt}%
\definecolor{currentstroke}{rgb}{0.000000,0.000000,0.000000}%
\pgfsetstrokecolor{currentstroke}%
\pgfsetdash{}{0pt}%
\pgfsys@defobject{currentmarker}{\pgfqpoint{0.000000in}{-0.048611in}}{\pgfqpoint{0.000000in}{0.000000in}}{%
\pgfpathmoveto{\pgfqpoint{0.000000in}{0.000000in}}%
\pgfpathlineto{\pgfqpoint{0.000000in}{-0.048611in}}%
\pgfusepath{stroke,fill}%
}%
\begin{pgfscope}%
\pgfsys@transformshift{2.962260in}{0.582778in}%
\pgfsys@useobject{currentmarker}{}%
\end{pgfscope}%
\end{pgfscope}%
\begin{pgfscope}%
\definecolor{textcolor}{rgb}{0.000000,0.000000,0.000000}%
\pgfsetstrokecolor{textcolor}%
\pgfsetfillcolor{textcolor}%
\pgftext[x=2.962260in,y=0.485556in,,top]{\color{textcolor}{\sffamily\fontsize{10.000000}{12.000000}\selectfont\catcode`\^=\active\def^{\ifmmode\sp\else\^{}\fi}\catcode`\%=\active\def
\end{pgfscope}%
\begin{pgfscope}%
\pgfpathrectangle{\pgfqpoint{2.546130in}{0.582778in}}{\pgfqpoint{1.525809in}{1.543889in}}%
\pgfusepath{clip}%
\pgfsetrectcap%
\pgfsetroundjoin%
\pgfsetlinewidth{0.803000pt}%
\definecolor{currentstroke}{rgb}{0.690196,0.690196,0.690196}%
\pgfsetstrokecolor{currentstroke}%
\pgfsetdash{}{0pt}%
\pgfpathmoveto{\pgfqpoint{3.309034in}{0.582778in}}%
\pgfpathlineto{\pgfqpoint{3.309034in}{2.126667in}}%
\pgfusepath{stroke}%
\end{pgfscope}%
\begin{pgfscope}%
\pgfsetbuttcap%
\pgfsetroundjoin%
\definecolor{currentfill}{rgb}{0.000000,0.000000,0.000000}%
\pgfsetfillcolor{currentfill}%
\pgfsetlinewidth{0.803000pt}%
\definecolor{currentstroke}{rgb}{0.000000,0.000000,0.000000}%
\pgfsetstrokecolor{currentstroke}%
\pgfsetdash{}{0pt}%
\pgfsys@defobject{currentmarker}{\pgfqpoint{0.000000in}{-0.048611in}}{\pgfqpoint{0.000000in}{0.000000in}}{%
\pgfpathmoveto{\pgfqpoint{0.000000in}{0.000000in}}%
\pgfpathlineto{\pgfqpoint{0.000000in}{-0.048611in}}%
\pgfusepath{stroke,fill}%
}%
\begin{pgfscope}%
\pgfsys@transformshift{3.309034in}{0.582778in}%
\pgfsys@useobject{currentmarker}{}%
\end{pgfscope}%
\end{pgfscope}%
\begin{pgfscope}%
\definecolor{textcolor}{rgb}{0.000000,0.000000,0.000000}%
\pgfsetstrokecolor{textcolor}%
\pgfsetfillcolor{textcolor}%
\pgftext[x=3.309034in,y=0.485556in,,top]{\color{textcolor}{\sffamily\fontsize{10.000000}{12.000000}\selectfont\catcode`\^=\active\def^{\ifmmode\sp\else\^{}\fi}\catcode`\%=\active\def
\end{pgfscope}%
\begin{pgfscope}%
\pgfpathrectangle{\pgfqpoint{2.546130in}{0.582778in}}{\pgfqpoint{1.525809in}{1.543889in}}%
\pgfusepath{clip}%
\pgfsetrectcap%
\pgfsetroundjoin%
\pgfsetlinewidth{0.803000pt}%
\definecolor{currentstroke}{rgb}{0.690196,0.690196,0.690196}%
\pgfsetstrokecolor{currentstroke}%
\pgfsetdash{}{0pt}%
\pgfpathmoveto{\pgfqpoint{3.655809in}{0.582778in}}%
\pgfpathlineto{\pgfqpoint{3.655809in}{2.126667in}}%
\pgfusepath{stroke}%
\end{pgfscope}%
\begin{pgfscope}%
\pgfsetbuttcap%
\pgfsetroundjoin%
\definecolor{currentfill}{rgb}{0.000000,0.000000,0.000000}%
\pgfsetfillcolor{currentfill}%
\pgfsetlinewidth{0.803000pt}%
\definecolor{currentstroke}{rgb}{0.000000,0.000000,0.000000}%
\pgfsetstrokecolor{currentstroke}%
\pgfsetdash{}{0pt}%
\pgfsys@defobject{currentmarker}{\pgfqpoint{0.000000in}{-0.048611in}}{\pgfqpoint{0.000000in}{0.000000in}}{%
\pgfpathmoveto{\pgfqpoint{0.000000in}{0.000000in}}%
\pgfpathlineto{\pgfqpoint{0.000000in}{-0.048611in}}%
\pgfusepath{stroke,fill}%
}%
\begin{pgfscope}%
\pgfsys@transformshift{3.655809in}{0.582778in}%
\pgfsys@useobject{currentmarker}{}%
\end{pgfscope}%
\end{pgfscope}%
\begin{pgfscope}%
\definecolor{textcolor}{rgb}{0.000000,0.000000,0.000000}%
\pgfsetstrokecolor{textcolor}%
\pgfsetfillcolor{textcolor}%
\pgftext[x=3.655809in,y=0.485556in,,top]{\color{textcolor}{\sffamily\fontsize{10.000000}{12.000000}\selectfont\catcode`\^=\active\def^{\ifmmode\sp\else\^{}\fi}\catcode`\%=\active\def
\end{pgfscope}%
\begin{pgfscope}%
\pgfpathrectangle{\pgfqpoint{2.546130in}{0.582778in}}{\pgfqpoint{1.525809in}{1.543889in}}%
\pgfusepath{clip}%
\pgfsetrectcap%
\pgfsetroundjoin%
\pgfsetlinewidth{0.803000pt}%
\definecolor{currentstroke}{rgb}{0.690196,0.690196,0.690196}%
\pgfsetstrokecolor{currentstroke}%
\pgfsetdash{}{0pt}%
\pgfpathmoveto{\pgfqpoint{4.002584in}{0.582778in}}%
\pgfpathlineto{\pgfqpoint{4.002584in}{2.126667in}}%
\pgfusepath{stroke}%
\end{pgfscope}%
\begin{pgfscope}%
\pgfsetbuttcap%
\pgfsetroundjoin%
\definecolor{currentfill}{rgb}{0.000000,0.000000,0.000000}%
\pgfsetfillcolor{currentfill}%
\pgfsetlinewidth{0.803000pt}%
\definecolor{currentstroke}{rgb}{0.000000,0.000000,0.000000}%
\pgfsetstrokecolor{currentstroke}%
\pgfsetdash{}{0pt}%
\pgfsys@defobject{currentmarker}{\pgfqpoint{0.000000in}{-0.048611in}}{\pgfqpoint{0.000000in}{0.000000in}}{%
\pgfpathmoveto{\pgfqpoint{0.000000in}{0.000000in}}%
\pgfpathlineto{\pgfqpoint{0.000000in}{-0.048611in}}%
\pgfusepath{stroke,fill}%
}%
\begin{pgfscope}%
\pgfsys@transformshift{4.002584in}{0.582778in}%
\pgfsys@useobject{currentmarker}{}%
\end{pgfscope}%
\end{pgfscope}%
\begin{pgfscope}%
\definecolor{textcolor}{rgb}{0.000000,0.000000,0.000000}%
\pgfsetstrokecolor{textcolor}%
\pgfsetfillcolor{textcolor}%
\pgftext[x=4.002584in,y=0.485556in,,top]{\color{textcolor}{\sffamily\fontsize{10.000000}{12.000000}\selectfont\catcode`\^=\active\def^{\ifmmode\sp\else\^{}\fi}\catcode`\%=\active\def
\end{pgfscope}%
\begin{pgfscope}%
\definecolor{textcolor}{rgb}{0.000000,0.000000,0.000000}%
\pgfsetstrokecolor{textcolor}%
\pgfsetfillcolor{textcolor}%
\pgftext[x=3.309034in,y=0.295587in,,top]{\color{textcolor}{\sffamily\fontsize{10.000000}{12.000000}\selectfont\catcode`\^=\active\def^{\ifmmode\sp\else\^{}\fi}\catcode`\%=\active\def
\end{pgfscope}%
\begin{pgfscope}%
\pgfpathrectangle{\pgfqpoint{2.546130in}{0.582778in}}{\pgfqpoint{1.525809in}{1.543889in}}%
\pgfusepath{clip}%
\pgfsetrectcap%
\pgfsetroundjoin%
\pgfsetlinewidth{0.803000pt}%
\definecolor{currentstroke}{rgb}{0.690196,0.690196,0.690196}%
\pgfsetstrokecolor{currentstroke}%
\pgfsetdash{}{0pt}%
\pgfpathmoveto{\pgfqpoint{2.546130in}{0.978352in}}%
\pgfpathlineto{\pgfqpoint{4.071939in}{0.978352in}}%
\pgfusepath{stroke}%
\end{pgfscope}%
\begin{pgfscope}%
\pgfsetbuttcap%
\pgfsetroundjoin%
\definecolor{currentfill}{rgb}{0.000000,0.000000,0.000000}%
\pgfsetfillcolor{currentfill}%
\pgfsetlinewidth{0.803000pt}%
\definecolor{currentstroke}{rgb}{0.000000,0.000000,0.000000}%
\pgfsetstrokecolor{currentstroke}%
\pgfsetdash{}{0pt}%
\pgfsys@defobject{currentmarker}{\pgfqpoint{-0.048611in}{0.000000in}}{\pgfqpoint{-0.000000in}{0.000000in}}{%
\pgfpathmoveto{\pgfqpoint{-0.000000in}{0.000000in}}%
\pgfpathlineto{\pgfqpoint{-0.048611in}{0.000000in}}%
\pgfusepath{stroke,fill}%
}%
\begin{pgfscope}%
\pgfsys@transformshift{2.546130in}{0.978352in}%
\pgfsys@useobject{currentmarker}{}%
\end{pgfscope}%
\end{pgfscope}%
\begin{pgfscope}%
\pgfpathrectangle{\pgfqpoint{2.546130in}{0.582778in}}{\pgfqpoint{1.525809in}{1.543889in}}%
\pgfusepath{clip}%
\pgfsetrectcap%
\pgfsetroundjoin%
\pgfsetlinewidth{0.803000pt}%
\definecolor{currentstroke}{rgb}{0.690196,0.690196,0.690196}%
\pgfsetstrokecolor{currentstroke}%
\pgfsetdash{}{0pt}%
\pgfpathmoveto{\pgfqpoint{2.546130in}{1.374323in}}%
\pgfpathlineto{\pgfqpoint{4.071939in}{1.374323in}}%
\pgfusepath{stroke}%
\end{pgfscope}%
\begin{pgfscope}%
\pgfsetbuttcap%
\pgfsetroundjoin%
\definecolor{currentfill}{rgb}{0.000000,0.000000,0.000000}%
\pgfsetfillcolor{currentfill}%
\pgfsetlinewidth{0.803000pt}%
\definecolor{currentstroke}{rgb}{0.000000,0.000000,0.000000}%
\pgfsetstrokecolor{currentstroke}%
\pgfsetdash{}{0pt}%
\pgfsys@defobject{currentmarker}{\pgfqpoint{-0.048611in}{0.000000in}}{\pgfqpoint{-0.000000in}{0.000000in}}{%
\pgfpathmoveto{\pgfqpoint{-0.000000in}{0.000000in}}%
\pgfpathlineto{\pgfqpoint{-0.048611in}{0.000000in}}%
\pgfusepath{stroke,fill}%
}%
\begin{pgfscope}%
\pgfsys@transformshift{2.546130in}{1.374323in}%
\pgfsys@useobject{currentmarker}{}%
\end{pgfscope}%
\end{pgfscope}%
\begin{pgfscope}%
\pgfpathrectangle{\pgfqpoint{2.546130in}{0.582778in}}{\pgfqpoint{1.525809in}{1.543889in}}%
\pgfusepath{clip}%
\pgfsetrectcap%
\pgfsetroundjoin%
\pgfsetlinewidth{0.803000pt}%
\definecolor{currentstroke}{rgb}{0.690196,0.690196,0.690196}%
\pgfsetstrokecolor{currentstroke}%
\pgfsetdash{}{0pt}%
\pgfpathmoveto{\pgfqpoint{2.546130in}{1.770293in}}%
\pgfpathlineto{\pgfqpoint{4.071939in}{1.770293in}}%
\pgfusepath{stroke}%
\end{pgfscope}%
\begin{pgfscope}%
\pgfsetbuttcap%
\pgfsetroundjoin%
\definecolor{currentfill}{rgb}{0.000000,0.000000,0.000000}%
\pgfsetfillcolor{currentfill}%
\pgfsetlinewidth{0.803000pt}%
\definecolor{currentstroke}{rgb}{0.000000,0.000000,0.000000}%
\pgfsetstrokecolor{currentstroke}%
\pgfsetdash{}{0pt}%
\pgfsys@defobject{currentmarker}{\pgfqpoint{-0.048611in}{0.000000in}}{\pgfqpoint{-0.000000in}{0.000000in}}{%
\pgfpathmoveto{\pgfqpoint{-0.000000in}{0.000000in}}%
\pgfpathlineto{\pgfqpoint{-0.048611in}{0.000000in}}%
\pgfusepath{stroke,fill}%
}%
\begin{pgfscope}%
\pgfsys@transformshift{2.546130in}{1.770293in}%
\pgfsys@useobject{currentmarker}{}%
\end{pgfscope}%
\end{pgfscope}%
\begin{pgfscope}%
\pgfsys@transformshift{2.580000in}{0.583333in}%
\pgftext[left,bottom]{\includegraphics[interpolate=true,width=1.456667in,height=1.396667in]{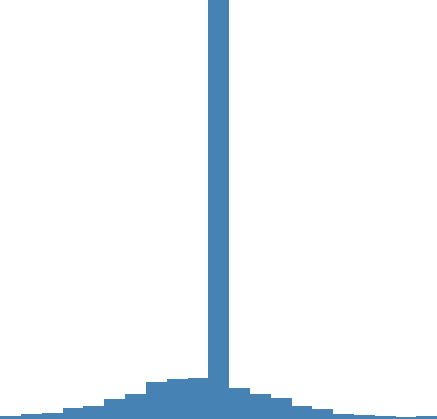}}%
\end{pgfscope}%
\begin{pgfscope}%
\pgfsetrectcap%
\pgfsetmiterjoin%
\pgfsetlinewidth{0.803000pt}%
\definecolor{currentstroke}{rgb}{0.000000,0.000000,0.000000}%
\pgfsetstrokecolor{currentstroke}%
\pgfsetdash{}{0pt}%
\pgfpathmoveto{\pgfqpoint{2.546130in}{0.582778in}}%
\pgfpathlineto{\pgfqpoint{2.546130in}{2.126667in}}%
\pgfusepath{stroke}%
\end{pgfscope}%
\begin{pgfscope}%
\pgfsetrectcap%
\pgfsetmiterjoin%
\pgfsetlinewidth{0.803000pt}%
\definecolor{currentstroke}{rgb}{0.000000,0.000000,0.000000}%
\pgfsetstrokecolor{currentstroke}%
\pgfsetdash{}{0pt}%
\pgfpathmoveto{\pgfqpoint{4.071939in}{0.582778in}}%
\pgfpathlineto{\pgfqpoint{4.071939in}{2.126667in}}%
\pgfusepath{stroke}%
\end{pgfscope}%
\begin{pgfscope}%
\pgfsetrectcap%
\pgfsetmiterjoin%
\pgfsetlinewidth{0.803000pt}%
\definecolor{currentstroke}{rgb}{0.000000,0.000000,0.000000}%
\pgfsetstrokecolor{currentstroke}%
\pgfsetdash{}{0pt}%
\pgfpathmoveto{\pgfqpoint{2.546130in}{0.582778in}}%
\pgfpathlineto{\pgfqpoint{4.071939in}{0.582778in}}%
\pgfusepath{stroke}%
\end{pgfscope}%
\begin{pgfscope}%
\pgfsetrectcap%
\pgfsetmiterjoin%
\pgfsetlinewidth{0.803000pt}%
\definecolor{currentstroke}{rgb}{0.000000,0.000000,0.000000}%
\pgfsetstrokecolor{currentstroke}%
\pgfsetdash{}{0pt}%
\pgfpathmoveto{\pgfqpoint{2.546130in}{2.126667in}}%
\pgfpathlineto{\pgfqpoint{4.071939in}{2.126667in}}%
\pgfusepath{stroke}%
\end{pgfscope}%
\begin{pgfscope}%
\definecolor{textcolor}{rgb}{0.000000,0.000000,0.000000}%
\pgfsetstrokecolor{textcolor}%
\pgfsetfillcolor{textcolor}%
\pgftext[x=3.309034in,y=2.210000in,,base]{\color{textcolor}{\sffamily\fontsize{12.000000}{14.400000}\selectfont\catcode`\^=\active\def^{\ifmmode\sp\else\^{}\fi}\catcode`\%=\active\def
\end{pgfscope}%
\begin{pgfscope}%
\pgfsetbuttcap%
\pgfsetmiterjoin%
\definecolor{currentfill}{rgb}{1.000000,1.000000,1.000000}%
\pgfsetfillcolor{currentfill}%
\pgfsetlinewidth{0.000000pt}%
\definecolor{currentstroke}{rgb}{0.000000,0.000000,0.000000}%
\pgfsetstrokecolor{currentstroke}%
\pgfsetstrokeopacity{0.000000}%
\pgfsetdash{}{0pt}%
\pgfpathmoveto{\pgfqpoint{4.298232in}{0.582778in}}%
\pgfpathlineto{\pgfqpoint{5.824041in}{0.582778in}}%
\pgfpathlineto{\pgfqpoint{5.824041in}{2.126667in}}%
\pgfpathlineto{\pgfqpoint{4.298232in}{2.126667in}}%
\pgfpathlineto{\pgfqpoint{4.298232in}{0.582778in}}%
\pgfpathclose%
\pgfusepath{fill}%
\end{pgfscope}%
\begin{pgfscope}%
\pgfpathrectangle{\pgfqpoint{4.298232in}{0.582778in}}{\pgfqpoint{1.525809in}{1.543889in}}%
\pgfusepath{clip}%
\pgfsetrectcap%
\pgfsetroundjoin%
\pgfsetlinewidth{0.803000pt}%
\definecolor{currentstroke}{rgb}{0.690196,0.690196,0.690196}%
\pgfsetstrokecolor{currentstroke}%
\pgfsetdash{}{0pt}%
\pgfpathmoveto{\pgfqpoint{4.367587in}{0.582778in}}%
\pgfpathlineto{\pgfqpoint{4.367587in}{2.126667in}}%
\pgfusepath{stroke}%
\end{pgfscope}%
\begin{pgfscope}%
\pgfsetbuttcap%
\pgfsetroundjoin%
\definecolor{currentfill}{rgb}{0.000000,0.000000,0.000000}%
\pgfsetfillcolor{currentfill}%
\pgfsetlinewidth{0.803000pt}%
\definecolor{currentstroke}{rgb}{0.000000,0.000000,0.000000}%
\pgfsetstrokecolor{currentstroke}%
\pgfsetdash{}{0pt}%
\pgfsys@defobject{currentmarker}{\pgfqpoint{0.000000in}{-0.048611in}}{\pgfqpoint{0.000000in}{0.000000in}}{%
\pgfpathmoveto{\pgfqpoint{0.000000in}{0.000000in}}%
\pgfpathlineto{\pgfqpoint{0.000000in}{-0.048611in}}%
\pgfusepath{stroke,fill}%
}%
\begin{pgfscope}%
\pgfsys@transformshift{4.367587in}{0.582778in}%
\pgfsys@useobject{currentmarker}{}%
\end{pgfscope}%
\end{pgfscope}%
\begin{pgfscope}%
\definecolor{textcolor}{rgb}{0.000000,0.000000,0.000000}%
\pgfsetstrokecolor{textcolor}%
\pgfsetfillcolor{textcolor}%
\pgftext[x=4.367587in,y=0.485556in,,top]{\color{textcolor}{\sffamily\fontsize{10.000000}{12.000000}\selectfont\catcode`\^=\active\def^{\ifmmode\sp\else\^{}\fi}\catcode`\%=\active\def
\end{pgfscope}%
\begin{pgfscope}%
\pgfpathrectangle{\pgfqpoint{4.298232in}{0.582778in}}{\pgfqpoint{1.525809in}{1.543889in}}%
\pgfusepath{clip}%
\pgfsetrectcap%
\pgfsetroundjoin%
\pgfsetlinewidth{0.803000pt}%
\definecolor{currentstroke}{rgb}{0.690196,0.690196,0.690196}%
\pgfsetstrokecolor{currentstroke}%
\pgfsetdash{}{0pt}%
\pgfpathmoveto{\pgfqpoint{4.714362in}{0.582778in}}%
\pgfpathlineto{\pgfqpoint{4.714362in}{2.126667in}}%
\pgfusepath{stroke}%
\end{pgfscope}%
\begin{pgfscope}%
\pgfsetbuttcap%
\pgfsetroundjoin%
\definecolor{currentfill}{rgb}{0.000000,0.000000,0.000000}%
\pgfsetfillcolor{currentfill}%
\pgfsetlinewidth{0.803000pt}%
\definecolor{currentstroke}{rgb}{0.000000,0.000000,0.000000}%
\pgfsetstrokecolor{currentstroke}%
\pgfsetdash{}{0pt}%
\pgfsys@defobject{currentmarker}{\pgfqpoint{0.000000in}{-0.048611in}}{\pgfqpoint{0.000000in}{0.000000in}}{%
\pgfpathmoveto{\pgfqpoint{0.000000in}{0.000000in}}%
\pgfpathlineto{\pgfqpoint{0.000000in}{-0.048611in}}%
\pgfusepath{stroke,fill}%
}%
\begin{pgfscope}%
\pgfsys@transformshift{4.714362in}{0.582778in}%
\pgfsys@useobject{currentmarker}{}%
\end{pgfscope}%
\end{pgfscope}%
\begin{pgfscope}%
\definecolor{textcolor}{rgb}{0.000000,0.000000,0.000000}%
\pgfsetstrokecolor{textcolor}%
\pgfsetfillcolor{textcolor}%
\pgftext[x=4.714362in,y=0.485556in,,top]{\color{textcolor}{\sffamily\fontsize{10.000000}{12.000000}\selectfont\catcode`\^=\active\def^{\ifmmode\sp\else\^{}\fi}\catcode`\%=\active\def
\end{pgfscope}%
\begin{pgfscope}%
\pgfpathrectangle{\pgfqpoint{4.298232in}{0.582778in}}{\pgfqpoint{1.525809in}{1.543889in}}%
\pgfusepath{clip}%
\pgfsetrectcap%
\pgfsetroundjoin%
\pgfsetlinewidth{0.803000pt}%
\definecolor{currentstroke}{rgb}{0.690196,0.690196,0.690196}%
\pgfsetstrokecolor{currentstroke}%
\pgfsetdash{}{0pt}%
\pgfpathmoveto{\pgfqpoint{5.061136in}{0.582778in}}%
\pgfpathlineto{\pgfqpoint{5.061136in}{2.126667in}}%
\pgfusepath{stroke}%
\end{pgfscope}%
\begin{pgfscope}%
\pgfsetbuttcap%
\pgfsetroundjoin%
\definecolor{currentfill}{rgb}{0.000000,0.000000,0.000000}%
\pgfsetfillcolor{currentfill}%
\pgfsetlinewidth{0.803000pt}%
\definecolor{currentstroke}{rgb}{0.000000,0.000000,0.000000}%
\pgfsetstrokecolor{currentstroke}%
\pgfsetdash{}{0pt}%
\pgfsys@defobject{currentmarker}{\pgfqpoint{0.000000in}{-0.048611in}}{\pgfqpoint{0.000000in}{0.000000in}}{%
\pgfpathmoveto{\pgfqpoint{0.000000in}{0.000000in}}%
\pgfpathlineto{\pgfqpoint{0.000000in}{-0.048611in}}%
\pgfusepath{stroke,fill}%
}%
\begin{pgfscope}%
\pgfsys@transformshift{5.061136in}{0.582778in}%
\pgfsys@useobject{currentmarker}{}%
\end{pgfscope}%
\end{pgfscope}%
\begin{pgfscope}%
\definecolor{textcolor}{rgb}{0.000000,0.000000,0.000000}%
\pgfsetstrokecolor{textcolor}%
\pgfsetfillcolor{textcolor}%
\pgftext[x=5.061136in,y=0.485556in,,top]{\color{textcolor}{\sffamily\fontsize{10.000000}{12.000000}\selectfont\catcode`\^=\active\def^{\ifmmode\sp\else\^{}\fi}\catcode`\%=\active\def
\end{pgfscope}%
\begin{pgfscope}%
\pgfpathrectangle{\pgfqpoint{4.298232in}{0.582778in}}{\pgfqpoint{1.525809in}{1.543889in}}%
\pgfusepath{clip}%
\pgfsetrectcap%
\pgfsetroundjoin%
\pgfsetlinewidth{0.803000pt}%
\definecolor{currentstroke}{rgb}{0.690196,0.690196,0.690196}%
\pgfsetstrokecolor{currentstroke}%
\pgfsetdash{}{0pt}%
\pgfpathmoveto{\pgfqpoint{5.407911in}{0.582778in}}%
\pgfpathlineto{\pgfqpoint{5.407911in}{2.126667in}}%
\pgfusepath{stroke}%
\end{pgfscope}%
\begin{pgfscope}%
\pgfsetbuttcap%
\pgfsetroundjoin%
\definecolor{currentfill}{rgb}{0.000000,0.000000,0.000000}%
\pgfsetfillcolor{currentfill}%
\pgfsetlinewidth{0.803000pt}%
\definecolor{currentstroke}{rgb}{0.000000,0.000000,0.000000}%
\pgfsetstrokecolor{currentstroke}%
\pgfsetdash{}{0pt}%
\pgfsys@defobject{currentmarker}{\pgfqpoint{0.000000in}{-0.048611in}}{\pgfqpoint{0.000000in}{0.000000in}}{%
\pgfpathmoveto{\pgfqpoint{0.000000in}{0.000000in}}%
\pgfpathlineto{\pgfqpoint{0.000000in}{-0.048611in}}%
\pgfusepath{stroke,fill}%
}%
\begin{pgfscope}%
\pgfsys@transformshift{5.407911in}{0.582778in}%
\pgfsys@useobject{currentmarker}{}%
\end{pgfscope}%
\end{pgfscope}%
\begin{pgfscope}%
\definecolor{textcolor}{rgb}{0.000000,0.000000,0.000000}%
\pgfsetstrokecolor{textcolor}%
\pgfsetfillcolor{textcolor}%
\pgftext[x=5.407911in,y=0.485556in,,top]{\color{textcolor}{\sffamily\fontsize{10.000000}{12.000000}\selectfont\catcode`\^=\active\def^{\ifmmode\sp\else\^{}\fi}\catcode`\%=\active\def
\end{pgfscope}%
\begin{pgfscope}%
\pgfpathrectangle{\pgfqpoint{4.298232in}{0.582778in}}{\pgfqpoint{1.525809in}{1.543889in}}%
\pgfusepath{clip}%
\pgfsetrectcap%
\pgfsetroundjoin%
\pgfsetlinewidth{0.803000pt}%
\definecolor{currentstroke}{rgb}{0.690196,0.690196,0.690196}%
\pgfsetstrokecolor{currentstroke}%
\pgfsetdash{}{0pt}%
\pgfpathmoveto{\pgfqpoint{5.754686in}{0.582778in}}%
\pgfpathlineto{\pgfqpoint{5.754686in}{2.126667in}}%
\pgfusepath{stroke}%
\end{pgfscope}%
\begin{pgfscope}%
\pgfsetbuttcap%
\pgfsetroundjoin%
\definecolor{currentfill}{rgb}{0.000000,0.000000,0.000000}%
\pgfsetfillcolor{currentfill}%
\pgfsetlinewidth{0.803000pt}%
\definecolor{currentstroke}{rgb}{0.000000,0.000000,0.000000}%
\pgfsetstrokecolor{currentstroke}%
\pgfsetdash{}{0pt}%
\pgfsys@defobject{currentmarker}{\pgfqpoint{0.000000in}{-0.048611in}}{\pgfqpoint{0.000000in}{0.000000in}}{%
\pgfpathmoveto{\pgfqpoint{0.000000in}{0.000000in}}%
\pgfpathlineto{\pgfqpoint{0.000000in}{-0.048611in}}%
\pgfusepath{stroke,fill}%
}%
\begin{pgfscope}%
\pgfsys@transformshift{5.754686in}{0.582778in}%
\pgfsys@useobject{currentmarker}{}%
\end{pgfscope}%
\end{pgfscope}%
\begin{pgfscope}%
\definecolor{textcolor}{rgb}{0.000000,0.000000,0.000000}%
\pgfsetstrokecolor{textcolor}%
\pgfsetfillcolor{textcolor}%
\pgftext[x=5.754686in,y=0.485556in,,top]{\color{textcolor}{\sffamily\fontsize{10.000000}{12.000000}\selectfont\catcode`\^=\active\def^{\ifmmode\sp\else\^{}\fi}\catcode`\%=\active\def
\end{pgfscope}%
\begin{pgfscope}%
\pgfpathrectangle{\pgfqpoint{4.298232in}{0.582778in}}{\pgfqpoint{1.525809in}{1.543889in}}%
\pgfusepath{clip}%
\pgfsetrectcap%
\pgfsetroundjoin%
\pgfsetlinewidth{0.803000pt}%
\definecolor{currentstroke}{rgb}{0.690196,0.690196,0.690196}%
\pgfsetstrokecolor{currentstroke}%
\pgfsetdash{}{0pt}%
\pgfpathmoveto{\pgfqpoint{4.298232in}{0.978352in}}%
\pgfpathlineto{\pgfqpoint{5.824041in}{0.978352in}}%
\pgfusepath{stroke}%
\end{pgfscope}%
\begin{pgfscope}%
\pgfsetbuttcap%
\pgfsetroundjoin%
\definecolor{currentfill}{rgb}{0.000000,0.000000,0.000000}%
\pgfsetfillcolor{currentfill}%
\pgfsetlinewidth{0.803000pt}%
\definecolor{currentstroke}{rgb}{0.000000,0.000000,0.000000}%
\pgfsetstrokecolor{currentstroke}%
\pgfsetdash{}{0pt}%
\pgfsys@defobject{currentmarker}{\pgfqpoint{-0.048611in}{0.000000in}}{\pgfqpoint{-0.000000in}{0.000000in}}{%
\pgfpathmoveto{\pgfqpoint{-0.000000in}{0.000000in}}%
\pgfpathlineto{\pgfqpoint{-0.048611in}{0.000000in}}%
\pgfusepath{stroke,fill}%
}%
\begin{pgfscope}%
\pgfsys@transformshift{4.298232in}{0.978352in}%
\pgfsys@useobject{currentmarker}{}%
\end{pgfscope}%
\end{pgfscope}%
\begin{pgfscope}%
\pgfpathrectangle{\pgfqpoint{4.298232in}{0.582778in}}{\pgfqpoint{1.525809in}{1.543889in}}%
\pgfusepath{clip}%
\pgfsetrectcap%
\pgfsetroundjoin%
\pgfsetlinewidth{0.803000pt}%
\definecolor{currentstroke}{rgb}{0.690196,0.690196,0.690196}%
\pgfsetstrokecolor{currentstroke}%
\pgfsetdash{}{0pt}%
\pgfpathmoveto{\pgfqpoint{4.298232in}{1.374323in}}%
\pgfpathlineto{\pgfqpoint{5.824041in}{1.374323in}}%
\pgfusepath{stroke}%
\end{pgfscope}%
\begin{pgfscope}%
\pgfsetbuttcap%
\pgfsetroundjoin%
\definecolor{currentfill}{rgb}{0.000000,0.000000,0.000000}%
\pgfsetfillcolor{currentfill}%
\pgfsetlinewidth{0.803000pt}%
\definecolor{currentstroke}{rgb}{0.000000,0.000000,0.000000}%
\pgfsetstrokecolor{currentstroke}%
\pgfsetdash{}{0pt}%
\pgfsys@defobject{currentmarker}{\pgfqpoint{-0.048611in}{0.000000in}}{\pgfqpoint{-0.000000in}{0.000000in}}{%
\pgfpathmoveto{\pgfqpoint{-0.000000in}{0.000000in}}%
\pgfpathlineto{\pgfqpoint{-0.048611in}{0.000000in}}%
\pgfusepath{stroke,fill}%
}%
\begin{pgfscope}%
\pgfsys@transformshift{4.298232in}{1.374323in}%
\pgfsys@useobject{currentmarker}{}%
\end{pgfscope}%
\end{pgfscope}%
\begin{pgfscope}%
\pgfpathrectangle{\pgfqpoint{4.298232in}{0.582778in}}{\pgfqpoint{1.525809in}{1.543889in}}%
\pgfusepath{clip}%
\pgfsetrectcap%
\pgfsetroundjoin%
\pgfsetlinewidth{0.803000pt}%
\definecolor{currentstroke}{rgb}{0.690196,0.690196,0.690196}%
\pgfsetstrokecolor{currentstroke}%
\pgfsetdash{}{0pt}%
\pgfpathmoveto{\pgfqpoint{4.298232in}{1.770293in}}%
\pgfpathlineto{\pgfqpoint{5.824041in}{1.770293in}}%
\pgfusepath{stroke}%
\end{pgfscope}%
\begin{pgfscope}%
\pgfsetbuttcap%
\pgfsetroundjoin%
\definecolor{currentfill}{rgb}{0.000000,0.000000,0.000000}%
\pgfsetfillcolor{currentfill}%
\pgfsetlinewidth{0.803000pt}%
\definecolor{currentstroke}{rgb}{0.000000,0.000000,0.000000}%
\pgfsetstrokecolor{currentstroke}%
\pgfsetdash{}{0pt}%
\pgfsys@defobject{currentmarker}{\pgfqpoint{-0.048611in}{0.000000in}}{\pgfqpoint{-0.000000in}{0.000000in}}{%
\pgfpathmoveto{\pgfqpoint{-0.000000in}{0.000000in}}%
\pgfpathlineto{\pgfqpoint{-0.048611in}{0.000000in}}%
\pgfusepath{stroke,fill}%
}%
\begin{pgfscope}%
\pgfsys@transformshift{4.298232in}{1.770293in}%
\pgfsys@useobject{currentmarker}{}%
\end{pgfscope}%
\end{pgfscope}%
\begin{pgfscope}%
\pgfsys@transformshift{4.296667in}{0.583333in}%
\pgftext[left,bottom]{\includegraphics[interpolate=true,width=1.493333in,height=1.386667in]{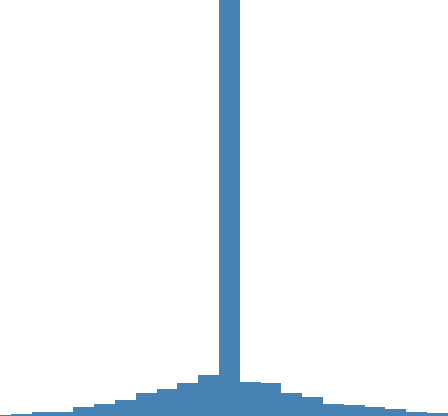}}%
\end{pgfscope}%
\begin{pgfscope}%
\pgfsetrectcap%
\pgfsetmiterjoin%
\pgfsetlinewidth{0.803000pt}%
\definecolor{currentstroke}{rgb}{0.000000,0.000000,0.000000}%
\pgfsetstrokecolor{currentstroke}%
\pgfsetdash{}{0pt}%
\pgfpathmoveto{\pgfqpoint{4.298232in}{0.582778in}}%
\pgfpathlineto{\pgfqpoint{4.298232in}{2.126667in}}%
\pgfusepath{stroke}%
\end{pgfscope}%
\begin{pgfscope}%
\pgfsetrectcap%
\pgfsetmiterjoin%
\pgfsetlinewidth{0.803000pt}%
\definecolor{currentstroke}{rgb}{0.000000,0.000000,0.000000}%
\pgfsetstrokecolor{currentstroke}%
\pgfsetdash{}{0pt}%
\pgfpathmoveto{\pgfqpoint{5.824041in}{0.582778in}}%
\pgfpathlineto{\pgfqpoint{5.824041in}{2.126667in}}%
\pgfusepath{stroke}%
\end{pgfscope}%
\begin{pgfscope}%
\pgfsetrectcap%
\pgfsetmiterjoin%
\pgfsetlinewidth{0.803000pt}%
\definecolor{currentstroke}{rgb}{0.000000,0.000000,0.000000}%
\pgfsetstrokecolor{currentstroke}%
\pgfsetdash{}{0pt}%
\pgfpathmoveto{\pgfqpoint{4.298232in}{0.582778in}}%
\pgfpathlineto{\pgfqpoint{5.824041in}{0.582778in}}%
\pgfusepath{stroke}%
\end{pgfscope}%
\begin{pgfscope}%
\pgfsetrectcap%
\pgfsetmiterjoin%
\pgfsetlinewidth{0.803000pt}%
\definecolor{currentstroke}{rgb}{0.000000,0.000000,0.000000}%
\pgfsetstrokecolor{currentstroke}%
\pgfsetdash{}{0pt}%
\pgfpathmoveto{\pgfqpoint{4.298232in}{2.126667in}}%
\pgfpathlineto{\pgfqpoint{5.824041in}{2.126667in}}%
\pgfusepath{stroke}%
\end{pgfscope}%
\begin{pgfscope}%
\definecolor{textcolor}{rgb}{0.000000,0.000000,0.000000}%
\pgfsetstrokecolor{textcolor}%
\pgfsetfillcolor{textcolor}%
\pgftext[x=5.061136in,y=2.210000in,,base]{\color{textcolor}{\sffamily\fontsize{12.000000}{14.400000}\selectfont\catcode`\^=\active\def^{\ifmmode\sp\else\^{}\fi}\catcode`\%=\active\def
\end{pgfscope}%
\end{pgfpicture}%
\makeatother%
\endgroup%

%% file: pipeline.tex
\begin{figure*}[htpb]
    \centering
    \resizebox{0.9\textwidth}{!}{
    \begin{tikzpicture}[
        fancybox/.style={
          draw,
          thick,
          rounded corners,
          fill={rgb,255:red,80;green,140;blue,200},
          minimum width=3.5cm,
          minimum height=1.3cm,
          align=center,
          font=\sffamily
        },
        fancyarrow/.style={
            -{Latex[round]}, thick,
        },
        invisiblenode/.style={
            draw=none, minimum size=0cm, font=\sffamily
        },
        node distance=2cm and 4cm
    ]

    \node[fancybox] (segmentation) at (-6, 0) {Semantic\\segmentation};
    \node[fancybox] (perspective) at (-2, 0) {Perspective correction \\ and cropping};
    \node[fancybox] (grid) at (3, -1.7) {Grid size\\extraction};
    \node[fancybox] (conversion) at (3, 0) {Segmentation-to-trace\\conversion};
    \node[fancybox] (layout) at (3, 1.7) {Layout\\identification};

    \node[invisiblenode] (image) at (-8.7, 0) {Image};
    \node[invisiblenode] (signal) at (7.5, 0) {ECG signal};

    \draw[fancyarrow] (image.east) to[out=0,in=180] (segmentation.west);
    \draw[fancyarrow] (segmentation.east) to[out=0,in=180,looseness=1] (perspective.west);

    \draw[fancyarrow] (perspective.east) to[out=0,in=180] (grid.west);
    \draw[fancyarrow] (perspective.east) to[out=0,in=180] (conversion.west);

    \draw[fancyarrow] (grid.east) to[in=180,out=0] (signal.west);
    \draw[fancyarrow] (conversion.east) to[in=180,out=0] (signal.west);
    \draw[fancyarrow] (layout.east) to[in=180,out=0] (signal.west);
    \draw[fancyarrow] (perspective.east) to[out=0,in=180,looseness=1] (layout.west);

    \end{tikzpicture}
    }
    \caption{The digitization pipeline consists of semantic segmentation, that separates gridlines, ecg signal, text and background, followed by perspective correction and cropping. These first two modules produce an image where the gridlines on the paper are aligned with the image. The last part converts the aligned image to 1D signals, identifies the lead layout and outputs a structured time series in the desired sample rate.}
    \label{fig:pipeline}
\end{figure*}

%% file: figures/perspective/angle_radius_domain.pgf
\begingroup%
\makeatletter%
\begin{pgfpicture}%
\pgfpathrectangle{\pgfpointorigin}{\pgfqpoint{13.888826in}{4.372741in}}%
\pgfusepath{use as bounding box, clip}%
\begin{pgfscope}%
\pgfsetbuttcap%
\pgfsetmiterjoin%
\definecolor{currentfill}{rgb}{1.000000,1.000000,1.000000}%
\pgfsetfillcolor{currentfill}%
\pgfsetlinewidth{0.000000pt}%
\definecolor{currentstroke}{rgb}{1.000000,1.000000,1.000000}%
\pgfsetstrokecolor{currentstroke}%
\pgfsetdash{}{0pt}%
\pgfpathmoveto{\pgfqpoint{0.000000in}{0.000000in}}%
\pgfpathlineto{\pgfqpoint{13.888826in}{0.000000in}}%
\pgfpathlineto{\pgfqpoint{13.888826in}{4.372741in}}%
\pgfpathlineto{\pgfqpoint{0.000000in}{4.372741in}}%
\pgfpathlineto{\pgfqpoint{0.000000in}{0.000000in}}%
\pgfpathclose%
\pgfusepath{fill}%
\end{pgfscope}%
\begin{pgfscope}%
\pgfsetbuttcap%
\pgfsetmiterjoin%
\definecolor{currentfill}{rgb}{1.000000,1.000000,1.000000}%
\pgfsetfillcolor{currentfill}%
\pgfsetlinewidth{0.000000pt}%
\definecolor{currentstroke}{rgb}{0.000000,0.000000,0.000000}%
\pgfsetstrokecolor{currentstroke}%
\pgfsetstrokeopacity{0.000000}%
\pgfsetdash{}{0pt}%
\pgfpathmoveto{\pgfqpoint{0.424381in}{0.809159in}}%
\pgfpathlineto{\pgfqpoint{4.479032in}{0.809159in}}%
\pgfpathlineto{\pgfqpoint{4.479032in}{3.859492in}}%
\pgfpathlineto{\pgfqpoint{0.424381in}{3.859492in}}%
\pgfpathlineto{\pgfqpoint{0.424381in}{0.809159in}}%
\pgfpathclose%
\pgfusepath{fill}%
\end{pgfscope}%
\begin{pgfscope}%
\pgfpathrectangle{\pgfqpoint{0.424381in}{0.809159in}}{\pgfqpoint{4.054651in}{3.050333in}}%
\pgfusepath{clip}%
\pgfsetbuttcap%
\pgfsetmiterjoin%
\pgfsetlinewidth{1.505625pt}%
\definecolor{currentstroke}{rgb}{0.000000,0.000000,0.000000}%
\pgfsetstrokecolor{currentstroke}%
\pgfsetdash{}{0pt}%
\pgfpathmoveto{\pgfqpoint{1.161590in}{0.809159in}}%
\pgfpathcurveto{\pgfqpoint{1.161590in}{0.851929in}}{\pgfqpoint{1.157664in}{0.894613in}}{\pgfqpoint{1.149857in}{0.936703in}}%
\pgfpathcurveto{\pgfqpoint{1.142050in}{0.978792in}}{\pgfqpoint{1.130395in}{1.020118in}}{\pgfqpoint{1.115029in}{1.060186in}}%
\pgfusepath{stroke}%
\end{pgfscope}%
\begin{pgfscope}%
\definecolor{textcolor}{rgb}{0.000000,0.000000,0.000000}%
\pgfsetstrokecolor{textcolor}%
\pgfsetfillcolor{textcolor}%
\pgftext[x=2.451707in,y=0.753603in,,top]{\color{textcolor}{\sffamily\fontsize{20.000000}{24.000000}\selectfont\catcode`\^=\active\def^{\ifmmode\sp\else\^{}\fi}\catcode`\%=\active\def
\end{pgfscope}%
\begin{pgfscope}%
\definecolor{textcolor}{rgb}{0.000000,0.000000,0.000000}%
\pgfsetstrokecolor{textcolor}%
\pgfsetfillcolor{textcolor}%
\pgftext[x=0.368826in,y=2.334325in,,bottom,rotate=90.000000]{\color{textcolor}{\sffamily\fontsize{20.000000}{24.000000}\selectfont\catcode`\^=\active\def^{\ifmmode\sp\else\^{}\fi}\catcode`\%=\active\def
\end{pgfscope}%
\begin{pgfscope}%
\pgfpathrectangle{\pgfqpoint{0.424381in}{0.809159in}}{\pgfqpoint{4.054651in}{3.050333in}}%
\pgfusepath{clip}%
\pgfsetrectcap%
\pgfsetroundjoin%
\pgfsetlinewidth{1.505625pt}%
\definecolor{currentstroke}{rgb}{0.121569,0.466667,0.705882}%
\pgfsetstrokecolor{currentstroke}%
\pgfsetdash{}{0pt}%
\pgfpathmoveto{\pgfqpoint{0.697874in}{3.869492in}}%
\pgfpathlineto{\pgfqpoint{0.978201in}{0.799159in}}%
\pgfusepath{stroke}%
\end{pgfscope}%
\begin{pgfscope}%
\pgfpathrectangle{\pgfqpoint{0.424381in}{0.809159in}}{\pgfqpoint{4.054651in}{3.050333in}}%
\pgfusepath{clip}%
\pgfsetrectcap%
\pgfsetroundjoin%
\pgfsetlinewidth{1.505625pt}%
\definecolor{currentstroke}{rgb}{0.121569,0.466667,0.705882}%
\pgfsetstrokecolor{currentstroke}%
\pgfsetdash{}{0pt}%
\pgfpathmoveto{\pgfqpoint{1.064110in}{3.869492in}}%
\pgfpathlineto{\pgfqpoint{1.568699in}{0.799159in}}%
\pgfusepath{stroke}%
\end{pgfscope}%
\begin{pgfscope}%
\pgfpathrectangle{\pgfqpoint{0.424381in}{0.809159in}}{\pgfqpoint{4.054651in}{3.050333in}}%
\pgfusepath{clip}%
\pgfsetrectcap%
\pgfsetroundjoin%
\pgfsetlinewidth{1.505625pt}%
\definecolor{currentstroke}{rgb}{0.121569,0.466667,0.705882}%
\pgfsetstrokecolor{currentstroke}%
\pgfsetdash{}{0pt}%
\pgfpathmoveto{\pgfqpoint{1.430346in}{3.869492in}}%
\pgfpathlineto{\pgfqpoint{2.159197in}{0.799159in}}%
\pgfusepath{stroke}%
\end{pgfscope}%
\begin{pgfscope}%
\pgfpathrectangle{\pgfqpoint{0.424381in}{0.809159in}}{\pgfqpoint{4.054651in}{3.050333in}}%
\pgfusepath{clip}%
\pgfsetrectcap%
\pgfsetroundjoin%
\pgfsetlinewidth{1.505625pt}%
\definecolor{currentstroke}{rgb}{0.121569,0.466667,0.705882}%
\pgfsetstrokecolor{currentstroke}%
\pgfsetdash{}{0pt}%
\pgfpathmoveto{\pgfqpoint{1.796582in}{3.869492in}}%
\pgfpathlineto{\pgfqpoint{2.749695in}{0.799159in}}%
\pgfusepath{stroke}%
\end{pgfscope}%
\begin{pgfscope}%
\pgfpathrectangle{\pgfqpoint{0.424381in}{0.809159in}}{\pgfqpoint{4.054651in}{3.050333in}}%
\pgfusepath{clip}%
\pgfsetrectcap%
\pgfsetroundjoin%
\pgfsetlinewidth{1.505625pt}%
\definecolor{currentstroke}{rgb}{0.000000,0.000000,0.000000}%
\pgfsetstrokecolor{currentstroke}%
\pgfsetdash{}{0pt}%
\pgfpathmoveto{\pgfqpoint{0.424381in}{0.809159in}}%
\pgfpathlineto{\pgfqpoint{1.063321in}{1.041392in}}%
\pgfpathlineto{\pgfqpoint{1.702261in}{1.273625in}}%
\pgfpathlineto{\pgfqpoint{2.341201in}{1.505859in}}%
\pgfpathlineto{\pgfqpoint{2.980141in}{1.738092in}}%
\pgfusepath{stroke}%
\end{pgfscope}%
\begin{pgfscope}%
\pgfpathrectangle{\pgfqpoint{0.424381in}{0.809159in}}{\pgfqpoint{4.054651in}{3.050333in}}%
\pgfusepath{clip}%
\pgfsetrectcap%
\pgfsetroundjoin%
\pgfsetlinewidth{1.505625pt}%
\definecolor{currentstroke}{rgb}{0.121569,0.466667,0.705882}%
\pgfsetstrokecolor{currentstroke}%
\pgfsetdash{}{0pt}%
\pgfpathmoveto{\pgfqpoint{2.162818in}{3.869492in}}%
\pgfpathlineto{\pgfqpoint{3.188916in}{1.193655in}}%
\pgfpathlineto{\pgfqpoint{3.340192in}{0.799159in}}%
\pgfusepath{stroke}%
\end{pgfscope}%
\begin{pgfscope}%
\pgfpathrectangle{\pgfqpoint{0.424381in}{0.809159in}}{\pgfqpoint{4.054651in}{3.050333in}}%
\pgfusepath{clip}%
\pgfsetrectcap%
\pgfsetroundjoin%
\pgfsetlinewidth{1.505625pt}%
\definecolor{currentstroke}{rgb}{0.121569,0.466667,0.705882}%
\pgfsetstrokecolor{currentstroke}%
\pgfsetdash{}{0pt}%
\pgfpathmoveto{\pgfqpoint{2.529054in}{3.869492in}}%
\pgfpathlineto{\pgfqpoint{3.188916in}{2.424041in}}%
\pgfpathlineto{\pgfqpoint{3.930690in}{0.799159in}}%
\pgfusepath{stroke}%
\end{pgfscope}%
\begin{pgfscope}%
\pgfpathrectangle{\pgfqpoint{0.424381in}{0.809159in}}{\pgfqpoint{4.054651in}{3.050333in}}%
\pgfusepath{clip}%
\pgfsetrectcap%
\pgfsetroundjoin%
\pgfsetlinewidth{1.505625pt}%
\definecolor{currentstroke}{rgb}{0.839216,0.152941,0.156863}%
\pgfsetstrokecolor{currentstroke}%
\pgfsetdash{}{0pt}%
\pgfpathmoveto{\pgfqpoint{0.414381in}{1.347453in}}%
\pgfpathlineto{\pgfqpoint{0.424381in}{1.347453in}}%
\pgfpathlineto{\pgfqpoint{3.188916in}{1.347453in}}%
\pgfpathlineto{\pgfqpoint{4.489032in}{1.347453in}}%
\pgfusepath{stroke}%
\end{pgfscope}%
\begin{pgfscope}%
\pgfpathrectangle{\pgfqpoint{0.424381in}{0.809159in}}{\pgfqpoint{4.054651in}{3.050333in}}%
\pgfusepath{clip}%
\pgfsetrectcap%
\pgfsetroundjoin%
\pgfsetlinewidth{1.505625pt}%
\definecolor{currentstroke}{rgb}{0.839216,0.152941,0.156863}%
\pgfsetstrokecolor{currentstroke}%
\pgfsetdash{}{0pt}%
\pgfpathmoveto{\pgfqpoint{0.414381in}{1.742630in}}%
\pgfpathlineto{\pgfqpoint{0.424381in}{1.742202in}}%
\pgfpathlineto{\pgfqpoint{3.188916in}{1.623777in}}%
\pgfpathlineto{\pgfqpoint{4.489032in}{1.568084in}}%
\pgfusepath{stroke}%
\end{pgfscope}%
\begin{pgfscope}%
\pgfpathrectangle{\pgfqpoint{0.424381in}{0.809159in}}{\pgfqpoint{4.054651in}{3.050333in}}%
\pgfusepath{clip}%
\pgfsetrectcap%
\pgfsetroundjoin%
\pgfsetlinewidth{1.505625pt}%
\definecolor{currentstroke}{rgb}{0.839216,0.152941,0.156863}%
\pgfsetstrokecolor{currentstroke}%
\pgfsetdash{}{0pt}%
\pgfpathmoveto{\pgfqpoint{0.414381in}{2.137808in}}%
\pgfpathlineto{\pgfqpoint{0.424381in}{2.136951in}}%
\pgfpathlineto{\pgfqpoint{3.188916in}{1.900101in}}%
\pgfpathlineto{\pgfqpoint{4.489032in}{1.788715in}}%
\pgfusepath{stroke}%
\end{pgfscope}%
\begin{pgfscope}%
\pgfpathrectangle{\pgfqpoint{0.424381in}{0.809159in}}{\pgfqpoint{4.054651in}{3.050333in}}%
\pgfusepath{clip}%
\pgfsetrectcap%
\pgfsetroundjoin%
\pgfsetlinewidth{1.505625pt}%
\definecolor{currentstroke}{rgb}{0.839216,0.152941,0.156863}%
\pgfsetstrokecolor{currentstroke}%
\pgfsetdash{}{0pt}%
\pgfpathmoveto{\pgfqpoint{0.414381in}{2.532985in}}%
\pgfpathlineto{\pgfqpoint{0.424381in}{2.531700in}}%
\pgfpathlineto{\pgfqpoint{3.188916in}{2.176426in}}%
\pgfpathlineto{\pgfqpoint{4.489032in}{2.009346in}}%
\pgfusepath{stroke}%
\end{pgfscope}%
\begin{pgfscope}%
\pgfpathrectangle{\pgfqpoint{0.424381in}{0.809159in}}{\pgfqpoint{4.054651in}{3.050333in}}%
\pgfusepath{clip}%
\pgfsetrectcap%
\pgfsetroundjoin%
\pgfsetlinewidth{1.505625pt}%
\definecolor{currentstroke}{rgb}{0.839216,0.152941,0.156863}%
\pgfsetstrokecolor{currentstroke}%
\pgfsetdash{}{0pt}%
\pgfpathmoveto{\pgfqpoint{0.414381in}{2.928162in}}%
\pgfpathlineto{\pgfqpoint{0.424381in}{2.926449in}}%
\pgfpathlineto{\pgfqpoint{3.188916in}{2.452750in}}%
\pgfpathlineto{\pgfqpoint{4.489032in}{2.229977in}}%
\pgfusepath{stroke}%
\end{pgfscope}%
\begin{pgfscope}%
\pgfpathrectangle{\pgfqpoint{0.424381in}{0.809159in}}{\pgfqpoint{4.054651in}{3.050333in}}%
\pgfusepath{clip}%
\pgfsetrectcap%
\pgfsetroundjoin%
\pgfsetlinewidth{1.505625pt}%
\definecolor{currentstroke}{rgb}{0.839216,0.152941,0.156863}%
\pgfsetstrokecolor{currentstroke}%
\pgfsetdash{}{0pt}%
\pgfpathmoveto{\pgfqpoint{0.414381in}{3.323340in}}%
\pgfpathlineto{\pgfqpoint{0.424381in}{3.321198in}}%
\pgfpathlineto{\pgfqpoint{3.188916in}{2.729074in}}%
\pgfpathlineto{\pgfqpoint{4.489032in}{2.450608in}}%
\pgfusepath{stroke}%
\end{pgfscope}%
\begin{pgfscope}%
\pgfsetrectcap%
\pgfsetmiterjoin%
\pgfsetlinewidth{1.505625pt}%
\definecolor{currentstroke}{rgb}{0.000000,0.000000,0.000000}%
\pgfsetstrokecolor{currentstroke}%
\pgfsetdash{}{0pt}%
\pgfpathmoveto{\pgfqpoint{0.424381in}{0.809159in}}%
\pgfpathlineto{\pgfqpoint{0.424381in}{3.859492in}}%
\pgfusepath{stroke}%
\end{pgfscope}%
\begin{pgfscope}%
\pgfsetrectcap%
\pgfsetmiterjoin%
\pgfsetlinewidth{1.505625pt}%
\definecolor{currentstroke}{rgb}{0.000000,0.000000,0.000000}%
\pgfsetstrokecolor{currentstroke}%
\pgfsetdash{}{0pt}%
\pgfpathmoveto{\pgfqpoint{4.479032in}{0.809159in}}%
\pgfpathlineto{\pgfqpoint{4.479032in}{3.859492in}}%
\pgfusepath{stroke}%
\end{pgfscope}%
\begin{pgfscope}%
\pgfsetrectcap%
\pgfsetmiterjoin%
\pgfsetlinewidth{1.505625pt}%
\definecolor{currentstroke}{rgb}{0.000000,0.000000,0.000000}%
\pgfsetstrokecolor{currentstroke}%
\pgfsetdash{}{0pt}%
\pgfpathmoveto{\pgfqpoint{0.424381in}{0.809159in}}%
\pgfpathlineto{\pgfqpoint{4.479032in}{0.809159in}}%
\pgfusepath{stroke}%
\end{pgfscope}%
\begin{pgfscope}%
\pgfsetrectcap%
\pgfsetmiterjoin%
\pgfsetlinewidth{1.505625pt}%
\definecolor{currentstroke}{rgb}{0.000000,0.000000,0.000000}%
\pgfsetstrokecolor{currentstroke}%
\pgfsetdash{}{0pt}%
\pgfpathmoveto{\pgfqpoint{0.424381in}{3.859492in}}%
\pgfpathlineto{\pgfqpoint{4.479032in}{3.859492in}}%
\pgfusepath{stroke}%
\end{pgfscope}%
\begin{pgfscope}%
\definecolor{textcolor}{rgb}{0.000000,0.000000,0.000000}%
\pgfsetstrokecolor{textcolor}%
\pgfsetfillcolor{textcolor}%
\pgftext[x=1.309032in,y=0.880931in,left,base]{\color{textcolor}{\sffamily\fontsize{26.000000}{31.200000}\selectfont\catcode`\^=\active\def^{\ifmmode\sp\else\^{}\fi}\catcode`\%=\active\def
\end{pgfscope}%
\begin{pgfscope}%
\definecolor{textcolor}{rgb}{0.000000,0.000000,0.000000}%
\pgfsetstrokecolor{textcolor}%
\pgfsetfillcolor{textcolor}%
\pgftext[x=1.622346in,y=1.383339in,left,base]{\color{textcolor}{\sffamily\fontsize{26.000000}{31.200000}\selectfont\catcode`\^=\active\def^{\ifmmode\sp\else\^{}\fi}\catcode`\%=\active\def
\end{pgfscope}%
\begin{pgfscope}%
\definecolor{textcolor}{rgb}{0.000000,0.000000,0.000000}%
\pgfsetstrokecolor{textcolor}%
\pgfsetfillcolor{textcolor}%
\pgftext[x=2.451707in,y=3.998381in,,base]{\color{textcolor}{\sffamily\fontsize{26.000000}{31.200000}\selectfont\catcode`\^=\active\def^{\ifmmode\sp\else\^{}\fi}\catcode`\%=\active\def
\end{pgfscope}%
\begin{pgfscope}%
\pgfsetbuttcap%
\pgfsetmiterjoin%
\definecolor{currentfill}{rgb}{1.000000,1.000000,1.000000}%
\pgfsetfillcolor{currentfill}%
\pgfsetlinewidth{0.000000pt}%
\definecolor{currentstroke}{rgb}{0.000000,0.000000,0.000000}%
\pgfsetstrokecolor{currentstroke}%
\pgfsetstrokeopacity{0.000000}%
\pgfsetdash{}{0pt}%
\pgfpathmoveto{\pgfqpoint{5.547122in}{0.809159in}}%
\pgfpathlineto{\pgfqpoint{9.601773in}{0.809159in}}%
\pgfpathlineto{\pgfqpoint{9.601773in}{3.859492in}}%
\pgfpathlineto{\pgfqpoint{5.547122in}{3.859492in}}%
\pgfpathlineto{\pgfqpoint{5.547122in}{0.809159in}}%
\pgfpathclose%
\pgfusepath{fill}%
\end{pgfscope}%
\begin{pgfscope}%
\pgfpathrectangle{\pgfqpoint{5.547122in}{0.809159in}}{\pgfqpoint{4.054651in}{3.050333in}}%
\pgfusepath{clip}%
\pgfsetrectcap%
\pgfsetroundjoin%
\pgfsetlinewidth{2.258437pt}%
\definecolor{currentstroke}{rgb}{0.000000,0.733333,0.000000}%
\pgfsetstrokecolor{currentstroke}%
\pgfsetdash{}{0pt}%
\pgfpathmoveto{\pgfqpoint{6.586598in}{0.809159in}}%
\pgfpathlineto{\pgfqpoint{7.115758in}{3.859492in}}%
\pgfusepath{stroke}%
\end{pgfscope}%
\begin{pgfscope}%
\pgfpathrectangle{\pgfqpoint{5.547122in}{0.809159in}}{\pgfqpoint{4.054651in}{3.050333in}}%
\pgfusepath{clip}%
\pgfsetrectcap%
\pgfsetroundjoin%
\pgfsetlinewidth{0.803000pt}%
\definecolor{currentstroke}{rgb}{0.690196,0.690196,0.690196}%
\pgfsetstrokecolor{currentstroke}%
\pgfsetdash{}{0pt}%
\pgfpathmoveto{\pgfqpoint{6.560785in}{0.809159in}}%
\pgfpathlineto{\pgfqpoint{6.560785in}{3.859492in}}%
\pgfusepath{stroke}%
\end{pgfscope}%
\begin{pgfscope}%
\pgfsetbuttcap%
\pgfsetroundjoin%
\definecolor{currentfill}{rgb}{0.000000,0.000000,0.000000}%
\pgfsetfillcolor{currentfill}%
\pgfsetlinewidth{0.803000pt}%
\definecolor{currentstroke}{rgb}{0.000000,0.000000,0.000000}%
\pgfsetstrokecolor{currentstroke}%
\pgfsetdash{}{0pt}%
\pgfsys@defobject{currentmarker}{\pgfqpoint{0.000000in}{-0.048611in}}{\pgfqpoint{0.000000in}{0.000000in}}{%
\pgfpathmoveto{\pgfqpoint{0.000000in}{0.000000in}}%
\pgfpathlineto{\pgfqpoint{0.000000in}{-0.048611in}}%
\pgfusepath{stroke,fill}%
}%
\begin{pgfscope}%
\pgfsys@transformshift{6.560785in}{0.809159in}%
\pgfsys@useobject{currentmarker}{}%
\end{pgfscope}%
\end{pgfscope}%
\begin{pgfscope}%
\definecolor{textcolor}{rgb}{0.000000,0.000000,0.000000}%
\pgfsetstrokecolor{textcolor}%
\pgfsetfillcolor{textcolor}%
\pgftext[x=6.560785in,y=0.711937in,,top]{\color{textcolor}{\sffamily\fontsize{20.000000}{24.000000}\selectfont\catcode`\^=\active\def^{\ifmmode\sp\else\^{}\fi}\catcode`\%=\active\def
\end{pgfscope}%
\begin{pgfscope}%
\pgfpathrectangle{\pgfqpoint{5.547122in}{0.809159in}}{\pgfqpoint{4.054651in}{3.050333in}}%
\pgfusepath{clip}%
\pgfsetrectcap%
\pgfsetroundjoin%
\pgfsetlinewidth{0.803000pt}%
\definecolor{currentstroke}{rgb}{0.690196,0.690196,0.690196}%
\pgfsetstrokecolor{currentstroke}%
\pgfsetdash{}{0pt}%
\pgfpathmoveto{\pgfqpoint{8.588110in}{0.809159in}}%
\pgfpathlineto{\pgfqpoint{8.588110in}{3.859492in}}%
\pgfusepath{stroke}%
\end{pgfscope}%
\begin{pgfscope}%
\pgfsetbuttcap%
\pgfsetroundjoin%
\definecolor{currentfill}{rgb}{0.000000,0.000000,0.000000}%
\pgfsetfillcolor{currentfill}%
\pgfsetlinewidth{0.803000pt}%
\definecolor{currentstroke}{rgb}{0.000000,0.000000,0.000000}%
\pgfsetstrokecolor{currentstroke}%
\pgfsetdash{}{0pt}%
\pgfsys@defobject{currentmarker}{\pgfqpoint{0.000000in}{-0.048611in}}{\pgfqpoint{0.000000in}{0.000000in}}{%
\pgfpathmoveto{\pgfqpoint{0.000000in}{0.000000in}}%
\pgfpathlineto{\pgfqpoint{0.000000in}{-0.048611in}}%
\pgfusepath{stroke,fill}%
}%
\begin{pgfscope}%
\pgfsys@transformshift{8.588110in}{0.809159in}%
\pgfsys@useobject{currentmarker}{}%
\end{pgfscope}%
\end{pgfscope}%
\begin{pgfscope}%
\definecolor{textcolor}{rgb}{0.000000,0.000000,0.000000}%
\pgfsetstrokecolor{textcolor}%
\pgfsetfillcolor{textcolor}%
\pgftext[x=8.588110in,y=0.711937in,,top]{\color{textcolor}{\sffamily\fontsize{20.000000}{24.000000}\selectfont\catcode`\^=\active\def^{\ifmmode\sp\else\^{}\fi}\catcode`\%=\active\def
\end{pgfscope}%
\begin{pgfscope}%
\definecolor{textcolor}{rgb}{0.000000,0.000000,0.000000}%
\pgfsetstrokecolor{textcolor}%
\pgfsetfillcolor{textcolor}%
\pgftext[x=7.574448in,y=0.368826in,,top]{\color{textcolor}{\sffamily\fontsize{20.000000}{24.000000}\selectfont\catcode`\^=\active\def^{\ifmmode\sp\else\^{}\fi}\catcode`\%=\active\def
\end{pgfscope}%
\begin{pgfscope}%
\definecolor{textcolor}{rgb}{0.000000,0.000000,0.000000}%
\pgfsetstrokecolor{textcolor}%
\pgfsetfillcolor{textcolor}%
\pgftext[x=5.491567in,y=2.334325in,,bottom,rotate=90.000000]{\color{textcolor}{\sffamily\fontsize{20.000000}{24.000000}\selectfont\catcode`\^=\active\def^{\ifmmode\sp\else\^{}\fi}\catcode`\%=\active\def
\end{pgfscope}%
\begin{pgfscope}%
\pgfpathrectangle{\pgfqpoint{5.547122in}{0.809159in}}{\pgfqpoint{4.054651in}{3.050333in}}%
\pgfusepath{clip}%
\pgfsetbuttcap%
\pgfsetmiterjoin%
\definecolor{currentfill}{rgb}{0.121569,0.466667,0.705882}%
\pgfsetfillcolor{currentfill}%
\pgfsetlinewidth{1.003750pt}%
\definecolor{currentstroke}{rgb}{0.121569,0.466667,0.705882}%
\pgfsetstrokecolor{currentstroke}%
\pgfsetdash{}{0pt}%
\pgfsys@defobject{currentmarker}{\pgfqpoint{-0.055556in}{-0.055556in}}{\pgfqpoint{0.055556in}{0.055556in}}{%
\pgfpathmoveto{\pgfqpoint{-0.055556in}{-0.055556in}}%
\pgfpathlineto{\pgfqpoint{0.055556in}{-0.055556in}}%
\pgfpathlineto{\pgfqpoint{0.055556in}{0.055556in}}%
\pgfpathlineto{\pgfqpoint{-0.055556in}{0.055556in}}%
\pgfpathlineto{\pgfqpoint{-0.055556in}{-0.055556in}}%
\pgfpathclose%
\pgfusepath{stroke,fill}%
}%
\begin{pgfscope}%
\pgfsys@transformshift{6.675208in}{1.315551in}%
\pgfsys@useobject{currentmarker}{}%
\end{pgfscope}%
\end{pgfscope}%
\begin{pgfscope}%
\pgfpathrectangle{\pgfqpoint{5.547122in}{0.809159in}}{\pgfqpoint{4.054651in}{3.050333in}}%
\pgfusepath{clip}%
\pgfsetbuttcap%
\pgfsetmiterjoin%
\definecolor{currentfill}{rgb}{0.121569,0.466667,0.705882}%
\pgfsetfillcolor{currentfill}%
\pgfsetlinewidth{1.003750pt}%
\definecolor{currentstroke}{rgb}{0.121569,0.466667,0.705882}%
\pgfsetstrokecolor{currentstroke}%
\pgfsetdash{}{0pt}%
\pgfsys@defobject{currentmarker}{\pgfqpoint{-0.055556in}{-0.055556in}}{\pgfqpoint{0.055556in}{0.055556in}}{%
\pgfpathmoveto{\pgfqpoint{-0.055556in}{-0.055556in}}%
\pgfpathlineto{\pgfqpoint{0.055556in}{-0.055556in}}%
\pgfpathlineto{\pgfqpoint{0.055556in}{0.055556in}}%
\pgfpathlineto{\pgfqpoint{-0.055556in}{0.055556in}}%
\pgfpathlineto{\pgfqpoint{-0.055556in}{-0.055556in}}%
\pgfpathclose%
\pgfusepath{stroke,fill}%
}%
\begin{pgfscope}%
\pgfsys@transformshift{6.765551in}{1.846633in}%
\pgfsys@useobject{currentmarker}{}%
\end{pgfscope}%
\end{pgfscope}%
\begin{pgfscope}%
\pgfpathrectangle{\pgfqpoint{5.547122in}{0.809159in}}{\pgfqpoint{4.054651in}{3.050333in}}%
\pgfusepath{clip}%
\pgfsetbuttcap%
\pgfsetmiterjoin%
\definecolor{currentfill}{rgb}{0.121569,0.466667,0.705882}%
\pgfsetfillcolor{currentfill}%
\pgfsetlinewidth{1.003750pt}%
\definecolor{currentstroke}{rgb}{0.121569,0.466667,0.705882}%
\pgfsetstrokecolor{currentstroke}%
\pgfsetdash{}{0pt}%
\pgfsys@defobject{currentmarker}{\pgfqpoint{-0.055556in}{-0.055556in}}{\pgfqpoint{0.055556in}{0.055556in}}{%
\pgfpathmoveto{\pgfqpoint{-0.055556in}{-0.055556in}}%
\pgfpathlineto{\pgfqpoint{0.055556in}{-0.055556in}}%
\pgfpathlineto{\pgfqpoint{0.055556in}{0.055556in}}%
\pgfpathlineto{\pgfqpoint{-0.055556in}{0.055556in}}%
\pgfpathlineto{\pgfqpoint{-0.055556in}{-0.055556in}}%
\pgfpathclose%
\pgfusepath{stroke,fill}%
}%
\begin{pgfscope}%
\pgfsys@transformshift{6.853919in}{2.361201in}%
\pgfsys@useobject{currentmarker}{}%
\end{pgfscope}%
\end{pgfscope}%
\begin{pgfscope}%
\pgfpathrectangle{\pgfqpoint{5.547122in}{0.809159in}}{\pgfqpoint{4.054651in}{3.050333in}}%
\pgfusepath{clip}%
\pgfsetbuttcap%
\pgfsetmiterjoin%
\definecolor{currentfill}{rgb}{0.121569,0.466667,0.705882}%
\pgfsetfillcolor{currentfill}%
\pgfsetlinewidth{1.003750pt}%
\definecolor{currentstroke}{rgb}{0.121569,0.466667,0.705882}%
\pgfsetstrokecolor{currentstroke}%
\pgfsetdash{}{0pt}%
\pgfsys@defobject{currentmarker}{\pgfqpoint{-0.055556in}{-0.055556in}}{\pgfqpoint{0.055556in}{0.055556in}}{%
\pgfpathmoveto{\pgfqpoint{-0.055556in}{-0.055556in}}%
\pgfpathlineto{\pgfqpoint{0.055556in}{-0.055556in}}%
\pgfpathlineto{\pgfqpoint{0.055556in}{0.055556in}}%
\pgfpathlineto{\pgfqpoint{-0.055556in}{0.055556in}}%
\pgfpathlineto{\pgfqpoint{-0.055556in}{-0.055556in}}%
\pgfpathclose%
\pgfusepath{stroke,fill}%
}%
\begin{pgfscope}%
\pgfsys@transformshift{6.939579in}{2.853087in}%
\pgfsys@useobject{currentmarker}{}%
\end{pgfscope}%
\end{pgfscope}%
\begin{pgfscope}%
\pgfpathrectangle{\pgfqpoint{5.547122in}{0.809159in}}{\pgfqpoint{4.054651in}{3.050333in}}%
\pgfusepath{clip}%
\pgfsetrectcap%
\pgfsetroundjoin%
\pgfsetlinewidth{1.505625pt}%
\definecolor{currentstroke}{rgb}{0.000000,0.000000,0.000000}%
\pgfsetstrokecolor{currentstroke}%
\pgfsetdash{}{0pt}%
\pgfpathmoveto{\pgfqpoint{7.021940in}{0.809159in}}%
\pgfpathlineto{\pgfqpoint{7.021940in}{3.317566in}}%
\pgfusepath{stroke}%
\end{pgfscope}%
\begin{pgfscope}%
\pgfpathrectangle{\pgfqpoint{5.547122in}{0.809159in}}{\pgfqpoint{4.054651in}{3.050333in}}%
\pgfusepath{clip}%
\pgfsetrectcap%
\pgfsetroundjoin%
\pgfsetlinewidth{1.505625pt}%
\definecolor{currentstroke}{rgb}{0.000000,0.000000,0.000000}%
\pgfsetstrokecolor{currentstroke}%
\pgfsetdash{}{0pt}%
\pgfpathmoveto{\pgfqpoint{5.537122in}{3.317566in}}%
\pgfpathlineto{\pgfqpoint{7.021940in}{3.317566in}}%
\pgfusepath{stroke}%
\end{pgfscope}%
\begin{pgfscope}%
\pgfpathrectangle{\pgfqpoint{5.547122in}{0.809159in}}{\pgfqpoint{4.054651in}{3.050333in}}%
\pgfusepath{clip}%
\pgfsetbuttcap%
\pgfsetmiterjoin%
\definecolor{currentfill}{rgb}{0.121569,0.466667,0.705882}%
\pgfsetfillcolor{currentfill}%
\pgfsetlinewidth{1.003750pt}%
\definecolor{currentstroke}{rgb}{0.121569,0.466667,0.705882}%
\pgfsetstrokecolor{currentstroke}%
\pgfsetdash{}{0pt}%
\pgfsys@defobject{currentmarker}{\pgfqpoint{-0.055556in}{-0.055556in}}{\pgfqpoint{0.055556in}{0.055556in}}{%
\pgfpathmoveto{\pgfqpoint{-0.055556in}{-0.055556in}}%
\pgfpathlineto{\pgfqpoint{0.055556in}{-0.055556in}}%
\pgfpathlineto{\pgfqpoint{0.055556in}{0.055556in}}%
\pgfpathlineto{\pgfqpoint{-0.055556in}{0.055556in}}%
\pgfpathlineto{\pgfqpoint{-0.055556in}{-0.055556in}}%
\pgfpathclose%
\pgfusepath{stroke,fill}%
}%
\begin{pgfscope}%
\pgfsys@transformshift{7.021940in}{3.317566in}%
\pgfsys@useobject{currentmarker}{}%
\end{pgfscope}%
\end{pgfscope}%
\begin{pgfscope}%
\pgfpathrectangle{\pgfqpoint{5.547122in}{0.809159in}}{\pgfqpoint{4.054651in}{3.050333in}}%
\pgfusepath{clip}%
\pgfsetbuttcap%
\pgfsetmiterjoin%
\definecolor{currentfill}{rgb}{0.121569,0.466667,0.705882}%
\pgfsetfillcolor{currentfill}%
\pgfsetlinewidth{1.003750pt}%
\definecolor{currentstroke}{rgb}{0.121569,0.466667,0.705882}%
\pgfsetstrokecolor{currentstroke}%
\pgfsetdash{}{0pt}%
\pgfsys@defobject{currentmarker}{\pgfqpoint{-0.055556in}{-0.055556in}}{\pgfqpoint{0.055556in}{0.055556in}}{%
\pgfpathmoveto{\pgfqpoint{-0.055556in}{-0.055556in}}%
\pgfpathlineto{\pgfqpoint{0.055556in}{-0.055556in}}%
\pgfpathlineto{\pgfqpoint{0.055556in}{0.055556in}}%
\pgfpathlineto{\pgfqpoint{-0.055556in}{0.055556in}}%
\pgfpathlineto{\pgfqpoint{-0.055556in}{-0.055556in}}%
\pgfpathclose%
\pgfusepath{stroke,fill}%
}%
\begin{pgfscope}%
\pgfsys@transformshift{7.100560in}{3.751446in}%
\pgfsys@useobject{currentmarker}{}%
\end{pgfscope}%
\end{pgfscope}%
\begin{pgfscope}%
\pgfpathrectangle{\pgfqpoint{5.547122in}{0.809159in}}{\pgfqpoint{4.054651in}{3.050333in}}%
\pgfusepath{clip}%
\pgfsetbuttcap%
\pgfsetmiterjoin%
\definecolor{currentfill}{rgb}{0.839216,0.152941,0.156863}%
\pgfsetfillcolor{currentfill}%
\pgfsetlinewidth{1.003750pt}%
\definecolor{currentstroke}{rgb}{0.839216,0.152941,0.156863}%
\pgfsetstrokecolor{currentstroke}%
\pgfsetdash{}{0pt}%
\pgfsys@defobject{currentmarker}{\pgfqpoint{-0.055556in}{-0.055556in}}{\pgfqpoint{0.055556in}{0.055556in}}{%
\pgfpathmoveto{\pgfqpoint{-0.055556in}{-0.055556in}}%
\pgfpathlineto{\pgfqpoint{0.055556in}{-0.055556in}}%
\pgfpathlineto{\pgfqpoint{0.055556in}{0.055556in}}%
\pgfpathlineto{\pgfqpoint{-0.055556in}{0.055556in}}%
\pgfpathlineto{\pgfqpoint{-0.055556in}{-0.055556in}}%
\pgfpathclose%
\pgfusepath{stroke,fill}%
}%
\begin{pgfscope}%
\pgfsys@transformshift{8.588110in}{1.317548in}%
\pgfsys@useobject{currentmarker}{}%
\end{pgfscope}%
\end{pgfscope}%
\begin{pgfscope}%
\pgfpathrectangle{\pgfqpoint{5.547122in}{0.809159in}}{\pgfqpoint{4.054651in}{3.050333in}}%
\pgfusepath{clip}%
\pgfsetbuttcap%
\pgfsetmiterjoin%
\definecolor{currentfill}{rgb}{0.839216,0.152941,0.156863}%
\pgfsetfillcolor{currentfill}%
\pgfsetlinewidth{1.003750pt}%
\definecolor{currentstroke}{rgb}{0.839216,0.152941,0.156863}%
\pgfsetstrokecolor{currentstroke}%
\pgfsetdash{}{0pt}%
\pgfsys@defobject{currentmarker}{\pgfqpoint{-0.055556in}{-0.055556in}}{\pgfqpoint{0.055556in}{0.055556in}}{%
\pgfpathmoveto{\pgfqpoint{-0.055556in}{-0.055556in}}%
\pgfpathlineto{\pgfqpoint{0.055556in}{-0.055556in}}%
\pgfpathlineto{\pgfqpoint{0.055556in}{0.055556in}}%
\pgfpathlineto{\pgfqpoint{-0.055556in}{0.055556in}}%
\pgfpathlineto{\pgfqpoint{-0.055556in}{-0.055556in}}%
\pgfpathclose%
\pgfusepath{stroke,fill}%
}%
\begin{pgfscope}%
\pgfsys@transformshift{8.531359in}{1.689514in}%
\pgfsys@useobject{currentmarker}{}%
\end{pgfscope}%
\end{pgfscope}%
\begin{pgfscope}%
\pgfpathrectangle{\pgfqpoint{5.547122in}{0.809159in}}{\pgfqpoint{4.054651in}{3.050333in}}%
\pgfusepath{clip}%
\pgfsetbuttcap%
\pgfsetmiterjoin%
\definecolor{currentfill}{rgb}{0.839216,0.152941,0.156863}%
\pgfsetfillcolor{currentfill}%
\pgfsetlinewidth{1.003750pt}%
\definecolor{currentstroke}{rgb}{0.839216,0.152941,0.156863}%
\pgfsetstrokecolor{currentstroke}%
\pgfsetdash{}{0pt}%
\pgfsys@defobject{currentmarker}{\pgfqpoint{-0.055556in}{-0.055556in}}{\pgfqpoint{0.055556in}{0.055556in}}{%
\pgfpathmoveto{\pgfqpoint{-0.055556in}{-0.055556in}}%
\pgfpathlineto{\pgfqpoint{0.055556in}{-0.055556in}}%
\pgfpathlineto{\pgfqpoint{0.055556in}{0.055556in}}%
\pgfpathlineto{\pgfqpoint{-0.055556in}{0.055556in}}%
\pgfpathlineto{\pgfqpoint{-0.055556in}{-0.055556in}}%
\pgfpathclose%
\pgfusepath{stroke,fill}%
}%
\begin{pgfscope}%
\pgfsys@transformshift{8.474826in}{2.058357in}%
\pgfsys@useobject{currentmarker}{}%
\end{pgfscope}%
\end{pgfscope}%
\begin{pgfscope}%
\pgfpathrectangle{\pgfqpoint{5.547122in}{0.809159in}}{\pgfqpoint{4.054651in}{3.050333in}}%
\pgfusepath{clip}%
\pgfsetbuttcap%
\pgfsetmiterjoin%
\definecolor{currentfill}{rgb}{0.839216,0.152941,0.156863}%
\pgfsetfillcolor{currentfill}%
\pgfsetlinewidth{1.003750pt}%
\definecolor{currentstroke}{rgb}{0.839216,0.152941,0.156863}%
\pgfsetstrokecolor{currentstroke}%
\pgfsetdash{}{0pt}%
\pgfsys@defobject{currentmarker}{\pgfqpoint{-0.055556in}{-0.055556in}}{\pgfqpoint{0.055556in}{0.055556in}}{%
\pgfpathmoveto{\pgfqpoint{-0.055556in}{-0.055556in}}%
\pgfpathlineto{\pgfqpoint{0.055556in}{-0.055556in}}%
\pgfpathlineto{\pgfqpoint{0.055556in}{0.055556in}}%
\pgfpathlineto{\pgfqpoint{-0.055556in}{0.055556in}}%
\pgfpathlineto{\pgfqpoint{-0.055556in}{-0.055556in}}%
\pgfpathclose%
\pgfusepath{stroke,fill}%
}%
\begin{pgfscope}%
\pgfsys@transformshift{8.418726in}{2.422013in}%
\pgfsys@useobject{currentmarker}{}%
\end{pgfscope}%
\end{pgfscope}%
\begin{pgfscope}%
\pgfpathrectangle{\pgfqpoint{5.547122in}{0.809159in}}{\pgfqpoint{4.054651in}{3.050333in}}%
\pgfusepath{clip}%
\pgfsetbuttcap%
\pgfsetmiterjoin%
\definecolor{currentfill}{rgb}{0.839216,0.152941,0.156863}%
\pgfsetfillcolor{currentfill}%
\pgfsetlinewidth{1.003750pt}%
\definecolor{currentstroke}{rgb}{0.839216,0.152941,0.156863}%
\pgfsetstrokecolor{currentstroke}%
\pgfsetdash{}{0pt}%
\pgfsys@defobject{currentmarker}{\pgfqpoint{-0.055556in}{-0.055556in}}{\pgfqpoint{0.055556in}{0.055556in}}{%
\pgfpathmoveto{\pgfqpoint{-0.055556in}{-0.055556in}}%
\pgfpathlineto{\pgfqpoint{0.055556in}{-0.055556in}}%
\pgfpathlineto{\pgfqpoint{0.055556in}{0.055556in}}%
\pgfpathlineto{\pgfqpoint{-0.055556in}{0.055556in}}%
\pgfpathlineto{\pgfqpoint{-0.055556in}{-0.055556in}}%
\pgfpathclose%
\pgfusepath{stroke,fill}%
}%
\begin{pgfscope}%
\pgfsys@transformshift{8.363261in}{2.778552in}%
\pgfsys@useobject{currentmarker}{}%
\end{pgfscope}%
\end{pgfscope}%
\begin{pgfscope}%
\pgfpathrectangle{\pgfqpoint{5.547122in}{0.809159in}}{\pgfqpoint{4.054651in}{3.050333in}}%
\pgfusepath{clip}%
\pgfsetbuttcap%
\pgfsetmiterjoin%
\definecolor{currentfill}{rgb}{0.839216,0.152941,0.156863}%
\pgfsetfillcolor{currentfill}%
\pgfsetlinewidth{1.003750pt}%
\definecolor{currentstroke}{rgb}{0.839216,0.152941,0.156863}%
\pgfsetstrokecolor{currentstroke}%
\pgfsetdash{}{0pt}%
\pgfsys@defobject{currentmarker}{\pgfqpoint{-0.055556in}{-0.055556in}}{\pgfqpoint{0.055556in}{0.055556in}}{%
\pgfpathmoveto{\pgfqpoint{-0.055556in}{-0.055556in}}%
\pgfpathlineto{\pgfqpoint{0.055556in}{-0.055556in}}%
\pgfpathlineto{\pgfqpoint{0.055556in}{0.055556in}}%
\pgfpathlineto{\pgfqpoint{-0.055556in}{0.055556in}}%
\pgfpathlineto{\pgfqpoint{-0.055556in}{-0.055556in}}%
\pgfpathclose%
\pgfusepath{stroke,fill}%
}%
\begin{pgfscope}%
\pgfsys@transformshift{8.308623in}{3.126230in}%
\pgfsys@useobject{currentmarker}{}%
\end{pgfscope}%
\end{pgfscope}%
\begin{pgfscope}%
\pgfsetrectcap%
\pgfsetmiterjoin%
\pgfsetlinewidth{1.505625pt}%
\definecolor{currentstroke}{rgb}{0.000000,0.000000,0.000000}%
\pgfsetstrokecolor{currentstroke}%
\pgfsetdash{}{0pt}%
\pgfpathmoveto{\pgfqpoint{5.547122in}{0.809159in}}%
\pgfpathlineto{\pgfqpoint{5.547122in}{3.859492in}}%
\pgfusepath{stroke}%
\end{pgfscope}%
\begin{pgfscope}%
\pgfsetrectcap%
\pgfsetmiterjoin%
\pgfsetlinewidth{1.505625pt}%
\definecolor{currentstroke}{rgb}{0.000000,0.000000,0.000000}%
\pgfsetstrokecolor{currentstroke}%
\pgfsetdash{}{0pt}%
\pgfpathmoveto{\pgfqpoint{9.601773in}{0.809159in}}%
\pgfpathlineto{\pgfqpoint{9.601773in}{3.859492in}}%
\pgfusepath{stroke}%
\end{pgfscope}%
\begin{pgfscope}%
\pgfsetrectcap%
\pgfsetmiterjoin%
\pgfsetlinewidth{1.505625pt}%
\definecolor{currentstroke}{rgb}{0.000000,0.000000,0.000000}%
\pgfsetstrokecolor{currentstroke}%
\pgfsetdash{}{0pt}%
\pgfpathmoveto{\pgfqpoint{5.547122in}{0.809159in}}%
\pgfpathlineto{\pgfqpoint{9.601773in}{0.809159in}}%
\pgfusepath{stroke}%
\end{pgfscope}%
\begin{pgfscope}%
\pgfsetrectcap%
\pgfsetmiterjoin%
\pgfsetlinewidth{1.505625pt}%
\definecolor{currentstroke}{rgb}{0.000000,0.000000,0.000000}%
\pgfsetstrokecolor{currentstroke}%
\pgfsetdash{}{0pt}%
\pgfpathmoveto{\pgfqpoint{5.547122in}{3.859492in}}%
\pgfpathlineto{\pgfqpoint{9.601773in}{3.859492in}}%
\pgfusepath{stroke}%
\end{pgfscope}%
\begin{pgfscope}%
\definecolor{textcolor}{rgb}{0.000000,0.000000,0.000000}%
\pgfsetstrokecolor{textcolor}%
\pgfsetfillcolor{textcolor}%
\pgftext[x=7.099378in,y=3.283673in,left,base]{\color{textcolor}{\sffamily\fontsize{26.000000}{31.200000}\selectfont\catcode`\^=\active\def^{\ifmmode\sp\else\^{}\fi}\catcode`\%=\active\def
\end{pgfscope}%
\begin{pgfscope}%
\definecolor{textcolor}{rgb}{0.000000,0.000000,0.000000}%
\pgfsetstrokecolor{textcolor}%
\pgfsetfillcolor{textcolor}%
\pgftext[x=7.574448in,y=3.998381in,,base]{\color{textcolor}{\sffamily\fontsize{26.000000}{31.200000}\selectfont\catcode`\^=\active\def^{\ifmmode\sp\else\^{}\fi}\catcode`\%=\active\def
\end{pgfscope}%
\begin{pgfscope}%
\pgfsetbuttcap%
\pgfsetmiterjoin%
\definecolor{currentfill}{rgb}{1.000000,1.000000,1.000000}%
\pgfsetfillcolor{currentfill}%
\pgfsetlinewidth{0.000000pt}%
\definecolor{currentstroke}{rgb}{0.000000,0.000000,0.000000}%
\pgfsetstrokecolor{currentstroke}%
\pgfsetstrokeopacity{0.000000}%
\pgfsetdash{}{0pt}%
\pgfpathmoveto{\pgfqpoint{10.669864in}{0.809159in}}%
\pgfpathlineto{\pgfqpoint{13.788826in}{0.809159in}}%
\pgfpathlineto{\pgfqpoint{13.788826in}{3.859492in}}%
\pgfpathlineto{\pgfqpoint{10.669864in}{3.859492in}}%
\pgfpathlineto{\pgfqpoint{10.669864in}{0.809159in}}%
\pgfpathclose%
\pgfusepath{fill}%
\end{pgfscope}%
\begin{pgfscope}%
\pgfpathrectangle{\pgfqpoint{10.669864in}{0.809159in}}{\pgfqpoint{3.118962in}{3.050333in}}%
\pgfusepath{clip}%
\pgfsetrectcap%
\pgfsetroundjoin%
\pgfsetlinewidth{0.803000pt}%
\definecolor{currentstroke}{rgb}{0.690196,0.690196,0.690196}%
\pgfsetstrokecolor{currentstroke}%
\pgfsetdash{}{0pt}%
\pgfpathmoveto{\pgfqpoint{11.449604in}{0.809159in}}%
\pgfpathlineto{\pgfqpoint{11.449604in}{3.859492in}}%
\pgfusepath{stroke}%
\end{pgfscope}%
\begin{pgfscope}%
\pgfsetbuttcap%
\pgfsetroundjoin%
\definecolor{currentfill}{rgb}{0.000000,0.000000,0.000000}%
\pgfsetfillcolor{currentfill}%
\pgfsetlinewidth{0.803000pt}%
\definecolor{currentstroke}{rgb}{0.000000,0.000000,0.000000}%
\pgfsetstrokecolor{currentstroke}%
\pgfsetdash{}{0pt}%
\pgfsys@defobject{currentmarker}{\pgfqpoint{0.000000in}{-0.048611in}}{\pgfqpoint{0.000000in}{0.000000in}}{%
\pgfpathmoveto{\pgfqpoint{0.000000in}{0.000000in}}%
\pgfpathlineto{\pgfqpoint{0.000000in}{-0.048611in}}%
\pgfusepath{stroke,fill}%
}%
\begin{pgfscope}%
\pgfsys@transformshift{11.449604in}{0.809159in}%
\pgfsys@useobject{currentmarker}{}%
\end{pgfscope}%
\end{pgfscope}%
\begin{pgfscope}%
\definecolor{textcolor}{rgb}{0.000000,0.000000,0.000000}%
\pgfsetstrokecolor{textcolor}%
\pgfsetfillcolor{textcolor}%
\pgftext[x=11.449604in,y=0.711937in,,top]{\color{textcolor}{\sffamily\fontsize{20.000000}{24.000000}\selectfont\catcode`\^=\active\def^{\ifmmode\sp\else\^{}\fi}\catcode`\%=\active\def
\end{pgfscope}%
\begin{pgfscope}%
\pgfpathrectangle{\pgfqpoint{10.669864in}{0.809159in}}{\pgfqpoint{3.118962in}{3.050333in}}%
\pgfusepath{clip}%
\pgfsetrectcap%
\pgfsetroundjoin%
\pgfsetlinewidth{0.803000pt}%
\definecolor{currentstroke}{rgb}{0.690196,0.690196,0.690196}%
\pgfsetstrokecolor{currentstroke}%
\pgfsetdash{}{0pt}%
\pgfpathmoveto{\pgfqpoint{13.009085in}{0.809159in}}%
\pgfpathlineto{\pgfqpoint{13.009085in}{3.859492in}}%
\pgfusepath{stroke}%
\end{pgfscope}%
\begin{pgfscope}%
\pgfsetbuttcap%
\pgfsetroundjoin%
\definecolor{currentfill}{rgb}{0.000000,0.000000,0.000000}%
\pgfsetfillcolor{currentfill}%
\pgfsetlinewidth{0.803000pt}%
\definecolor{currentstroke}{rgb}{0.000000,0.000000,0.000000}%
\pgfsetstrokecolor{currentstroke}%
\pgfsetdash{}{0pt}%
\pgfsys@defobject{currentmarker}{\pgfqpoint{0.000000in}{-0.048611in}}{\pgfqpoint{0.000000in}{0.000000in}}{%
\pgfpathmoveto{\pgfqpoint{0.000000in}{0.000000in}}%
\pgfpathlineto{\pgfqpoint{0.000000in}{-0.048611in}}%
\pgfusepath{stroke,fill}%
}%
\begin{pgfscope}%
\pgfsys@transformshift{13.009085in}{0.809159in}%
\pgfsys@useobject{currentmarker}{}%
\end{pgfscope}%
\end{pgfscope}%
\begin{pgfscope}%
\definecolor{textcolor}{rgb}{0.000000,0.000000,0.000000}%
\pgfsetstrokecolor{textcolor}%
\pgfsetfillcolor{textcolor}%
\pgftext[x=13.009085in,y=0.711937in,,top]{\color{textcolor}{\sffamily\fontsize{20.000000}{24.000000}\selectfont\catcode`\^=\active\def^{\ifmmode\sp\else\^{}\fi}\catcode`\%=\active\def
\end{pgfscope}%
\begin{pgfscope}%
\definecolor{textcolor}{rgb}{0.000000,0.000000,0.000000}%
\pgfsetstrokecolor{textcolor}%
\pgfsetfillcolor{textcolor}%
\pgftext[x=12.229345in,y=0.368826in,,top]{\color{textcolor}{\sffamily\fontsize{20.000000}{24.000000}\selectfont\catcode`\^=\active\def^{\ifmmode\sp\else\^{}\fi}\catcode`\%=\active\def
\end{pgfscope}%
\begin{pgfscope}%
\pgfpathrectangle{\pgfqpoint{10.669864in}{0.809159in}}{\pgfqpoint{3.118962in}{3.050333in}}%
\pgfusepath{clip}%
\pgfsetrectcap%
\pgfsetroundjoin%
\pgfsetlinewidth{0.803000pt}%
\definecolor{currentstroke}{rgb}{0.690196,0.690196,0.690196}%
\pgfsetstrokecolor{currentstroke}%
\pgfsetdash{}{0pt}%
\pgfpathmoveto{\pgfqpoint{10.669864in}{1.571742in}}%
\pgfpathlineto{\pgfqpoint{13.788826in}{1.571742in}}%
\pgfusepath{stroke}%
\end{pgfscope}%
\begin{pgfscope}%
\pgfsetbuttcap%
\pgfsetroundjoin%
\definecolor{currentfill}{rgb}{0.000000,0.000000,0.000000}%
\pgfsetfillcolor{currentfill}%
\pgfsetlinewidth{0.803000pt}%
\definecolor{currentstroke}{rgb}{0.000000,0.000000,0.000000}%
\pgfsetstrokecolor{currentstroke}%
\pgfsetdash{}{0pt}%
\pgfsys@defobject{currentmarker}{\pgfqpoint{-0.048611in}{0.000000in}}{\pgfqpoint{-0.000000in}{0.000000in}}{%
\pgfpathmoveto{\pgfqpoint{-0.000000in}{0.000000in}}%
\pgfpathlineto{\pgfqpoint{-0.048611in}{0.000000in}}%
\pgfusepath{stroke,fill}%
}%
\begin{pgfscope}%
\pgfsys@transformshift{10.669864in}{1.571742in}%
\pgfsys@useobject{currentmarker}{}%
\end{pgfscope}%
\end{pgfscope}%
\begin{pgfscope}%
\definecolor{textcolor}{rgb}{0.000000,0.000000,0.000000}%
\pgfsetstrokecolor{textcolor}%
\pgfsetfillcolor{textcolor}%
\pgftext[x=10.440534in, y=1.466219in, left, base]{\color{textcolor}{\sffamily\fontsize{20.000000}{24.000000}\selectfont\catcode`\^=\active\def^{\ifmmode\sp\else\^{}\fi}\catcode`\%=\active\def
\end{pgfscope}%
\begin{pgfscope}%
\pgfpathrectangle{\pgfqpoint{10.669864in}{0.809159in}}{\pgfqpoint{3.118962in}{3.050333in}}%
\pgfusepath{clip}%
\pgfsetrectcap%
\pgfsetroundjoin%
\pgfsetlinewidth{0.803000pt}%
\definecolor{currentstroke}{rgb}{0.690196,0.690196,0.690196}%
\pgfsetstrokecolor{currentstroke}%
\pgfsetdash{}{0pt}%
\pgfpathmoveto{\pgfqpoint{10.669864in}{3.096909in}}%
\pgfpathlineto{\pgfqpoint{13.788826in}{3.096909in}}%
\pgfusepath{stroke}%
\end{pgfscope}%
\begin{pgfscope}%
\pgfsetbuttcap%
\pgfsetroundjoin%
\definecolor{currentfill}{rgb}{0.000000,0.000000,0.000000}%
\pgfsetfillcolor{currentfill}%
\pgfsetlinewidth{0.803000pt}%
\definecolor{currentstroke}{rgb}{0.000000,0.000000,0.000000}%
\pgfsetstrokecolor{currentstroke}%
\pgfsetdash{}{0pt}%
\pgfsys@defobject{currentmarker}{\pgfqpoint{-0.048611in}{0.000000in}}{\pgfqpoint{-0.000000in}{0.000000in}}{%
\pgfpathmoveto{\pgfqpoint{-0.000000in}{0.000000in}}%
\pgfpathlineto{\pgfqpoint{-0.048611in}{0.000000in}}%
\pgfusepath{stroke,fill}%
}%
\begin{pgfscope}%
\pgfsys@transformshift{10.669864in}{3.096909in}%
\pgfsys@useobject{currentmarker}{}%
\end{pgfscope}%
\end{pgfscope}%
\begin{pgfscope}%
\definecolor{textcolor}{rgb}{0.000000,0.000000,0.000000}%
\pgfsetstrokecolor{textcolor}%
\pgfsetfillcolor{textcolor}%
\pgftext[x=10.138095in, y=2.989018in, left, base]{\color{textcolor}{\sffamily\fontsize{20.000000}{24.000000}\selectfont\catcode`\^=\active\def^{\ifmmode\sp\else\^{}\fi}\catcode`\%=\active\def
\end{pgfscope}%
\begin{pgfscope}%
\definecolor{textcolor}{rgb}{0.000000,0.000000,0.000000}%
\pgfsetstrokecolor{textcolor}%
\pgfsetfillcolor{textcolor}%
\pgftext[x=10.082540in,y=2.334325in,,bottom,rotate=90.000000]{\color{textcolor}{\sffamily\fontsize{20.000000}{24.000000}\selectfont\catcode`\^=\active\def^{\ifmmode\sp\else\^{}\fi}\catcode`\%=\active\def
\end{pgfscope}%
\begin{pgfscope}%
\pgfpathrectangle{\pgfqpoint{10.669864in}{0.809159in}}{\pgfqpoint{3.118962in}{3.050333in}}%
\pgfusepath{clip}%
\pgfsetbuttcap%
\pgfsetbeveljoin%
\definecolor{currentfill}{rgb}{0.000000,0.733333,0.000000}%
\pgfsetfillcolor{currentfill}%
\pgfsetlinewidth{1.003750pt}%
\definecolor{currentstroke}{rgb}{0.000000,0.733333,0.000000}%
\pgfsetstrokecolor{currentstroke}%
\pgfsetdash{}{0pt}%
\pgfsys@defobject{currentmarker}{\pgfqpoint{-0.132091in}{-0.112363in}}{\pgfqpoint{0.132091in}{0.138889in}}{%
\pgfpathmoveto{\pgfqpoint{0.000000in}{0.138889in}}%
\pgfpathlineto{\pgfqpoint{-0.031182in}{0.042919in}}%
\pgfpathlineto{\pgfqpoint{-0.132091in}{0.042919in}}%
\pgfpathlineto{\pgfqpoint{-0.050454in}{-0.016394in}}%
\pgfpathlineto{\pgfqpoint{-0.081637in}{-0.112363in}}%
\pgfpathlineto{\pgfqpoint{-0.000000in}{-0.053051in}}%
\pgfpathlineto{\pgfqpoint{0.081637in}{-0.112363in}}%
\pgfpathlineto{\pgfqpoint{0.050454in}{-0.016394in}}%
\pgfpathlineto{\pgfqpoint{0.132091in}{0.042919in}}%
\pgfpathlineto{\pgfqpoint{0.031182in}{0.042919in}}%
\pgfpathlineto{\pgfqpoint{0.000000in}{0.138889in}}%
\pgfpathclose%
\pgfusepath{stroke,fill}%
}%
\begin{pgfscope}%
\pgfsys@transformshift{11.469460in}{1.989251in}%
\pgfsys@useobject{currentmarker}{}%
\end{pgfscope}%
\end{pgfscope}%
\begin{pgfscope}%
\pgfsetrectcap%
\pgfsetmiterjoin%
\pgfsetlinewidth{1.505625pt}%
\definecolor{currentstroke}{rgb}{0.000000,0.000000,0.000000}%
\pgfsetstrokecolor{currentstroke}%
\pgfsetdash{}{0pt}%
\pgfpathmoveto{\pgfqpoint{10.669864in}{0.809159in}}%
\pgfpathlineto{\pgfqpoint{10.669864in}{3.859492in}}%
\pgfusepath{stroke}%
\end{pgfscope}%
\begin{pgfscope}%
\pgfsetrectcap%
\pgfsetmiterjoin%
\pgfsetlinewidth{1.505625pt}%
\definecolor{currentstroke}{rgb}{0.000000,0.000000,0.000000}%
\pgfsetstrokecolor{currentstroke}%
\pgfsetdash{}{0pt}%
\pgfpathmoveto{\pgfqpoint{13.788826in}{0.809159in}}%
\pgfpathlineto{\pgfqpoint{13.788826in}{3.859492in}}%
\pgfusepath{stroke}%
\end{pgfscope}%
\begin{pgfscope}%
\pgfsetrectcap%
\pgfsetmiterjoin%
\pgfsetlinewidth{1.505625pt}%
\definecolor{currentstroke}{rgb}{0.000000,0.000000,0.000000}%
\pgfsetstrokecolor{currentstroke}%
\pgfsetdash{}{0pt}%
\pgfpathmoveto{\pgfqpoint{10.669864in}{0.809159in}}%
\pgfpathlineto{\pgfqpoint{13.788826in}{0.809159in}}%
\pgfusepath{stroke}%
\end{pgfscope}%
\begin{pgfscope}%
\pgfsetrectcap%
\pgfsetmiterjoin%
\pgfsetlinewidth{1.505625pt}%
\definecolor{currentstroke}{rgb}{0.000000,0.000000,0.000000}%
\pgfsetstrokecolor{currentstroke}%
\pgfsetdash{}{0pt}%
\pgfpathmoveto{\pgfqpoint{10.669864in}{3.859492in}}%
\pgfpathlineto{\pgfqpoint{13.788826in}{3.859492in}}%
\pgfusepath{stroke}%
\end{pgfscope}%
\begin{pgfscope}%
\definecolor{textcolor}{rgb}{0.000000,0.000000,0.000000}%
\pgfsetstrokecolor{textcolor}%
\pgfsetfillcolor{textcolor}%
\pgftext[x=12.229345in,y=3.998381in,,base]{\color{textcolor}{\sffamily\fontsize{26.000000}{31.200000}\selectfont\catcode`\^=\active\def^{\ifmmode\sp\else\^{}\fi}\catcode`\%=\active\def
\end{pgfscope}%
\end{pgfpicture}%
\makeatother%
\endgroup%

%% file: figures/perspective/hough_transform.pgf
\begingroup%
\makeatletter%
\begin{pgfpicture}%
\pgfpathrectangle{\pgfpointorigin}{\pgfqpoint{9.462220in}{2.587482in}}%
\pgfusepath{use as bounding box, clip}%
\begin{pgfscope}%
\pgfsetbuttcap%
\pgfsetmiterjoin%
\definecolor{currentfill}{rgb}{1.000000,1.000000,1.000000}%
\pgfsetfillcolor{currentfill}%
\pgfsetlinewidth{0.000000pt}%
\definecolor{currentstroke}{rgb}{1.000000,1.000000,1.000000}%
\pgfsetstrokecolor{currentstroke}%
\pgfsetdash{}{0pt}%
\pgfpathmoveto{\pgfqpoint{0.000000in}{0.000000in}}%
\pgfpathlineto{\pgfqpoint{9.462220in}{0.000000in}}%
\pgfpathlineto{\pgfqpoint{9.462220in}{2.587482in}}%
\pgfpathlineto{\pgfqpoint{0.000000in}{2.587482in}}%
\pgfpathlineto{\pgfqpoint{0.000000in}{0.000000in}}%
\pgfpathclose%
\pgfusepath{fill}%
\end{pgfscope}%
\begin{pgfscope}%
\pgfsetbuttcap%
\pgfsetmiterjoin%
\definecolor{currentfill}{rgb}{1.000000,1.000000,1.000000}%
\pgfsetfillcolor{currentfill}%
\pgfsetlinewidth{0.000000pt}%
\definecolor{currentstroke}{rgb}{0.000000,0.000000,0.000000}%
\pgfsetstrokecolor{currentstroke}%
\pgfsetstrokeopacity{0.000000}%
\pgfsetdash{}{0pt}%
\pgfpathmoveto{\pgfqpoint{0.189968in}{0.346981in}}%
\pgfpathlineto{\pgfqpoint{2.925714in}{0.346981in}}%
\pgfpathlineto{\pgfqpoint{2.925714in}{2.556622in}}%
\pgfpathlineto{\pgfqpoint{0.189968in}{2.556622in}}%
\pgfpathlineto{\pgfqpoint{0.189968in}{0.346981in}}%
\pgfpathclose%
\pgfusepath{fill}%
\end{pgfscope}%
\begin{pgfscope}%
\pgfpathrectangle{\pgfqpoint{0.189968in}{0.346981in}}{\pgfqpoint{2.735746in}{2.209641in}}%
\pgfusepath{clip}%
\pgfsys@transformshift{0.189968in}{0.346981in}%
\pgftext[left,bottom]{\includegraphics[interpolate=true,width=2.737500in,height=2.210000in]{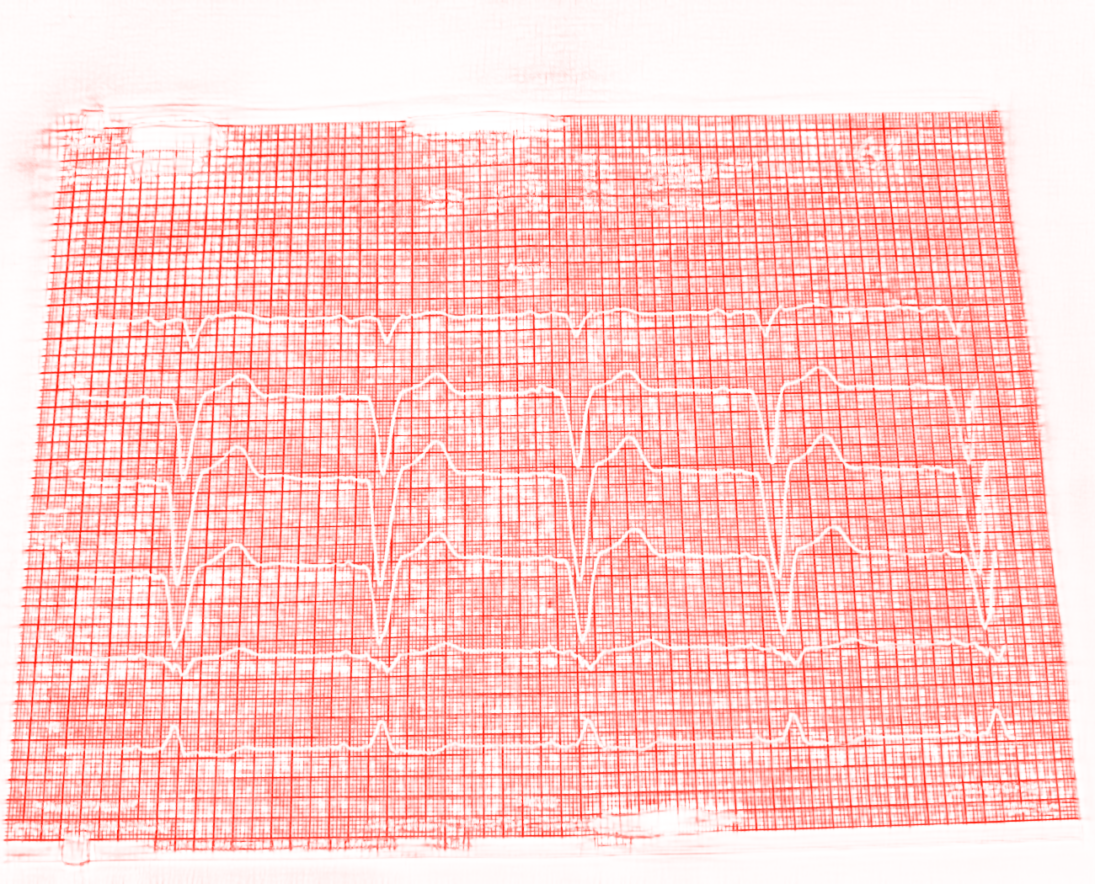}}%
\end{pgfscope}%
\begin{pgfscope}%
\definecolor{textcolor}{rgb}{0.000000,0.000000,0.000000}%
\pgfsetstrokecolor{textcolor}%
\pgfsetfillcolor{textcolor}%
\pgftext[x=1.557841in,y=0.170210in,,top]{\color{textcolor}{\sffamily\fontsize{10.000000}{12.000000}\selectfont\catcode`\^=\active\def^{\ifmmode\sp\else\^{}\fi}\catcode`\%=\active\def
\end{pgfscope}%
\begin{pgfscope}%
\definecolor{textcolor}{rgb}{0.000000,0.000000,0.000000}%
\pgfsetstrokecolor{textcolor}%
\pgfsetfillcolor{textcolor}%
\pgftext[x=0.134413in,y=1.451802in,,bottom,rotate=90.000000]{\color{textcolor}{\sffamily\fontsize{10.000000}{12.000000}\selectfont\catcode`\^=\active\def^{\ifmmode\sp\else\^{}\fi}\catcode`\%=\active\def
\end{pgfscope}%
\begin{pgfscope}%
\pgfsetrectcap%
\pgfsetmiterjoin%
\pgfsetlinewidth{0.501875pt}%
\definecolor{currentstroke}{rgb}{0.501961,0.501961,0.501961}%
\pgfsetstrokecolor{currentstroke}%
\pgfsetdash{}{0pt}%
\pgfpathmoveto{\pgfqpoint{0.189968in}{0.346981in}}%
\pgfpathlineto{\pgfqpoint{0.189968in}{2.556622in}}%
\pgfusepath{stroke}%
\end{pgfscope}%
\begin{pgfscope}%
\pgfsetrectcap%
\pgfsetmiterjoin%
\pgfsetlinewidth{0.501875pt}%
\definecolor{currentstroke}{rgb}{0.501961,0.501961,0.501961}%
\pgfsetstrokecolor{currentstroke}%
\pgfsetdash{}{0pt}%
\pgfpathmoveto{\pgfqpoint{2.925714in}{0.346981in}}%
\pgfpathlineto{\pgfqpoint{2.925714in}{2.556622in}}%
\pgfusepath{stroke}%
\end{pgfscope}%
\begin{pgfscope}%
\pgfsetrectcap%
\pgfsetmiterjoin%
\pgfsetlinewidth{0.501875pt}%
\definecolor{currentstroke}{rgb}{0.501961,0.501961,0.501961}%
\pgfsetstrokecolor{currentstroke}%
\pgfsetdash{}{0pt}%
\pgfpathmoveto{\pgfqpoint{0.189968in}{0.346981in}}%
\pgfpathlineto{\pgfqpoint{2.925714in}{0.346981in}}%
\pgfusepath{stroke}%
\end{pgfscope}%
\begin{pgfscope}%
\pgfsetrectcap%
\pgfsetmiterjoin%
\pgfsetlinewidth{0.501875pt}%
\definecolor{currentstroke}{rgb}{0.501961,0.501961,0.501961}%
\pgfsetstrokecolor{currentstroke}%
\pgfsetdash{}{0pt}%
\pgfpathmoveto{\pgfqpoint{0.189968in}{2.556622in}}%
\pgfpathlineto{\pgfqpoint{2.925714in}{2.556622in}}%
\pgfusepath{stroke}%
\end{pgfscope}%
\begin{pgfscope}%
\pgfsetbuttcap%
\pgfsetmiterjoin%
\definecolor{currentfill}{rgb}{1.000000,1.000000,1.000000}%
\pgfsetfillcolor{currentfill}%
\pgfsetlinewidth{0.000000pt}%
\definecolor{currentstroke}{rgb}{0.000000,0.000000,0.000000}%
\pgfsetstrokecolor{currentstroke}%
\pgfsetstrokeopacity{0.000000}%
\pgfsetdash{}{0pt}%
\pgfpathmoveto{\pgfqpoint{3.574318in}{0.351214in}}%
\pgfpathlineto{\pgfqpoint{6.310064in}{0.351214in}}%
\pgfpathlineto{\pgfqpoint{6.310064in}{2.552389in}}%
\pgfpathlineto{\pgfqpoint{3.574318in}{2.552389in}}%
\pgfpathlineto{\pgfqpoint{3.574318in}{0.351214in}}%
\pgfpathclose%
\pgfusepath{fill}%
\end{pgfscope}%
\begin{pgfscope}%
\pgfpathrectangle{\pgfqpoint{3.574318in}{0.351214in}}{\pgfqpoint{2.735746in}{2.201175in}}%
\pgfusepath{clip}%
\pgfsys@transformshift{3.574318in}{0.351214in}%
\pgftext[left,bottom]{\includegraphics[interpolate=true,width=2.737500in,height=2.202500in]{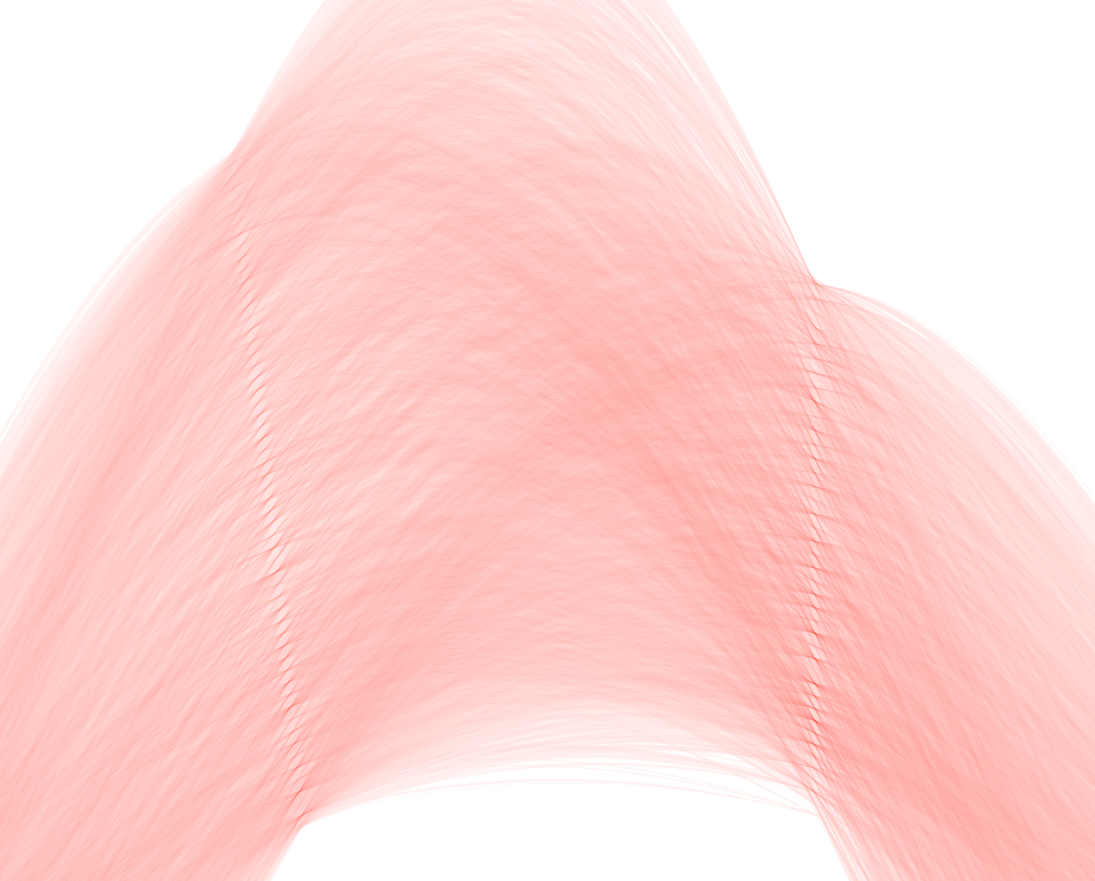}}%
\end{pgfscope}%
\begin{pgfscope}%
\pgfpathrectangle{\pgfqpoint{3.574318in}{0.351214in}}{\pgfqpoint{2.735746in}{2.201175in}}%
\pgfusepath{clip}%
\pgfsetbuttcap%
\pgfsetmiterjoin%
\pgfsetlinewidth{0.501875pt}%
\definecolor{currentstroke}{rgb}{0.501961,0.501961,0.501961}%
\pgfsetstrokecolor{currentstroke}%
\pgfsetdash{}{0pt}%
\pgfpathmoveto{\pgfqpoint{5.513710in}{1.578369in}}%
\pgfpathlineto{\pgfqpoint{5.670937in}{1.578369in}}%
\pgfpathlineto{\pgfqpoint{5.670937in}{1.704151in}}%
\pgfpathlineto{\pgfqpoint{5.513710in}{1.704151in}}%
\pgfpathlineto{\pgfqpoint{5.513710in}{1.578369in}}%
\pgfpathclose%
\pgfusepath{stroke}%
\end{pgfscope}%
\begin{pgfscope}%
\pgfpathrectangle{\pgfqpoint{3.574318in}{0.351214in}}{\pgfqpoint{2.735746in}{2.201175in}}%
\pgfusepath{clip}%
\pgfsetbuttcap%
\pgfsetmiterjoin%
\pgfsetlinewidth{0.501875pt}%
\definecolor{currentstroke}{rgb}{0.500000,0.500000,0.500000}%
\pgfsetstrokecolor{currentstroke}%
\pgfsetdash{}{0pt}%
\pgfpathmoveto{\pgfqpoint{5.513710in}{1.578369in}}%
\pgfpathlineto{\pgfqpoint{5.670937in}{1.578369in}}%
\pgfpathlineto{\pgfqpoint{5.670937in}{1.704151in}}%
\pgfpathlineto{\pgfqpoint{5.513710in}{1.704151in}}%
\pgfpathlineto{\pgfqpoint{5.513710in}{1.578369in}}%
\pgfpathclose%
\pgfusepath{stroke}%
\end{pgfscope}%
\begin{pgfscope}%
\pgfpathrectangle{\pgfqpoint{3.574318in}{0.351214in}}{\pgfqpoint{2.735746in}{2.201175in}}%
\pgfusepath{clip}%
\pgfsetrectcap%
\pgfsetroundjoin%
\pgfsetlinewidth{0.803000pt}%
\definecolor{currentstroke}{rgb}{0.501961,0.501961,0.501961}%
\pgfsetstrokecolor{currentstroke}%
\pgfsetstrokeopacity{0.500000}%
\pgfsetdash{}{0pt}%
\pgfpathmoveto{\pgfqpoint{4.257468in}{0.351214in}}%
\pgfpathlineto{\pgfqpoint{4.257468in}{2.552389in}}%
\pgfusepath{stroke}%
\end{pgfscope}%
\begin{pgfscope}%
\pgfsetbuttcap%
\pgfsetroundjoin%
\definecolor{currentfill}{rgb}{0.000000,0.000000,0.000000}%
\pgfsetfillcolor{currentfill}%
\pgfsetlinewidth{0.803000pt}%
\definecolor{currentstroke}{rgb}{0.000000,0.000000,0.000000}%
\pgfsetstrokecolor{currentstroke}%
\pgfsetdash{}{0pt}%
\pgfsys@defobject{currentmarker}{\pgfqpoint{0.000000in}{-0.048611in}}{\pgfqpoint{0.000000in}{0.000000in}}{%
\pgfpathmoveto{\pgfqpoint{0.000000in}{0.000000in}}%
\pgfpathlineto{\pgfqpoint{0.000000in}{-0.048611in}}%
\pgfusepath{stroke,fill}%
}%
\begin{pgfscope}%
\pgfsys@transformshift{4.257468in}{0.351214in}%
\pgfsys@useobject{currentmarker}{}%
\end{pgfscope}%
\end{pgfscope}%
\begin{pgfscope}%
\definecolor{textcolor}{rgb}{0.000000,0.000000,0.000000}%
\pgfsetstrokecolor{textcolor}%
\pgfsetfillcolor{textcolor}%
\pgftext[x=4.257468in,y=0.253992in,,top]{\color{textcolor}{\sffamily\fontsize{10.000000}{12.000000}\selectfont\catcode`\^=\active\def^{\ifmmode\sp\else\^{}\fi}\catcode`\%=\active\def
\end{pgfscope}%
\begin{pgfscope}%
\pgfpathrectangle{\pgfqpoint{3.574318in}{0.351214in}}{\pgfqpoint{2.735746in}{2.201175in}}%
\pgfusepath{clip}%
\pgfsetrectcap%
\pgfsetroundjoin%
\pgfsetlinewidth{0.803000pt}%
\definecolor{currentstroke}{rgb}{0.501961,0.501961,0.501961}%
\pgfsetstrokecolor{currentstroke}%
\pgfsetstrokeopacity{0.500000}%
\pgfsetdash{}{0pt}%
\pgfpathmoveto{\pgfqpoint{5.625341in}{0.351214in}}%
\pgfpathlineto{\pgfqpoint{5.625341in}{2.552389in}}%
\pgfusepath{stroke}%
\end{pgfscope}%
\begin{pgfscope}%
\pgfsetbuttcap%
\pgfsetroundjoin%
\definecolor{currentfill}{rgb}{0.000000,0.000000,0.000000}%
\pgfsetfillcolor{currentfill}%
\pgfsetlinewidth{0.803000pt}%
\definecolor{currentstroke}{rgb}{0.000000,0.000000,0.000000}%
\pgfsetstrokecolor{currentstroke}%
\pgfsetdash{}{0pt}%
\pgfsys@defobject{currentmarker}{\pgfqpoint{0.000000in}{-0.048611in}}{\pgfqpoint{0.000000in}{0.000000in}}{%
\pgfpathmoveto{\pgfqpoint{0.000000in}{0.000000in}}%
\pgfpathlineto{\pgfqpoint{0.000000in}{-0.048611in}}%
\pgfusepath{stroke,fill}%
}%
\begin{pgfscope}%
\pgfsys@transformshift{5.625341in}{0.351214in}%
\pgfsys@useobject{currentmarker}{}%
\end{pgfscope}%
\end{pgfscope}%
\begin{pgfscope}%
\definecolor{textcolor}{rgb}{0.000000,0.000000,0.000000}%
\pgfsetstrokecolor{textcolor}%
\pgfsetfillcolor{textcolor}%
\pgftext[x=5.625341in,y=0.253992in,,top]{\color{textcolor}{\sffamily\fontsize{10.000000}{12.000000}\selectfont\catcode`\^=\active\def^{\ifmmode\sp\else\^{}\fi}\catcode`\%=\active\def
\end{pgfscope}%
\begin{pgfscope}%
\definecolor{textcolor}{rgb}{0.000000,0.000000,0.000000}%
\pgfsetstrokecolor{textcolor}%
\pgfsetfillcolor{textcolor}%
\pgftext[x=4.942191in,y=0.175120in,,top]{\color{textcolor}{\sffamily\fontsize{10.000000}{12.000000}\selectfont\catcode`\^=\active\def^{\ifmmode\sp\else\^{}\fi}\catcode`\%=\active\def
\end{pgfscope}%
\begin{pgfscope}%
\definecolor{textcolor}{rgb}{0.000000,0.000000,0.000000}%
\pgfsetstrokecolor{textcolor}%
\pgfsetfillcolor{textcolor}%
\pgftext[x=3.518762in,y=1.451802in,,bottom,rotate=90.000000]{\color{textcolor}{\sffamily\fontsize{10.000000}{12.000000}\selectfont\catcode`\^=\active\def^{\ifmmode\sp\else\^{}\fi}\catcode`\%=\active\def
\end{pgfscope}%
\begin{pgfscope}%
\pgfsetrectcap%
\pgfsetmiterjoin%
\pgfsetlinewidth{0.501875pt}%
\definecolor{currentstroke}{rgb}{0.501961,0.501961,0.501961}%
\pgfsetstrokecolor{currentstroke}%
\pgfsetdash{}{0pt}%
\pgfpathmoveto{\pgfqpoint{3.574318in}{0.351214in}}%
\pgfpathlineto{\pgfqpoint{3.574318in}{2.552389in}}%
\pgfusepath{stroke}%
\end{pgfscope}%
\begin{pgfscope}%
\pgfsetrectcap%
\pgfsetmiterjoin%
\pgfsetlinewidth{0.501875pt}%
\definecolor{currentstroke}{rgb}{0.501961,0.501961,0.501961}%
\pgfsetstrokecolor{currentstroke}%
\pgfsetdash{}{0pt}%
\pgfpathmoveto{\pgfqpoint{6.310064in}{0.351214in}}%
\pgfpathlineto{\pgfqpoint{6.310064in}{2.552389in}}%
\pgfusepath{stroke}%
\end{pgfscope}%
\begin{pgfscope}%
\pgfsetrectcap%
\pgfsetmiterjoin%
\pgfsetlinewidth{0.501875pt}%
\definecolor{currentstroke}{rgb}{0.501961,0.501961,0.501961}%
\pgfsetstrokecolor{currentstroke}%
\pgfsetdash{}{0pt}%
\pgfpathmoveto{\pgfqpoint{3.574318in}{0.351214in}}%
\pgfpathlineto{\pgfqpoint{6.310064in}{0.351214in}}%
\pgfusepath{stroke}%
\end{pgfscope}%
\begin{pgfscope}%
\pgfsetrectcap%
\pgfsetmiterjoin%
\pgfsetlinewidth{0.501875pt}%
\definecolor{currentstroke}{rgb}{0.501961,0.501961,0.501961}%
\pgfsetstrokecolor{currentstroke}%
\pgfsetdash{}{0pt}%
\pgfpathmoveto{\pgfqpoint{3.574318in}{2.552389in}}%
\pgfpathlineto{\pgfqpoint{6.310064in}{2.552389in}}%
\pgfusepath{stroke}%
\end{pgfscope}%
\begin{pgfscope}%
\pgfsetbuttcap%
\pgfsetmiterjoin%
\definecolor{currentfill}{rgb}{1.000000,1.000000,1.000000}%
\pgfsetfillcolor{currentfill}%
\pgfsetlinewidth{0.000000pt}%
\definecolor{currentstroke}{rgb}{0.000000,0.000000,0.000000}%
\pgfsetstrokecolor{currentstroke}%
\pgfsetstrokeopacity{0.000000}%
\pgfsetdash{}{0pt}%
\pgfpathmoveto{\pgfqpoint{7.190860in}{0.316122in}}%
\pgfpathlineto{\pgfqpoint{9.462220in}{0.316122in}}%
\pgfpathlineto{\pgfqpoint{9.462220in}{2.587482in}}%
\pgfpathlineto{\pgfqpoint{7.190860in}{2.587482in}}%
\pgfpathlineto{\pgfqpoint{7.190860in}{0.316122in}}%
\pgfpathclose%
\pgfusepath{fill}%
\end{pgfscope}%
\begin{pgfscope}%
\pgfpathrectangle{\pgfqpoint{7.190860in}{0.316122in}}{\pgfqpoint{2.271360in}{2.271360in}}%
\pgfusepath{clip}%
\pgfsys@transformshift{7.190860in}{0.316122in}%
\pgftext[left,bottom]{\includegraphics[interpolate=true,width=2.272500in,height=2.272500in]{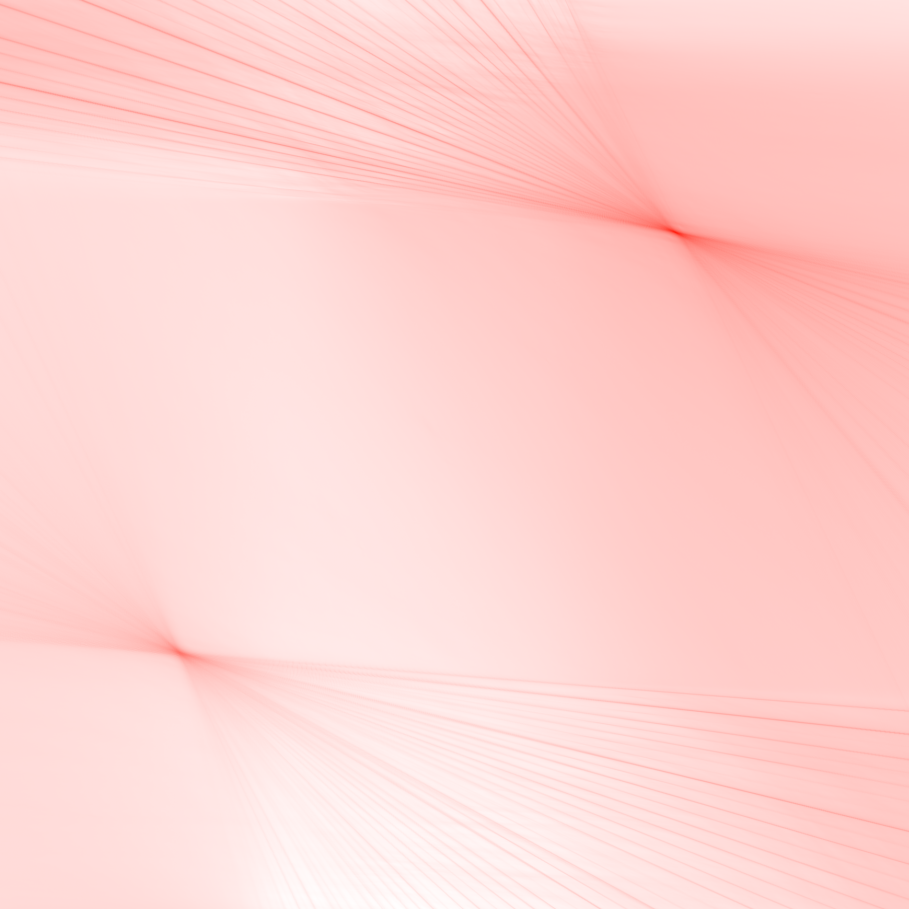}}%
\end{pgfscope}%
\begin{pgfscope}%
\pgfpathrectangle{\pgfqpoint{7.190860in}{0.316122in}}{\pgfqpoint{2.271360in}{2.271360in}}%
\pgfusepath{clip}%
\pgfsetrectcap%
\pgfsetroundjoin%
\pgfsetlinewidth{0.803000pt}%
\definecolor{currentstroke}{rgb}{0.501961,0.501961,0.501961}%
\pgfsetstrokecolor{currentstroke}%
\pgfsetstrokeopacity{0.500000}%
\pgfsetdash{}{0pt}%
\pgfpathmoveto{\pgfqpoint{7.757394in}{0.316122in}}%
\pgfpathlineto{\pgfqpoint{7.757394in}{2.587482in}}%
\pgfusepath{stroke}%
\end{pgfscope}%
\begin{pgfscope}%
\pgfsetbuttcap%
\pgfsetroundjoin%
\definecolor{currentfill}{rgb}{0.000000,0.000000,0.000000}%
\pgfsetfillcolor{currentfill}%
\pgfsetlinewidth{0.803000pt}%
\definecolor{currentstroke}{rgb}{0.000000,0.000000,0.000000}%
\pgfsetstrokecolor{currentstroke}%
\pgfsetdash{}{0pt}%
\pgfsys@defobject{currentmarker}{\pgfqpoint{0.000000in}{-0.048611in}}{\pgfqpoint{0.000000in}{0.000000in}}{%
\pgfpathmoveto{\pgfqpoint{0.000000in}{0.000000in}}%
\pgfpathlineto{\pgfqpoint{0.000000in}{-0.048611in}}%
\pgfusepath{stroke,fill}%
}%
\begin{pgfscope}%
\pgfsys@transformshift{7.757394in}{0.316122in}%
\pgfsys@useobject{currentmarker}{}%
\end{pgfscope}%
\end{pgfscope}%
\begin{pgfscope}%
\definecolor{textcolor}{rgb}{0.000000,0.000000,0.000000}%
\pgfsetstrokecolor{textcolor}%
\pgfsetfillcolor{textcolor}%
\pgftext[x=7.757394in,y=0.218899in,,top]{\color{textcolor}{\sffamily\fontsize{10.000000}{12.000000}\selectfont\catcode`\^=\active\def^{\ifmmode\sp\else\^{}\fi}\catcode`\%=\active\def
\end{pgfscope}%
\begin{pgfscope}%
\pgfpathrectangle{\pgfqpoint{7.190860in}{0.316122in}}{\pgfqpoint{2.271360in}{2.271360in}}%
\pgfusepath{clip}%
\pgfsetrectcap%
\pgfsetroundjoin%
\pgfsetlinewidth{0.803000pt}%
\definecolor{currentstroke}{rgb}{0.501961,0.501961,0.501961}%
\pgfsetstrokecolor{currentstroke}%
\pgfsetstrokeopacity{0.500000}%
\pgfsetdash{}{0pt}%
\pgfpathmoveto{\pgfqpoint{8.893074in}{0.316122in}}%
\pgfpathlineto{\pgfqpoint{8.893074in}{2.587482in}}%
\pgfusepath{stroke}%
\end{pgfscope}%
\begin{pgfscope}%
\pgfsetbuttcap%
\pgfsetroundjoin%
\definecolor{currentfill}{rgb}{0.000000,0.000000,0.000000}%
\pgfsetfillcolor{currentfill}%
\pgfsetlinewidth{0.803000pt}%
\definecolor{currentstroke}{rgb}{0.000000,0.000000,0.000000}%
\pgfsetstrokecolor{currentstroke}%
\pgfsetdash{}{0pt}%
\pgfsys@defobject{currentmarker}{\pgfqpoint{0.000000in}{-0.048611in}}{\pgfqpoint{0.000000in}{0.000000in}}{%
\pgfpathmoveto{\pgfqpoint{0.000000in}{0.000000in}}%
\pgfpathlineto{\pgfqpoint{0.000000in}{-0.048611in}}%
\pgfusepath{stroke,fill}%
}%
\begin{pgfscope}%
\pgfsys@transformshift{8.893074in}{0.316122in}%
\pgfsys@useobject{currentmarker}{}%
\end{pgfscope}%
\end{pgfscope}%
\begin{pgfscope}%
\definecolor{textcolor}{rgb}{0.000000,0.000000,0.000000}%
\pgfsetstrokecolor{textcolor}%
\pgfsetfillcolor{textcolor}%
\pgftext[x=8.893074in,y=0.218899in,,top]{\color{textcolor}{\sffamily\fontsize{10.000000}{12.000000}\selectfont\catcode`\^=\active\def^{\ifmmode\sp\else\^{}\fi}\catcode`\%=\active\def
\end{pgfscope}%
\begin{pgfscope}%
\definecolor{textcolor}{rgb}{0.000000,0.000000,0.000000}%
\pgfsetstrokecolor{textcolor}%
\pgfsetfillcolor{textcolor}%
\pgftext[x=8.326540in,y=0.134413in,,top]{\color{textcolor}{\sffamily\fontsize{10.000000}{12.000000}\selectfont\catcode`\^=\active\def^{\ifmmode\sp\else\^{}\fi}\catcode`\%=\active\def
\end{pgfscope}%
\begin{pgfscope}%
\pgfpathrectangle{\pgfqpoint{7.190860in}{0.316122in}}{\pgfqpoint{2.271360in}{2.271360in}}%
\pgfusepath{clip}%
\pgfsetrectcap%
\pgfsetroundjoin%
\pgfsetlinewidth{0.803000pt}%
\definecolor{currentstroke}{rgb}{0.501961,0.501961,0.501961}%
\pgfsetstrokecolor{currentstroke}%
\pgfsetstrokeopacity{0.500000}%
\pgfsetdash{}{0pt}%
\pgfpathmoveto{\pgfqpoint{7.190860in}{0.882656in}}%
\pgfpathlineto{\pgfqpoint{9.462220in}{0.882656in}}%
\pgfusepath{stroke}%
\end{pgfscope}%
\begin{pgfscope}%
\pgfsetbuttcap%
\pgfsetroundjoin%
\definecolor{currentfill}{rgb}{0.000000,0.000000,0.000000}%
\pgfsetfillcolor{currentfill}%
\pgfsetlinewidth{0.803000pt}%
\definecolor{currentstroke}{rgb}{0.000000,0.000000,0.000000}%
\pgfsetstrokecolor{currentstroke}%
\pgfsetdash{}{0pt}%
\pgfsys@defobject{currentmarker}{\pgfqpoint{-0.048611in}{0.000000in}}{\pgfqpoint{-0.000000in}{0.000000in}}{%
\pgfpathmoveto{\pgfqpoint{-0.000000in}{0.000000in}}%
\pgfpathlineto{\pgfqpoint{-0.048611in}{0.000000in}}%
\pgfusepath{stroke,fill}%
}%
\begin{pgfscope}%
\pgfsys@transformshift{7.190860in}{0.882656in}%
\pgfsys@useobject{currentmarker}{}%
\end{pgfscope}%
\end{pgfscope}%
\begin{pgfscope}%
\definecolor{textcolor}{rgb}{0.000000,0.000000,0.000000}%
\pgfsetstrokecolor{textcolor}%
\pgfsetfillcolor{textcolor}%
\pgftext[x=7.005272in, y=0.829895in, left, base]{\color{textcolor}{\sffamily\fontsize{10.000000}{12.000000}\selectfont\catcode`\^=\active\def^{\ifmmode\sp\else\^{}\fi}\catcode`\%=\active\def
\end{pgfscope}%
\begin{pgfscope}%
\pgfpathrectangle{\pgfqpoint{7.190860in}{0.316122in}}{\pgfqpoint{2.271360in}{2.271360in}}%
\pgfusepath{clip}%
\pgfsetrectcap%
\pgfsetroundjoin%
\pgfsetlinewidth{0.803000pt}%
\definecolor{currentstroke}{rgb}{0.501961,0.501961,0.501961}%
\pgfsetstrokecolor{currentstroke}%
\pgfsetstrokeopacity{0.500000}%
\pgfsetdash{}{0pt}%
\pgfpathmoveto{\pgfqpoint{7.190860in}{2.018336in}}%
\pgfpathlineto{\pgfqpoint{9.462220in}{2.018336in}}%
\pgfusepath{stroke}%
\end{pgfscope}%
\begin{pgfscope}%
\pgfsetbuttcap%
\pgfsetroundjoin%
\definecolor{currentfill}{rgb}{0.000000,0.000000,0.000000}%
\pgfsetfillcolor{currentfill}%
\pgfsetlinewidth{0.803000pt}%
\definecolor{currentstroke}{rgb}{0.000000,0.000000,0.000000}%
\pgfsetstrokecolor{currentstroke}%
\pgfsetdash{}{0pt}%
\pgfsys@defobject{currentmarker}{\pgfqpoint{-0.048611in}{0.000000in}}{\pgfqpoint{-0.000000in}{0.000000in}}{%
\pgfpathmoveto{\pgfqpoint{-0.000000in}{0.000000in}}%
\pgfpathlineto{\pgfqpoint{-0.048611in}{0.000000in}}%
\pgfusepath{stroke,fill}%
}%
\begin{pgfscope}%
\pgfsys@transformshift{7.190860in}{2.018336in}%
\pgfsys@useobject{currentmarker}{}%
\end{pgfscope}%
\end{pgfscope}%
\begin{pgfscope}%
\definecolor{textcolor}{rgb}{0.000000,0.000000,0.000000}%
\pgfsetstrokecolor{textcolor}%
\pgfsetfillcolor{textcolor}%
\pgftext[x=6.870595in, y=1.966253in, left, base]{\color{textcolor}{\sffamily\fontsize{10.000000}{12.000000}\selectfont\catcode`\^=\active\def^{\ifmmode\sp\else\^{}\fi}\catcode`\%=\active\def
\end{pgfscope}%
\begin{pgfscope}%
\definecolor{textcolor}{rgb}{0.000000,0.000000,0.000000}%
\pgfsetstrokecolor{textcolor}%
\pgfsetfillcolor{textcolor}%
\pgftext[x=6.815039in,y=1.451802in,,bottom,rotate=90.000000]{\color{textcolor}{\sffamily\fontsize{10.000000}{12.000000}\selectfont\catcode`\^=\active\def^{\ifmmode\sp\else\^{}\fi}\catcode`\%=\active\def
\end{pgfscope}%
\begin{pgfscope}%
\pgfsetrectcap%
\pgfsetmiterjoin%
\pgfsetlinewidth{0.501875pt}%
\definecolor{currentstroke}{rgb}{0.501961,0.501961,0.501961}%
\pgfsetstrokecolor{currentstroke}%
\pgfsetdash{}{0pt}%
\pgfpathmoveto{\pgfqpoint{7.190860in}{0.316122in}}%
\pgfpathlineto{\pgfqpoint{7.190860in}{2.587482in}}%
\pgfusepath{stroke}%
\end{pgfscope}%
\begin{pgfscope}%
\pgfsetrectcap%
\pgfsetmiterjoin%
\pgfsetlinewidth{0.501875pt}%
\definecolor{currentstroke}{rgb}{0.501961,0.501961,0.501961}%
\pgfsetstrokecolor{currentstroke}%
\pgfsetdash{}{0pt}%
\pgfpathmoveto{\pgfqpoint{9.462220in}{0.316122in}}%
\pgfpathlineto{\pgfqpoint{9.462220in}{2.587482in}}%
\pgfusepath{stroke}%
\end{pgfscope}%
\begin{pgfscope}%
\pgfsetrectcap%
\pgfsetmiterjoin%
\pgfsetlinewidth{0.501875pt}%
\definecolor{currentstroke}{rgb}{0.501961,0.501961,0.501961}%
\pgfsetstrokecolor{currentstroke}%
\pgfsetdash{}{0pt}%
\pgfpathmoveto{\pgfqpoint{7.190860in}{0.316122in}}%
\pgfpathlineto{\pgfqpoint{9.462220in}{0.316122in}}%
\pgfusepath{stroke}%
\end{pgfscope}%
\begin{pgfscope}%
\pgfsetrectcap%
\pgfsetmiterjoin%
\pgfsetlinewidth{0.501875pt}%
\definecolor{currentstroke}{rgb}{0.501961,0.501961,0.501961}%
\pgfsetstrokecolor{currentstroke}%
\pgfsetdash{}{0pt}%
\pgfpathmoveto{\pgfqpoint{7.190860in}{2.587482in}}%
\pgfpathlineto{\pgfqpoint{9.462220in}{2.587482in}}%
\pgfusepath{stroke}%
\end{pgfscope}%
\begin{pgfscope}%
\pgfsetbuttcap%
\pgfsetmiterjoin%
\definecolor{currentfill}{rgb}{1.000000,1.000000,1.000000}%
\pgfsetfillcolor{currentfill}%
\pgfsetlinewidth{0.000000pt}%
\definecolor{currentstroke}{rgb}{0.000000,0.000000,0.000000}%
\pgfsetstrokecolor{currentstroke}%
\pgfsetstrokeopacity{0.000000}%
\pgfsetdash{}{0pt}%
\pgfpathmoveto{\pgfqpoint{5.560876in}{1.944706in}}%
\pgfpathlineto{\pgfqpoint{6.268397in}{1.944706in}}%
\pgfpathlineto{\pgfqpoint{6.268397in}{2.510722in}}%
\pgfpathlineto{\pgfqpoint{5.560876in}{2.510722in}}%
\pgfpathlineto{\pgfqpoint{5.560876in}{1.944706in}}%
\pgfpathclose%
\pgfusepath{fill}%
\end{pgfscope}%
\begin{pgfscope}%
\pgfpathrectangle{\pgfqpoint{5.560876in}{1.944706in}}{\pgfqpoint{0.707521in}{0.566016in}}%
\pgfusepath{clip}%
\pgfsys@transformshift{5.560876in}{1.944706in}%
\pgftext[left,bottom]{\includegraphics[interpolate=true,width=0.710000in,height=0.567500in]{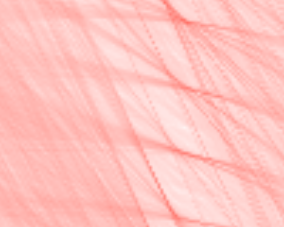}}%
\end{pgfscope}%
\begin{pgfscope}%
\pgfsetbuttcap%
\pgfsetmiterjoin%
\pgfsetlinewidth{0.501875pt}%
\definecolor{currentstroke}{rgb}{0.500000,0.500000,0.500000}%
\pgfsetstrokecolor{currentstroke}%
\pgfsetdash{}{0pt}%
\pgfpathmoveto{\pgfqpoint{5.560876in}{2.510722in}}%
\pgfpathlineto{\pgfqpoint{5.513710in}{1.704151in}}%
\pgfusepath{stroke}%
\end{pgfscope}%
\begin{pgfscope}%
\pgfsetbuttcap%
\pgfsetmiterjoin%
\pgfsetlinewidth{0.501875pt}%
\definecolor{currentstroke}{rgb}{0.500000,0.500000,0.500000}%
\pgfsetstrokecolor{currentstroke}%
\pgfsetdash{}{0pt}%
\pgfpathmoveto{\pgfqpoint{6.268397in}{1.944706in}}%
\pgfpathlineto{\pgfqpoint{5.670937in}{1.578369in}}%
\pgfusepath{stroke}%
\end{pgfscope}%
\begin{pgfscope}%
\pgfsetrectcap%
\pgfsetmiterjoin%
\pgfsetlinewidth{0.803000pt}%
\definecolor{currentstroke}{rgb}{0.501961,0.501961,0.501961}%
\pgfsetstrokecolor{currentstroke}%
\pgfsetdash{}{0pt}%
\pgfpathmoveto{\pgfqpoint{5.560876in}{1.944706in}}%
\pgfpathlineto{\pgfqpoint{5.560876in}{2.510722in}}%
\pgfusepath{stroke}%
\end{pgfscope}%
\begin{pgfscope}%
\pgfsetrectcap%
\pgfsetmiterjoin%
\pgfsetlinewidth{0.803000pt}%
\definecolor{currentstroke}{rgb}{0.501961,0.501961,0.501961}%
\pgfsetstrokecolor{currentstroke}%
\pgfsetdash{}{0pt}%
\pgfpathmoveto{\pgfqpoint{6.268397in}{1.944706in}}%
\pgfpathlineto{\pgfqpoint{6.268397in}{2.510722in}}%
\pgfusepath{stroke}%
\end{pgfscope}%
\begin{pgfscope}%
\pgfsetrectcap%
\pgfsetmiterjoin%
\pgfsetlinewidth{0.803000pt}%
\definecolor{currentstroke}{rgb}{0.501961,0.501961,0.501961}%
\pgfsetstrokecolor{currentstroke}%
\pgfsetdash{}{0pt}%
\pgfpathmoveto{\pgfqpoint{5.560876in}{1.944706in}}%
\pgfpathlineto{\pgfqpoint{6.268397in}{1.944706in}}%
\pgfusepath{stroke}%
\end{pgfscope}%
\begin{pgfscope}%
\pgfsetrectcap%
\pgfsetmiterjoin%
\pgfsetlinewidth{0.803000pt}%
\definecolor{currentstroke}{rgb}{0.501961,0.501961,0.501961}%
\pgfsetstrokecolor{currentstroke}%
\pgfsetdash{}{0pt}%
\pgfpathmoveto{\pgfqpoint{5.560876in}{2.510722in}}%
\pgfpathlineto{\pgfqpoint{6.268397in}{2.510722in}}%
\pgfusepath{stroke}%
\end{pgfscope}%
\end{pgfpicture}%
\makeatother%
\endgroup%

%% file: figures/pipeline/autocorr_grid_search.pgf
\begingroup%
\makeatletter%
\begin{pgfpicture}%
\pgfpathrectangle{\pgfpointorigin}{\pgfqpoint{5.000000in}{2.500000in}}%
\pgfusepath{use as bounding box, clip}%
\begin{pgfscope}%
\pgfsetbuttcap%
\pgfsetmiterjoin%
\definecolor{currentfill}{rgb}{1.000000,1.000000,1.000000}%
\pgfsetfillcolor{currentfill}%
\pgfsetlinewidth{0.000000pt}%
\definecolor{currentstroke}{rgb}{1.000000,1.000000,1.000000}%
\pgfsetstrokecolor{currentstroke}%
\pgfsetdash{}{0pt}%
\pgfpathmoveto{\pgfqpoint{0.000000in}{0.000000in}}%
\pgfpathlineto{\pgfqpoint{5.000000in}{0.000000in}}%
\pgfpathlineto{\pgfqpoint{5.000000in}{2.500000in}}%
\pgfpathlineto{\pgfqpoint{0.000000in}{2.500000in}}%
\pgfpathlineto{\pgfqpoint{0.000000in}{0.000000in}}%
\pgfpathclose%
\pgfusepath{fill}%
\end{pgfscope}%
\begin{pgfscope}%
\pgfsetbuttcap%
\pgfsetmiterjoin%
\definecolor{currentfill}{rgb}{1.000000,1.000000,1.000000}%
\pgfsetfillcolor{currentfill}%
\pgfsetlinewidth{0.000000pt}%
\definecolor{currentstroke}{rgb}{0.000000,0.000000,0.000000}%
\pgfsetstrokecolor{currentstroke}%
\pgfsetstrokeopacity{0.000000}%
\pgfsetdash{}{0pt}%
\pgfpathmoveto{\pgfqpoint{0.335972in}{0.582778in}}%
\pgfpathlineto{\pgfqpoint{2.425000in}{0.582778in}}%
\pgfpathlineto{\pgfqpoint{2.425000in}{2.126667in}}%
\pgfpathlineto{\pgfqpoint{0.335972in}{2.126667in}}%
\pgfpathlineto{\pgfqpoint{0.335972in}{0.582778in}}%
\pgfpathclose%
\pgfusepath{fill}%
\end{pgfscope}%
\begin{pgfscope}%
\pgfpathrectangle{\pgfqpoint{0.335972in}{0.582778in}}{\pgfqpoint{2.089028in}{1.543889in}}%
\pgfusepath{clip}%
\pgfsetrectcap%
\pgfsetroundjoin%
\pgfsetlinewidth{0.803000pt}%
\definecolor{currentstroke}{rgb}{0.690196,0.690196,0.690196}%
\pgfsetstrokecolor{currentstroke}%
\pgfsetdash{}{0pt}%
\pgfpathmoveto{\pgfqpoint{0.430928in}{0.582778in}}%
\pgfpathlineto{\pgfqpoint{0.430928in}{2.126667in}}%
\pgfusepath{stroke}%
\end{pgfscope}%
\begin{pgfscope}%
\pgfsetbuttcap%
\pgfsetroundjoin%
\definecolor{currentfill}{rgb}{0.000000,0.000000,0.000000}%
\pgfsetfillcolor{currentfill}%
\pgfsetlinewidth{0.803000pt}%
\definecolor{currentstroke}{rgb}{0.000000,0.000000,0.000000}%
\pgfsetstrokecolor{currentstroke}%
\pgfsetdash{}{0pt}%
\pgfsys@defobject{currentmarker}{\pgfqpoint{0.000000in}{-0.048611in}}{\pgfqpoint{0.000000in}{0.000000in}}{%
\pgfpathmoveto{\pgfqpoint{0.000000in}{0.000000in}}%
\pgfpathlineto{\pgfqpoint{0.000000in}{-0.048611in}}%
\pgfusepath{stroke,fill}%
}%
\begin{pgfscope}%
\pgfsys@transformshift{0.430928in}{0.582778in}%
\pgfsys@useobject{currentmarker}{}%
\end{pgfscope}%
\end{pgfscope}%
\begin{pgfscope}%
\definecolor{textcolor}{rgb}{0.000000,0.000000,0.000000}%
\pgfsetstrokecolor{textcolor}%
\pgfsetfillcolor{textcolor}%
\pgftext[x=0.430928in,y=0.485556in,,top]{\color{textcolor}{\sffamily\fontsize{10.000000}{12.000000}\selectfont\catcode`\^=\active\def^{\ifmmode\sp\else\^{}\fi}\catcode`\%=\active\def
\end{pgfscope}%
\begin{pgfscope}%
\pgfpathrectangle{\pgfqpoint{0.335972in}{0.582778in}}{\pgfqpoint{2.089028in}{1.543889in}}%
\pgfusepath{clip}%
\pgfsetrectcap%
\pgfsetroundjoin%
\pgfsetlinewidth{0.803000pt}%
\definecolor{currentstroke}{rgb}{0.690196,0.690196,0.690196}%
\pgfsetstrokecolor{currentstroke}%
\pgfsetdash{}{0pt}%
\pgfpathmoveto{\pgfqpoint{1.339596in}{0.582778in}}%
\pgfpathlineto{\pgfqpoint{1.339596in}{2.126667in}}%
\pgfusepath{stroke}%
\end{pgfscope}%
\begin{pgfscope}%
\pgfsetbuttcap%
\pgfsetroundjoin%
\definecolor{currentfill}{rgb}{0.000000,0.000000,0.000000}%
\pgfsetfillcolor{currentfill}%
\pgfsetlinewidth{0.803000pt}%
\definecolor{currentstroke}{rgb}{0.000000,0.000000,0.000000}%
\pgfsetstrokecolor{currentstroke}%
\pgfsetdash{}{0pt}%
\pgfsys@defobject{currentmarker}{\pgfqpoint{0.000000in}{-0.048611in}}{\pgfqpoint{0.000000in}{0.000000in}}{%
\pgfpathmoveto{\pgfqpoint{0.000000in}{0.000000in}}%
\pgfpathlineto{\pgfqpoint{0.000000in}{-0.048611in}}%
\pgfusepath{stroke,fill}%
}%
\begin{pgfscope}%
\pgfsys@transformshift{1.339596in}{0.582778in}%
\pgfsys@useobject{currentmarker}{}%
\end{pgfscope}%
\end{pgfscope}%
\begin{pgfscope}%
\definecolor{textcolor}{rgb}{0.000000,0.000000,0.000000}%
\pgfsetstrokecolor{textcolor}%
\pgfsetfillcolor{textcolor}%
\pgftext[x=1.339596in,y=0.485556in,,top]{\color{textcolor}{\sffamily\fontsize{10.000000}{12.000000}\selectfont\catcode`\^=\active\def^{\ifmmode\sp\else\^{}\fi}\catcode`\%=\active\def
\end{pgfscope}%
\begin{pgfscope}%
\pgfpathrectangle{\pgfqpoint{0.335972in}{0.582778in}}{\pgfqpoint{2.089028in}{1.543889in}}%
\pgfusepath{clip}%
\pgfsetrectcap%
\pgfsetroundjoin%
\pgfsetlinewidth{0.803000pt}%
\definecolor{currentstroke}{rgb}{0.690196,0.690196,0.690196}%
\pgfsetstrokecolor{currentstroke}%
\pgfsetdash{}{0pt}%
\pgfpathmoveto{\pgfqpoint{2.248264in}{0.582778in}}%
\pgfpathlineto{\pgfqpoint{2.248264in}{2.126667in}}%
\pgfusepath{stroke}%
\end{pgfscope}%
\begin{pgfscope}%
\pgfsetbuttcap%
\pgfsetroundjoin%
\definecolor{currentfill}{rgb}{0.000000,0.000000,0.000000}%
\pgfsetfillcolor{currentfill}%
\pgfsetlinewidth{0.803000pt}%
\definecolor{currentstroke}{rgb}{0.000000,0.000000,0.000000}%
\pgfsetstrokecolor{currentstroke}%
\pgfsetdash{}{0pt}%
\pgfsys@defobject{currentmarker}{\pgfqpoint{0.000000in}{-0.048611in}}{\pgfqpoint{0.000000in}{0.000000in}}{%
\pgfpathmoveto{\pgfqpoint{0.000000in}{0.000000in}}%
\pgfpathlineto{\pgfqpoint{0.000000in}{-0.048611in}}%
\pgfusepath{stroke,fill}%
}%
\begin{pgfscope}%
\pgfsys@transformshift{2.248264in}{0.582778in}%
\pgfsys@useobject{currentmarker}{}%
\end{pgfscope}%
\end{pgfscope}%
\begin{pgfscope}%
\definecolor{textcolor}{rgb}{0.000000,0.000000,0.000000}%
\pgfsetstrokecolor{textcolor}%
\pgfsetfillcolor{textcolor}%
\pgftext[x=2.248264in,y=0.485556in,,top]{\color{textcolor}{\sffamily\fontsize{10.000000}{12.000000}\selectfont\catcode`\^=\active\def^{\ifmmode\sp\else\^{}\fi}\catcode`\%=\active\def
\end{pgfscope}%
\begin{pgfscope}%
\definecolor{textcolor}{rgb}{0.000000,0.000000,0.000000}%
\pgfsetstrokecolor{textcolor}%
\pgfsetfillcolor{textcolor}%
\pgftext[x=1.380486in,y=0.295587in,,top]{\color{textcolor}{\sffamily\fontsize{10.000000}{12.000000}\selectfont\catcode`\^=\active\def^{\ifmmode\sp\else\^{}\fi}\catcode`\%=\active\def
\end{pgfscope}%
\begin{pgfscope}%
\pgfpathrectangle{\pgfqpoint{0.335972in}{0.582778in}}{\pgfqpoint{2.089028in}{1.543889in}}%
\pgfusepath{clip}%
\pgfsetrectcap%
\pgfsetroundjoin%
\pgfsetlinewidth{0.803000pt}%
\definecolor{currentstroke}{rgb}{0.690196,0.690196,0.690196}%
\pgfsetstrokecolor{currentstroke}%
\pgfsetdash{}{0pt}%
\pgfpathmoveto{\pgfqpoint{0.335972in}{1.065243in}}%
\pgfpathlineto{\pgfqpoint{2.425000in}{1.065243in}}%
\pgfusepath{stroke}%
\end{pgfscope}%
\begin{pgfscope}%
\pgfsetbuttcap%
\pgfsetroundjoin%
\definecolor{currentfill}{rgb}{0.000000,0.000000,0.000000}%
\pgfsetfillcolor{currentfill}%
\pgfsetlinewidth{0.803000pt}%
\definecolor{currentstroke}{rgb}{0.000000,0.000000,0.000000}%
\pgfsetstrokecolor{currentstroke}%
\pgfsetdash{}{0pt}%
\pgfsys@defobject{currentmarker}{\pgfqpoint{-0.048611in}{0.000000in}}{\pgfqpoint{-0.000000in}{0.000000in}}{%
\pgfpathmoveto{\pgfqpoint{-0.000000in}{0.000000in}}%
\pgfpathlineto{\pgfqpoint{-0.048611in}{0.000000in}}%
\pgfusepath{stroke,fill}%
}%
\begin{pgfscope}%
\pgfsys@transformshift{0.335972in}{1.065243in}%
\pgfsys@useobject{currentmarker}{}%
\end{pgfscope}%
\end{pgfscope}%
\begin{pgfscope}%
\definecolor{textcolor}{rgb}{0.000000,0.000000,0.000000}%
\pgfsetstrokecolor{textcolor}%
\pgfsetfillcolor{textcolor}%
\pgftext[x=0.150385in, y=1.012482in, left, base]{\color{textcolor}{\sffamily\fontsize{10.000000}{12.000000}\selectfont\catcode`\^=\active\def^{\ifmmode\sp\else\^{}\fi}\catcode`\%=\active\def
\end{pgfscope}%
\begin{pgfscope}%
\pgfpathrectangle{\pgfqpoint{0.335972in}{0.582778in}}{\pgfqpoint{2.089028in}{1.543889in}}%
\pgfusepath{clip}%
\pgfsetrectcap%
\pgfsetroundjoin%
\pgfsetlinewidth{0.803000pt}%
\definecolor{currentstroke}{rgb}{0.690196,0.690196,0.690196}%
\pgfsetstrokecolor{currentstroke}%
\pgfsetdash{}{0pt}%
\pgfpathmoveto{\pgfqpoint{0.335972in}{2.030174in}}%
\pgfpathlineto{\pgfqpoint{2.425000in}{2.030174in}}%
\pgfusepath{stroke}%
\end{pgfscope}%
\begin{pgfscope}%
\pgfsetbuttcap%
\pgfsetroundjoin%
\definecolor{currentfill}{rgb}{0.000000,0.000000,0.000000}%
\pgfsetfillcolor{currentfill}%
\pgfsetlinewidth{0.803000pt}%
\definecolor{currentstroke}{rgb}{0.000000,0.000000,0.000000}%
\pgfsetstrokecolor{currentstroke}%
\pgfsetdash{}{0pt}%
\pgfsys@defobject{currentmarker}{\pgfqpoint{-0.048611in}{0.000000in}}{\pgfqpoint{-0.000000in}{0.000000in}}{%
\pgfpathmoveto{\pgfqpoint{-0.000000in}{0.000000in}}%
\pgfpathlineto{\pgfqpoint{-0.048611in}{0.000000in}}%
\pgfusepath{stroke,fill}%
}%
\begin{pgfscope}%
\pgfsys@transformshift{0.335972in}{2.030174in}%
\pgfsys@useobject{currentmarker}{}%
\end{pgfscope}%
\end{pgfscope}%
\begin{pgfscope}%
\definecolor{textcolor}{rgb}{0.000000,0.000000,0.000000}%
\pgfsetstrokecolor{textcolor}%
\pgfsetfillcolor{textcolor}%
\pgftext[x=0.150385in, y=1.977412in, left, base]{\color{textcolor}{\sffamily\fontsize{10.000000}{12.000000}\selectfont\catcode`\^=\active\def^{\ifmmode\sp\else\^{}\fi}\catcode`\%=\active\def
\end{pgfscope}%
\begin{pgfscope}%
\pgfpathrectangle{\pgfqpoint{0.335972in}{0.582778in}}{\pgfqpoint{2.089028in}{1.543889in}}%
\pgfusepath{clip}%
\pgfsetrectcap%
\pgfsetroundjoin%
\pgfsetlinewidth{1.505625pt}%
\definecolor{currentstroke}{rgb}{0.274510,0.509804,0.705882}%
\pgfsetstrokecolor{currentstroke}%
\pgfsetdash{}{0pt}%
\pgfpathmoveto{\pgfqpoint{0.430928in}{2.030174in}}%
\pgfpathlineto{\pgfqpoint{0.440015in}{1.355151in}}%
\pgfpathlineto{\pgfqpoint{0.449101in}{1.011984in}}%
\pgfpathlineto{\pgfqpoint{0.458188in}{0.868615in}}%
\pgfpathlineto{\pgfqpoint{0.467275in}{0.819995in}}%
\pgfpathlineto{\pgfqpoint{0.476361in}{0.809436in}}%
\pgfpathlineto{\pgfqpoint{0.485448in}{0.833545in}}%
\pgfpathlineto{\pgfqpoint{0.494535in}{0.940126in}}%
\pgfpathlineto{\pgfqpoint{0.503621in}{1.223025in}}%
\pgfpathlineto{\pgfqpoint{0.512708in}{1.648726in}}%
\pgfpathlineto{\pgfqpoint{0.521795in}{1.846815in}}%
\pgfpathlineto{\pgfqpoint{0.530882in}{1.570666in}}%
\pgfpathlineto{\pgfqpoint{0.539968in}{1.146714in}}%
\pgfpathlineto{\pgfqpoint{0.549055in}{0.894542in}}%
\pgfpathlineto{\pgfqpoint{0.558142in}{0.800522in}}%
\pgfpathlineto{\pgfqpoint{0.567228in}{0.774148in}}%
\pgfpathlineto{\pgfqpoint{0.576315in}{0.775902in}}%
\pgfpathlineto{\pgfqpoint{0.585402in}{0.810632in}}%
\pgfpathlineto{\pgfqpoint{0.594488in}{0.932439in}}%
\pgfpathlineto{\pgfqpoint{0.603575in}{1.234842in}}%
\pgfpathlineto{\pgfqpoint{0.612662in}{1.661047in}}%
\pgfpathlineto{\pgfqpoint{0.621748in}{1.824656in}}%
\pgfpathlineto{\pgfqpoint{0.630835in}{1.524268in}}%
\pgfpathlineto{\pgfqpoint{0.639922in}{1.111605in}}%
\pgfpathlineto{\pgfqpoint{0.649008in}{0.877359in}}%
\pgfpathlineto{\pgfqpoint{0.658095in}{0.791493in}}%
\pgfpathlineto{\pgfqpoint{0.667182in}{0.767900in}}%
\pgfpathlineto{\pgfqpoint{0.676268in}{0.771761in}}%
\pgfpathlineto{\pgfqpoint{0.685355in}{0.811263in}}%
\pgfpathlineto{\pgfqpoint{0.694442in}{0.944296in}}%
\pgfpathlineto{\pgfqpoint{0.703528in}{1.261385in}}%
\pgfpathlineto{\pgfqpoint{0.712615in}{1.679748in}}%
\pgfpathlineto{\pgfqpoint{0.721702in}{1.803009in}}%
\pgfpathlineto{\pgfqpoint{0.730788in}{1.477894in}}%
\pgfpathlineto{\pgfqpoint{0.739875in}{1.075731in}}%
\pgfpathlineto{\pgfqpoint{0.748962in}{0.857244in}}%
\pgfpathlineto{\pgfqpoint{0.758049in}{0.778801in}}%
\pgfpathlineto{\pgfqpoint{0.767135in}{0.757976in}}%
\pgfpathlineto{\pgfqpoint{0.776222in}{0.763895in}}%
\pgfpathlineto{\pgfqpoint{0.785309in}{0.807743in}}%
\pgfpathlineto{\pgfqpoint{0.794395in}{0.952012in}}%
\pgfpathlineto{\pgfqpoint{0.803482in}{1.284447in}}%
\pgfpathlineto{\pgfqpoint{0.812569in}{1.696468in}}%
\pgfpathlineto{\pgfqpoint{0.821655in}{1.784051in}}%
\pgfpathlineto{\pgfqpoint{0.830742in}{1.441428in}}%
\pgfpathlineto{\pgfqpoint{0.839829in}{1.054846in}}%
\pgfpathlineto{\pgfqpoint{0.848915in}{0.855568in}}%
\pgfpathlineto{\pgfqpoint{0.858002in}{0.788669in}}%
\pgfpathlineto{\pgfqpoint{0.867089in}{0.777352in}}%
\pgfpathlineto{\pgfqpoint{0.876175in}{0.797526in}}%
\pgfpathlineto{\pgfqpoint{0.885262in}{0.870934in}}%
\pgfpathlineto{\pgfqpoint{0.894349in}{1.079461in}}%
\pgfpathlineto{\pgfqpoint{0.912522in}{1.976399in}}%
\pgfpathlineto{\pgfqpoint{0.921609in}{2.030174in}}%
\pgfpathlineto{\pgfqpoint{0.939782in}{1.150673in}}%
\pgfpathlineto{\pgfqpoint{0.948869in}{0.904371in}}%
\pgfpathlineto{\pgfqpoint{0.957955in}{0.808235in}}%
\pgfpathlineto{\pgfqpoint{0.967042in}{0.775963in}}%
\pgfpathlineto{\pgfqpoint{0.976129in}{0.775858in}}%
\pgfpathlineto{\pgfqpoint{0.985216in}{0.822390in}}%
\pgfpathlineto{\pgfqpoint{0.994302in}{0.986590in}}%
\pgfpathlineto{\pgfqpoint{1.012476in}{1.720043in}}%
\pgfpathlineto{\pgfqpoint{1.021562in}{1.720804in}}%
\pgfpathlineto{\pgfqpoint{1.039736in}{0.984437in}}%
\pgfpathlineto{\pgfqpoint{1.048822in}{0.811247in}}%
\pgfpathlineto{\pgfqpoint{1.057909in}{0.750541in}}%
\pgfpathlineto{\pgfqpoint{1.066996in}{0.735387in}}%
\pgfpathlineto{\pgfqpoint{1.076082in}{0.746502in}}%
\pgfpathlineto{\pgfqpoint{1.085169in}{0.804849in}}%
\pgfpathlineto{\pgfqpoint{1.094256in}{0.985708in}}%
\pgfpathlineto{\pgfqpoint{1.112429in}{1.715824in}}%
\pgfpathlineto{\pgfqpoint{1.121516in}{1.682808in}}%
\pgfpathlineto{\pgfqpoint{1.139689in}{0.960502in}}%
\pgfpathlineto{\pgfqpoint{1.148776in}{0.800540in}}%
\pgfpathlineto{\pgfqpoint{1.157862in}{0.744412in}}%
\pgfpathlineto{\pgfqpoint{1.166949in}{0.730963in}}%
\pgfpathlineto{\pgfqpoint{1.176036in}{0.744559in}}%
\pgfpathlineto{\pgfqpoint{1.185122in}{0.809426in}}%
\pgfpathlineto{\pgfqpoint{1.194209in}{1.003355in}}%
\pgfpathlineto{\pgfqpoint{1.212383in}{1.718764in}}%
\pgfpathlineto{\pgfqpoint{1.221469in}{1.648196in}}%
\pgfpathlineto{\pgfqpoint{1.230556in}{1.259053in}}%
\pgfpathlineto{\pgfqpoint{1.239643in}{0.934828in}}%
\pgfpathlineto{\pgfqpoint{1.248729in}{0.786397in}}%
\pgfpathlineto{\pgfqpoint{1.257816in}{0.734479in}}%
\pgfpathlineto{\pgfqpoint{1.266903in}{0.722686in}}%
\pgfpathlineto{\pgfqpoint{1.275989in}{0.738327in}}%
\pgfpathlineto{\pgfqpoint{1.285076in}{0.808833in}}%
\pgfpathlineto{\pgfqpoint{1.294163in}{1.015713in}}%
\pgfpathlineto{\pgfqpoint{1.303249in}{1.401466in}}%
\pgfpathlineto{\pgfqpoint{1.312336in}{1.718814in}}%
\pgfpathlineto{\pgfqpoint{1.321423in}{1.616289in}}%
\pgfpathlineto{\pgfqpoint{1.330509in}{1.226855in}}%
\pgfpathlineto{\pgfqpoint{1.339596in}{0.922494in}}%
\pgfpathlineto{\pgfqpoint{1.348683in}{0.788991in}}%
\pgfpathlineto{\pgfqpoint{1.357769in}{0.746145in}}%
\pgfpathlineto{\pgfqpoint{1.366856in}{0.744255in}}%
\pgfpathlineto{\pgfqpoint{1.375943in}{0.777646in}}%
\pgfpathlineto{\pgfqpoint{1.385029in}{0.888218in}}%
\pgfpathlineto{\pgfqpoint{1.394116in}{1.173306in}}%
\pgfpathlineto{\pgfqpoint{1.403203in}{1.646491in}}%
\pgfpathlineto{\pgfqpoint{1.412289in}{1.987165in}}%
\pgfpathlineto{\pgfqpoint{1.421376in}{1.830308in}}%
\pgfpathlineto{\pgfqpoint{1.430463in}{1.367921in}}%
\pgfpathlineto{\pgfqpoint{1.439550in}{1.006300in}}%
\pgfpathlineto{\pgfqpoint{1.448636in}{0.834343in}}%
\pgfpathlineto{\pgfqpoint{1.457723in}{0.766238in}}%
\pgfpathlineto{\pgfqpoint{1.466810in}{0.744119in}}%
\pgfpathlineto{\pgfqpoint{1.475896in}{0.755802in}}%
\pgfpathlineto{\pgfqpoint{1.484983in}{0.835277in}}%
\pgfpathlineto{\pgfqpoint{1.494070in}{1.068612in}}%
\pgfpathlineto{\pgfqpoint{1.503156in}{1.461361in}}%
\pgfpathlineto{\pgfqpoint{1.512243in}{1.716651in}}%
\pgfpathlineto{\pgfqpoint{1.521330in}{1.542518in}}%
\pgfpathlineto{\pgfqpoint{1.530416in}{1.152264in}}%
\pgfpathlineto{\pgfqpoint{1.539503in}{0.880877in}}%
\pgfpathlineto{\pgfqpoint{1.548590in}{0.764351in}}%
\pgfpathlineto{\pgfqpoint{1.557676in}{0.722587in}}%
\pgfpathlineto{\pgfqpoint{1.566763in}{0.714644in}}%
\pgfpathlineto{\pgfqpoint{1.575850in}{0.737186in}}%
\pgfpathlineto{\pgfqpoint{1.584936in}{0.830413in}}%
\pgfpathlineto{\pgfqpoint{1.594023in}{1.081506in}}%
\pgfpathlineto{\pgfqpoint{1.603110in}{1.476821in}}%
\pgfpathlineto{\pgfqpoint{1.612196in}{1.704243in}}%
\pgfpathlineto{\pgfqpoint{1.621283in}{1.506955in}}%
\pgfpathlineto{\pgfqpoint{1.630370in}{1.126176in}}%
\pgfpathlineto{\pgfqpoint{1.639456in}{0.872942in}}%
\pgfpathlineto{\pgfqpoint{1.648543in}{0.765050in}}%
\pgfpathlineto{\pgfqpoint{1.657630in}{0.725813in}}%
\pgfpathlineto{\pgfqpoint{1.666717in}{0.719129in}}%
\pgfpathlineto{\pgfqpoint{1.675803in}{0.744832in}}%
\pgfpathlineto{\pgfqpoint{1.684890in}{0.847360in}}%
\pgfpathlineto{\pgfqpoint{1.693977in}{1.112268in}}%
\pgfpathlineto{\pgfqpoint{1.703063in}{1.504635in}}%
\pgfpathlineto{\pgfqpoint{1.712150in}{1.698535in}}%
\pgfpathlineto{\pgfqpoint{1.721237in}{1.474378in}}%
\pgfpathlineto{\pgfqpoint{1.730323in}{1.099810in}}%
\pgfpathlineto{\pgfqpoint{1.739410in}{0.861989in}}%
\pgfpathlineto{\pgfqpoint{1.748497in}{0.761583in}}%
\pgfpathlineto{\pgfqpoint{1.757583in}{0.724542in}}%
\pgfpathlineto{\pgfqpoint{1.766670in}{0.718729in}}%
\pgfpathlineto{\pgfqpoint{1.775757in}{0.747271in}}%
\pgfpathlineto{\pgfqpoint{1.784843in}{0.859171in}}%
\pgfpathlineto{\pgfqpoint{1.793930in}{1.138099in}}%
\pgfpathlineto{\pgfqpoint{1.803017in}{1.527223in}}%
\pgfpathlineto{\pgfqpoint{1.812103in}{1.690525in}}%
\pgfpathlineto{\pgfqpoint{1.821190in}{1.446486in}}%
\pgfpathlineto{\pgfqpoint{1.830277in}{1.083479in}}%
\pgfpathlineto{\pgfqpoint{1.839363in}{0.864156in}}%
\pgfpathlineto{\pgfqpoint{1.848450in}{0.774838in}}%
\pgfpathlineto{\pgfqpoint{1.857537in}{0.745803in}}%
\pgfpathlineto{\pgfqpoint{1.866624in}{0.751405in}}%
\pgfpathlineto{\pgfqpoint{1.875710in}{0.804431in}}%
\pgfpathlineto{\pgfqpoint{1.884797in}{0.971029in}}%
\pgfpathlineto{\pgfqpoint{1.893884in}{1.336220in}}%
\pgfpathlineto{\pgfqpoint{1.902970in}{1.791577in}}%
\pgfpathlineto{\pgfqpoint{1.912057in}{1.945174in}}%
\pgfpathlineto{\pgfqpoint{1.921144in}{1.638194in}}%
\pgfpathlineto{\pgfqpoint{1.930230in}{1.212160in}}%
\pgfpathlineto{\pgfqpoint{1.939317in}{0.945547in}}%
\pgfpathlineto{\pgfqpoint{1.948404in}{0.823628in}}%
\pgfpathlineto{\pgfqpoint{1.957490in}{0.770873in}}%
\pgfpathlineto{\pgfqpoint{1.966577in}{0.755028in}}%
\pgfpathlineto{\pgfqpoint{1.975664in}{0.782879in}}%
\pgfpathlineto{\pgfqpoint{1.984750in}{0.911351in}}%
\pgfpathlineto{\pgfqpoint{1.993837in}{1.213010in}}%
\pgfpathlineto{\pgfqpoint{2.002924in}{1.579216in}}%
\pgfpathlineto{\pgfqpoint{2.012010in}{1.665997in}}%
\pgfpathlineto{\pgfqpoint{2.021097in}{1.378178in}}%
\pgfpathlineto{\pgfqpoint{2.030184in}{1.035002in}}%
\pgfpathlineto{\pgfqpoint{2.039270in}{0.842668in}}%
\pgfpathlineto{\pgfqpoint{2.048357in}{0.761420in}}%
\pgfpathlineto{\pgfqpoint{2.057444in}{0.729352in}}%
\pgfpathlineto{\pgfqpoint{2.066530in}{0.726355in}}%
\pgfpathlineto{\pgfqpoint{2.075617in}{0.766222in}}%
\pgfpathlineto{\pgfqpoint{2.084704in}{0.911086in}}%
\pgfpathlineto{\pgfqpoint{2.093791in}{1.227009in}}%
\pgfpathlineto{\pgfqpoint{2.102877in}{1.583419in}}%
\pgfpathlineto{\pgfqpoint{2.111964in}{1.640502in}}%
\pgfpathlineto{\pgfqpoint{2.130137in}{1.016734in}}%
\pgfpathlineto{\pgfqpoint{2.139224in}{0.838789in}}%
\pgfpathlineto{\pgfqpoint{2.148311in}{0.762795in}}%
\pgfpathlineto{\pgfqpoint{2.157397in}{0.732246in}}%
\pgfpathlineto{\pgfqpoint{2.166484in}{0.730921in}}%
\pgfpathlineto{\pgfqpoint{2.175571in}{0.776307in}}%
\pgfpathlineto{\pgfqpoint{2.184657in}{0.933552in}}%
\pgfpathlineto{\pgfqpoint{2.202831in}{1.597648in}}%
\pgfpathlineto{\pgfqpoint{2.211917in}{1.619977in}}%
\pgfpathlineto{\pgfqpoint{2.230091in}{0.998215in}}%
\pgfpathlineto{\pgfqpoint{2.239177in}{0.832164in}}%
\pgfpathlineto{\pgfqpoint{2.248264in}{0.760574in}}%
\pgfpathlineto{\pgfqpoint{2.257351in}{0.731350in}}%
\pgfpathlineto{\pgfqpoint{2.266437in}{0.731384in}}%
\pgfpathlineto{\pgfqpoint{2.275524in}{0.782123in}}%
\pgfpathlineto{\pgfqpoint{2.284611in}{0.951747in}}%
\pgfpathlineto{\pgfqpoint{2.302784in}{1.609223in}}%
\pgfpathlineto{\pgfqpoint{2.311871in}{1.601035in}}%
\pgfpathlineto{\pgfqpoint{2.330044in}{0.989516in}}%
\pgfpathlineto{\pgfqpoint{2.330044in}{0.989516in}}%
\pgfusepath{stroke}%
\end{pgfscope}%
\begin{pgfscope}%
\pgfsetrectcap%
\pgfsetmiterjoin%
\pgfsetlinewidth{0.803000pt}%
\definecolor{currentstroke}{rgb}{0.000000,0.000000,0.000000}%
\pgfsetstrokecolor{currentstroke}%
\pgfsetdash{}{0pt}%
\pgfpathmoveto{\pgfqpoint{0.335972in}{0.582778in}}%
\pgfpathlineto{\pgfqpoint{0.335972in}{2.126667in}}%
\pgfusepath{stroke}%
\end{pgfscope}%
\begin{pgfscope}%
\pgfsetrectcap%
\pgfsetmiterjoin%
\pgfsetlinewidth{0.803000pt}%
\definecolor{currentstroke}{rgb}{0.000000,0.000000,0.000000}%
\pgfsetstrokecolor{currentstroke}%
\pgfsetdash{}{0pt}%
\pgfpathmoveto{\pgfqpoint{2.425000in}{0.582778in}}%
\pgfpathlineto{\pgfqpoint{2.425000in}{2.126667in}}%
\pgfusepath{stroke}%
\end{pgfscope}%
\begin{pgfscope}%
\pgfsetrectcap%
\pgfsetmiterjoin%
\pgfsetlinewidth{0.803000pt}%
\definecolor{currentstroke}{rgb}{0.000000,0.000000,0.000000}%
\pgfsetstrokecolor{currentstroke}%
\pgfsetdash{}{0pt}%
\pgfpathmoveto{\pgfqpoint{0.335972in}{0.582778in}}%
\pgfpathlineto{\pgfqpoint{2.425000in}{0.582778in}}%
\pgfusepath{stroke}%
\end{pgfscope}%
\begin{pgfscope}%
\pgfsetrectcap%
\pgfsetmiterjoin%
\pgfsetlinewidth{0.803000pt}%
\definecolor{currentstroke}{rgb}{0.000000,0.000000,0.000000}%
\pgfsetstrokecolor{currentstroke}%
\pgfsetdash{}{0pt}%
\pgfpathmoveto{\pgfqpoint{0.335972in}{2.126667in}}%
\pgfpathlineto{\pgfqpoint{2.425000in}{2.126667in}}%
\pgfusepath{stroke}%
\end{pgfscope}%
\begin{pgfscope}%
\definecolor{textcolor}{rgb}{0.000000,0.000000,0.000000}%
\pgfsetstrokecolor{textcolor}%
\pgfsetfillcolor{textcolor}%
\pgftext[x=1.380486in,y=2.210000in,,base]{\color{textcolor}{\sffamily\fontsize{12.000000}{14.400000}\selectfont\catcode`\^=\active\def^{\ifmmode\sp\else\^{}\fi}\catcode`\%=\active\def
\end{pgfscope}%
\begin{pgfscope}%
\pgfsetbuttcap%
\pgfsetmiterjoin%
\definecolor{currentfill}{rgb}{1.000000,1.000000,1.000000}%
\pgfsetfillcolor{currentfill}%
\pgfsetlinewidth{0.000000pt}%
\definecolor{currentstroke}{rgb}{0.000000,0.000000,0.000000}%
\pgfsetstrokecolor{currentstroke}%
\pgfsetstrokeopacity{0.000000}%
\pgfsetdash{}{0pt}%
\pgfpathmoveto{\pgfqpoint{2.760972in}{0.582778in}}%
\pgfpathlineto{\pgfqpoint{4.850000in}{0.582778in}}%
\pgfpathlineto{\pgfqpoint{4.850000in}{2.126667in}}%
\pgfpathlineto{\pgfqpoint{2.760972in}{2.126667in}}%
\pgfpathlineto{\pgfqpoint{2.760972in}{0.582778in}}%
\pgfpathclose%
\pgfusepath{fill}%
\end{pgfscope}%
\begin{pgfscope}%
\pgfpathrectangle{\pgfqpoint{2.760972in}{0.582778in}}{\pgfqpoint{2.089028in}{1.543889in}}%
\pgfusepath{clip}%
\pgfsetrectcap%
\pgfsetroundjoin%
\pgfsetlinewidth{0.803000pt}%
\definecolor{currentstroke}{rgb}{0.690196,0.690196,0.690196}%
\pgfsetstrokecolor{currentstroke}%
\pgfsetdash{}{0pt}%
\pgfpathmoveto{\pgfqpoint{2.855928in}{0.582778in}}%
\pgfpathlineto{\pgfqpoint{2.855928in}{2.126667in}}%
\pgfusepath{stroke}%
\end{pgfscope}%
\begin{pgfscope}%
\pgfsetbuttcap%
\pgfsetroundjoin%
\definecolor{currentfill}{rgb}{0.000000,0.000000,0.000000}%
\pgfsetfillcolor{currentfill}%
\pgfsetlinewidth{0.803000pt}%
\definecolor{currentstroke}{rgb}{0.000000,0.000000,0.000000}%
\pgfsetstrokecolor{currentstroke}%
\pgfsetdash{}{0pt}%
\pgfsys@defobject{currentmarker}{\pgfqpoint{0.000000in}{-0.048611in}}{\pgfqpoint{0.000000in}{0.000000in}}{%
\pgfpathmoveto{\pgfqpoint{0.000000in}{0.000000in}}%
\pgfpathlineto{\pgfqpoint{0.000000in}{-0.048611in}}%
\pgfusepath{stroke,fill}%
}%
\begin{pgfscope}%
\pgfsys@transformshift{2.855928in}{0.582778in}%
\pgfsys@useobject{currentmarker}{}%
\end{pgfscope}%
\end{pgfscope}%
\begin{pgfscope}%
\definecolor{textcolor}{rgb}{0.000000,0.000000,0.000000}%
\pgfsetstrokecolor{textcolor}%
\pgfsetfillcolor{textcolor}%
\pgftext[x=2.855928in,y=0.485556in,,top]{\color{textcolor}{\sffamily\fontsize{10.000000}{12.000000}\selectfont\catcode`\^=\active\def^{\ifmmode\sp\else\^{}\fi}\catcode`\%=\active\def
\end{pgfscope}%
\begin{pgfscope}%
\pgfpathrectangle{\pgfqpoint{2.760972in}{0.582778in}}{\pgfqpoint{2.089028in}{1.543889in}}%
\pgfusepath{clip}%
\pgfsetrectcap%
\pgfsetroundjoin%
\pgfsetlinewidth{0.803000pt}%
\definecolor{currentstroke}{rgb}{0.690196,0.690196,0.690196}%
\pgfsetstrokecolor{currentstroke}%
\pgfsetdash{}{0pt}%
\pgfpathmoveto{\pgfqpoint{3.764596in}{0.582778in}}%
\pgfpathlineto{\pgfqpoint{3.764596in}{2.126667in}}%
\pgfusepath{stroke}%
\end{pgfscope}%
\begin{pgfscope}%
\pgfsetbuttcap%
\pgfsetroundjoin%
\definecolor{currentfill}{rgb}{0.000000,0.000000,0.000000}%
\pgfsetfillcolor{currentfill}%
\pgfsetlinewidth{0.803000pt}%
\definecolor{currentstroke}{rgb}{0.000000,0.000000,0.000000}%
\pgfsetstrokecolor{currentstroke}%
\pgfsetdash{}{0pt}%
\pgfsys@defobject{currentmarker}{\pgfqpoint{0.000000in}{-0.048611in}}{\pgfqpoint{0.000000in}{0.000000in}}{%
\pgfpathmoveto{\pgfqpoint{0.000000in}{0.000000in}}%
\pgfpathlineto{\pgfqpoint{0.000000in}{-0.048611in}}%
\pgfusepath{stroke,fill}%
}%
\begin{pgfscope}%
\pgfsys@transformshift{3.764596in}{0.582778in}%
\pgfsys@useobject{currentmarker}{}%
\end{pgfscope}%
\end{pgfscope}%
\begin{pgfscope}%
\definecolor{textcolor}{rgb}{0.000000,0.000000,0.000000}%
\pgfsetstrokecolor{textcolor}%
\pgfsetfillcolor{textcolor}%
\pgftext[x=3.764596in,y=0.485556in,,top]{\color{textcolor}{\sffamily\fontsize{10.000000}{12.000000}\selectfont\catcode`\^=\active\def^{\ifmmode\sp\else\^{}\fi}\catcode`\%=\active\def
\end{pgfscope}%
\begin{pgfscope}%
\pgfpathrectangle{\pgfqpoint{2.760972in}{0.582778in}}{\pgfqpoint{2.089028in}{1.543889in}}%
\pgfusepath{clip}%
\pgfsetrectcap%
\pgfsetroundjoin%
\pgfsetlinewidth{0.803000pt}%
\definecolor{currentstroke}{rgb}{0.690196,0.690196,0.690196}%
\pgfsetstrokecolor{currentstroke}%
\pgfsetdash{}{0pt}%
\pgfpathmoveto{\pgfqpoint{4.673264in}{0.582778in}}%
\pgfpathlineto{\pgfqpoint{4.673264in}{2.126667in}}%
\pgfusepath{stroke}%
\end{pgfscope}%
\begin{pgfscope}%
\pgfsetbuttcap%
\pgfsetroundjoin%
\definecolor{currentfill}{rgb}{0.000000,0.000000,0.000000}%
\pgfsetfillcolor{currentfill}%
\pgfsetlinewidth{0.803000pt}%
\definecolor{currentstroke}{rgb}{0.000000,0.000000,0.000000}%
\pgfsetstrokecolor{currentstroke}%
\pgfsetdash{}{0pt}%
\pgfsys@defobject{currentmarker}{\pgfqpoint{0.000000in}{-0.048611in}}{\pgfqpoint{0.000000in}{0.000000in}}{%
\pgfpathmoveto{\pgfqpoint{0.000000in}{0.000000in}}%
\pgfpathlineto{\pgfqpoint{0.000000in}{-0.048611in}}%
\pgfusepath{stroke,fill}%
}%
\begin{pgfscope}%
\pgfsys@transformshift{4.673264in}{0.582778in}%
\pgfsys@useobject{currentmarker}{}%
\end{pgfscope}%
\end{pgfscope}%
\begin{pgfscope}%
\definecolor{textcolor}{rgb}{0.000000,0.000000,0.000000}%
\pgfsetstrokecolor{textcolor}%
\pgfsetfillcolor{textcolor}%
\pgftext[x=4.673264in,y=0.485556in,,top]{\color{textcolor}{\sffamily\fontsize{10.000000}{12.000000}\selectfont\catcode`\^=\active\def^{\ifmmode\sp\else\^{}\fi}\catcode`\%=\active\def
\end{pgfscope}%
\begin{pgfscope}%
\pgfpathrectangle{\pgfqpoint{2.760972in}{0.582778in}}{\pgfqpoint{2.089028in}{1.543889in}}%
\pgfusepath{clip}%
\pgfsetrectcap%
\pgfsetroundjoin%
\pgfsetlinewidth{0.803000pt}%
\definecolor{currentstroke}{rgb}{0.690196,0.690196,0.690196}%
\pgfsetstrokecolor{currentstroke}%
\pgfsetdash{}{0pt}%
\pgfpathmoveto{\pgfqpoint{2.760972in}{1.065243in}}%
\pgfpathlineto{\pgfqpoint{4.850000in}{1.065243in}}%
\pgfusepath{stroke}%
\end{pgfscope}%
\begin{pgfscope}%
\pgfsetbuttcap%
\pgfsetroundjoin%
\definecolor{currentfill}{rgb}{0.000000,0.000000,0.000000}%
\pgfsetfillcolor{currentfill}%
\pgfsetlinewidth{0.803000pt}%
\definecolor{currentstroke}{rgb}{0.000000,0.000000,0.000000}%
\pgfsetstrokecolor{currentstroke}%
\pgfsetdash{}{0pt}%
\pgfsys@defobject{currentmarker}{\pgfqpoint{-0.048611in}{0.000000in}}{\pgfqpoint{-0.000000in}{0.000000in}}{%
\pgfpathmoveto{\pgfqpoint{-0.000000in}{0.000000in}}%
\pgfpathlineto{\pgfqpoint{-0.048611in}{0.000000in}}%
\pgfusepath{stroke,fill}%
}%
\begin{pgfscope}%
\pgfsys@transformshift{2.760972in}{1.065243in}%
\pgfsys@useobject{currentmarker}{}%
\end{pgfscope}%
\end{pgfscope}%
\begin{pgfscope}%
\definecolor{textcolor}{rgb}{0.000000,0.000000,0.000000}%
\pgfsetstrokecolor{textcolor}%
\pgfsetfillcolor{textcolor}%
\pgftext[x=2.575385in, y=1.012482in, left, base]{\color{textcolor}{\sffamily\fontsize{10.000000}{12.000000}\selectfont\catcode`\^=\active\def^{\ifmmode\sp\else\^{}\fi}\catcode`\%=\active\def
\end{pgfscope}%
\begin{pgfscope}%
\pgfpathrectangle{\pgfqpoint{2.760972in}{0.582778in}}{\pgfqpoint{2.089028in}{1.543889in}}%
\pgfusepath{clip}%
\pgfsetrectcap%
\pgfsetroundjoin%
\pgfsetlinewidth{0.803000pt}%
\definecolor{currentstroke}{rgb}{0.690196,0.690196,0.690196}%
\pgfsetstrokecolor{currentstroke}%
\pgfsetdash{}{0pt}%
\pgfpathmoveto{\pgfqpoint{2.760972in}{2.030174in}}%
\pgfpathlineto{\pgfqpoint{4.850000in}{2.030174in}}%
\pgfusepath{stroke}%
\end{pgfscope}%
\begin{pgfscope}%
\pgfsetbuttcap%
\pgfsetroundjoin%
\definecolor{currentfill}{rgb}{0.000000,0.000000,0.000000}%
\pgfsetfillcolor{currentfill}%
\pgfsetlinewidth{0.803000pt}%
\definecolor{currentstroke}{rgb}{0.000000,0.000000,0.000000}%
\pgfsetstrokecolor{currentstroke}%
\pgfsetdash{}{0pt}%
\pgfsys@defobject{currentmarker}{\pgfqpoint{-0.048611in}{0.000000in}}{\pgfqpoint{-0.000000in}{0.000000in}}{%
\pgfpathmoveto{\pgfqpoint{-0.000000in}{0.000000in}}%
\pgfpathlineto{\pgfqpoint{-0.048611in}{0.000000in}}%
\pgfusepath{stroke,fill}%
}%
\begin{pgfscope}%
\pgfsys@transformshift{2.760972in}{2.030174in}%
\pgfsys@useobject{currentmarker}{}%
\end{pgfscope}%
\end{pgfscope}%
\begin{pgfscope}%
\definecolor{textcolor}{rgb}{0.000000,0.000000,0.000000}%
\pgfsetstrokecolor{textcolor}%
\pgfsetfillcolor{textcolor}%
\pgftext[x=2.575385in, y=1.977412in, left, base]{\color{textcolor}{\sffamily\fontsize{10.000000}{12.000000}\selectfont\catcode`\^=\active\def^{\ifmmode\sp\else\^{}\fi}\catcode`\%=\active\def
\end{pgfscope}%
\begin{pgfscope}%
\pgfpathrectangle{\pgfqpoint{2.760972in}{0.582778in}}{\pgfqpoint{2.089028in}{1.543889in}}%
\pgfusepath{clip}%
\pgfsetrectcap%
\pgfsetroundjoin%
\pgfsetlinewidth{1.505625pt}%
\definecolor{currentstroke}{rgb}{0.274510,0.509804,0.705882}%
\pgfsetstrokecolor{currentstroke}%
\pgfsetdash{}{0pt}%
\pgfpathmoveto{\pgfqpoint{2.855928in}{2.030174in}}%
\pgfpathlineto{\pgfqpoint{2.865015in}{1.065243in}}%
\pgfpathlineto{\pgfqpoint{2.946795in}{1.065243in}}%
\pgfpathlineto{\pgfqpoint{2.955882in}{1.837188in}}%
\pgfpathlineto{\pgfqpoint{2.964968in}{1.065243in}}%
\pgfpathlineto{\pgfqpoint{3.046748in}{1.065243in}}%
\pgfpathlineto{\pgfqpoint{3.055835in}{1.837188in}}%
\pgfpathlineto{\pgfqpoint{3.064922in}{1.065243in}}%
\pgfpathlineto{\pgfqpoint{3.146702in}{1.065243in}}%
\pgfpathlineto{\pgfqpoint{3.155788in}{1.837188in}}%
\pgfpathlineto{\pgfqpoint{3.164875in}{1.065243in}}%
\pgfpathlineto{\pgfqpoint{3.246655in}{1.065243in}}%
\pgfpathlineto{\pgfqpoint{3.255742in}{1.837188in}}%
\pgfpathlineto{\pgfqpoint{3.264829in}{1.065243in}}%
\pgfpathlineto{\pgfqpoint{3.346609in}{1.065243in}}%
\pgfpathlineto{\pgfqpoint{3.355695in}{2.030174in}}%
\pgfpathlineto{\pgfqpoint{3.364782in}{1.065243in}}%
\pgfpathlineto{\pgfqpoint{3.437476in}{1.065243in}}%
\pgfpathlineto{\pgfqpoint{3.446562in}{1.837188in}}%
\pgfpathlineto{\pgfqpoint{3.455649in}{1.065243in}}%
\pgfpathlineto{\pgfqpoint{3.537429in}{1.065243in}}%
\pgfpathlineto{\pgfqpoint{3.546516in}{1.837188in}}%
\pgfpathlineto{\pgfqpoint{3.555602in}{1.065243in}}%
\pgfpathlineto{\pgfqpoint{3.637383in}{1.065243in}}%
\pgfpathlineto{\pgfqpoint{3.646469in}{1.837188in}}%
\pgfpathlineto{\pgfqpoint{3.655556in}{1.065243in}}%
\pgfpathlineto{\pgfqpoint{3.737336in}{1.065243in}}%
\pgfpathlineto{\pgfqpoint{3.746423in}{1.837188in}}%
\pgfpathlineto{\pgfqpoint{3.755509in}{1.065243in}}%
\pgfpathlineto{\pgfqpoint{3.837289in}{1.065243in}}%
\pgfpathlineto{\pgfqpoint{3.846376in}{2.030174in}}%
\pgfpathlineto{\pgfqpoint{3.855463in}{1.065243in}}%
\pgfpathlineto{\pgfqpoint{3.937243in}{1.065243in}}%
\pgfpathlineto{\pgfqpoint{3.946330in}{1.837188in}}%
\pgfpathlineto{\pgfqpoint{3.955416in}{1.065243in}}%
\pgfpathlineto{\pgfqpoint{4.037196in}{1.065243in}}%
\pgfpathlineto{\pgfqpoint{4.046283in}{1.837188in}}%
\pgfpathlineto{\pgfqpoint{4.055370in}{1.065243in}}%
\pgfpathlineto{\pgfqpoint{4.137150in}{1.065243in}}%
\pgfpathlineto{\pgfqpoint{4.146237in}{1.837188in}}%
\pgfpathlineto{\pgfqpoint{4.155323in}{1.065243in}}%
\pgfpathlineto{\pgfqpoint{4.237103in}{1.065243in}}%
\pgfpathlineto{\pgfqpoint{4.246190in}{1.837188in}}%
\pgfpathlineto{\pgfqpoint{4.255277in}{1.065243in}}%
\pgfpathlineto{\pgfqpoint{4.337057in}{1.065243in}}%
\pgfpathlineto{\pgfqpoint{4.346144in}{2.030174in}}%
\pgfpathlineto{\pgfqpoint{4.355230in}{1.065243in}}%
\pgfpathlineto{\pgfqpoint{4.437010in}{1.065243in}}%
\pgfpathlineto{\pgfqpoint{4.446097in}{1.837188in}}%
\pgfpathlineto{\pgfqpoint{4.455184in}{1.065243in}}%
\pgfpathlineto{\pgfqpoint{4.536964in}{1.065243in}}%
\pgfpathlineto{\pgfqpoint{4.546051in}{1.837188in}}%
\pgfpathlineto{\pgfqpoint{4.555137in}{1.065243in}}%
\pgfpathlineto{\pgfqpoint{4.627831in}{1.065243in}}%
\pgfpathlineto{\pgfqpoint{4.636917in}{1.837188in}}%
\pgfpathlineto{\pgfqpoint{4.646004in}{1.065243in}}%
\pgfpathlineto{\pgfqpoint{4.727784in}{1.065243in}}%
\pgfpathlineto{\pgfqpoint{4.736871in}{1.837188in}}%
\pgfpathlineto{\pgfqpoint{4.745958in}{1.065243in}}%
\pgfpathlineto{\pgfqpoint{4.755044in}{1.065243in}}%
\pgfpathlineto{\pgfqpoint{4.755044in}{1.065243in}}%
\pgfusepath{stroke}%
\end{pgfscope}%
\begin{pgfscope}%
\pgfsetrectcap%
\pgfsetmiterjoin%
\pgfsetlinewidth{0.803000pt}%
\definecolor{currentstroke}{rgb}{0.000000,0.000000,0.000000}%
\pgfsetstrokecolor{currentstroke}%
\pgfsetdash{}{0pt}%
\pgfpathmoveto{\pgfqpoint{2.760972in}{0.582778in}}%
\pgfpathlineto{\pgfqpoint{2.760972in}{2.126667in}}%
\pgfusepath{stroke}%
\end{pgfscope}%
\begin{pgfscope}%
\pgfsetrectcap%
\pgfsetmiterjoin%
\pgfsetlinewidth{0.803000pt}%
\definecolor{currentstroke}{rgb}{0.000000,0.000000,0.000000}%
\pgfsetstrokecolor{currentstroke}%
\pgfsetdash{}{0pt}%
\pgfpathmoveto{\pgfqpoint{4.850000in}{0.582778in}}%
\pgfpathlineto{\pgfqpoint{4.850000in}{2.126667in}}%
\pgfusepath{stroke}%
\end{pgfscope}%
\begin{pgfscope}%
\pgfsetrectcap%
\pgfsetmiterjoin%
\pgfsetlinewidth{0.803000pt}%
\definecolor{currentstroke}{rgb}{0.000000,0.000000,0.000000}%
\pgfsetstrokecolor{currentstroke}%
\pgfsetdash{}{0pt}%
\pgfpathmoveto{\pgfqpoint{2.760972in}{0.582778in}}%
\pgfpathlineto{\pgfqpoint{4.850000in}{0.582778in}}%
\pgfusepath{stroke}%
\end{pgfscope}%
\begin{pgfscope}%
\pgfsetrectcap%
\pgfsetmiterjoin%
\pgfsetlinewidth{0.803000pt}%
\definecolor{currentstroke}{rgb}{0.000000,0.000000,0.000000}%
\pgfsetstrokecolor{currentstroke}%
\pgfsetdash{}{0pt}%
\pgfpathmoveto{\pgfqpoint{2.760972in}{2.126667in}}%
\pgfpathlineto{\pgfqpoint{4.850000in}{2.126667in}}%
\pgfusepath{stroke}%
\end{pgfscope}%
\begin{pgfscope}%
\definecolor{textcolor}{rgb}{0.000000,0.000000,0.000000}%
\pgfsetstrokecolor{textcolor}%
\pgfsetfillcolor{textcolor}%
\pgftext[x=3.805486in,y=2.210000in,,base]{\color{textcolor}{\sffamily\fontsize{12.000000}{14.400000}\selectfont\catcode`\^=\active\def^{\ifmmode\sp\else\^{}\fi}\catcode`\%=\active\def
\end{pgfscope}%
\end{pgfpicture}%
\makeatother%
\endgroup%

%% file: figures/pipeline/linear_sum_assignment_example.pgf
\begingroup%
\makeatletter%
\begin{pgfpicture}%
\pgfpathrectangle{\pgfpointorigin}{\pgfqpoint{5.000000in}{2.800000in}}%
\pgfusepath{use as bounding box, clip}%
\begin{pgfscope}%
\pgfsetbuttcap%
\pgfsetmiterjoin%
\definecolor{currentfill}{rgb}{1.000000,1.000000,1.000000}%
\pgfsetfillcolor{currentfill}%
\pgfsetlinewidth{0.000000pt}%
\definecolor{currentstroke}{rgb}{1.000000,1.000000,1.000000}%
\pgfsetstrokecolor{currentstroke}%
\pgfsetdash{}{0pt}%
\pgfpathmoveto{\pgfqpoint{0.000000in}{0.000000in}}%
\pgfpathlineto{\pgfqpoint{5.000000in}{0.000000in}}%
\pgfpathlineto{\pgfqpoint{5.000000in}{2.800000in}}%
\pgfpathlineto{\pgfqpoint{0.000000in}{2.800000in}}%
\pgfpathlineto{\pgfqpoint{0.000000in}{0.000000in}}%
\pgfpathclose%
\pgfusepath{fill}%
\end{pgfscope}%
\begin{pgfscope}%
\pgfsetbuttcap%
\pgfsetmiterjoin%
\definecolor{currentfill}{rgb}{1.000000,1.000000,1.000000}%
\pgfsetfillcolor{currentfill}%
\pgfsetlinewidth{0.000000pt}%
\definecolor{currentstroke}{rgb}{0.000000,0.000000,0.000000}%
\pgfsetstrokecolor{currentstroke}%
\pgfsetstrokeopacity{0.000000}%
\pgfsetdash{}{0pt}%
\pgfpathmoveto{\pgfqpoint{0.150000in}{0.150000in}}%
\pgfpathlineto{\pgfqpoint{4.850000in}{0.150000in}}%
\pgfpathlineto{\pgfqpoint{4.850000in}{2.650000in}}%
\pgfpathlineto{\pgfqpoint{0.150000in}{2.650000in}}%
\pgfpathlineto{\pgfqpoint{0.150000in}{0.150000in}}%
\pgfpathclose%
\pgfusepath{fill}%
\end{pgfscope}%
\begin{pgfscope}%
\pgfpathrectangle{\pgfqpoint{0.150000in}{0.150000in}}{\pgfqpoint{4.700000in}{2.500000in}}%
\pgfusepath{clip}%
\pgfsetrectcap%
\pgfsetroundjoin%
\pgfsetlinewidth{2.007500pt}%
\definecolor{currentstroke}{rgb}{0.000000,0.000000,0.000000}%
\pgfsetstrokecolor{currentstroke}%
\pgfsetdash{}{0pt}%
\pgfpathmoveto{\pgfqpoint{4.302184in}{2.400000in}}%
\pgfpathlineto{\pgfqpoint{3.705434in}{2.337500in}}%
\pgfusepath{stroke}%
\end{pgfscope}%
\begin{pgfscope}%
\pgfpathrectangle{\pgfqpoint{0.150000in}{0.150000in}}{\pgfqpoint{4.700000in}{2.500000in}}%
\pgfusepath{clip}%
\pgfsetrectcap%
\pgfsetroundjoin%
\pgfsetlinewidth{2.007500pt}%
\definecolor{currentstroke}{rgb}{0.000000,0.000000,0.000000}%
\pgfsetstrokecolor{currentstroke}%
\pgfsetdash{}{0pt}%
\pgfpathmoveto{\pgfqpoint{0.453149in}{1.650000in}}%
\pgfpathlineto{\pgfqpoint{1.527298in}{1.650000in}}%
\pgfusepath{stroke}%
\end{pgfscope}%
\begin{pgfscope}%
\pgfpathrectangle{\pgfqpoint{0.150000in}{0.150000in}}{\pgfqpoint{4.700000in}{2.500000in}}%
\pgfusepath{clip}%
\pgfsetrectcap%
\pgfsetroundjoin%
\pgfsetlinewidth{2.007500pt}%
\definecolor{currentstroke}{rgb}{0.000000,0.000000,0.000000}%
\pgfsetstrokecolor{currentstroke}%
\pgfsetdash{}{0pt}%
\pgfpathmoveto{\pgfqpoint{2.004698in}{1.531250in}}%
\pgfpathlineto{\pgfqpoint{3.078847in}{1.531250in}}%
\pgfusepath{stroke}%
\end{pgfscope}%
\begin{pgfscope}%
\pgfpathrectangle{\pgfqpoint{0.150000in}{0.150000in}}{\pgfqpoint{4.700000in}{2.500000in}}%
\pgfusepath{clip}%
\pgfsetrectcap%
\pgfsetroundjoin%
\pgfsetlinewidth{2.007500pt}%
\definecolor{currentstroke}{rgb}{0.000000,0.000000,0.000000}%
\pgfsetstrokecolor{currentstroke}%
\pgfsetdash{}{0pt}%
\pgfpathmoveto{\pgfqpoint{3.245937in}{1.350000in}}%
\pgfpathlineto{\pgfqpoint{4.320086in}{1.350000in}}%
\pgfusepath{stroke}%
\end{pgfscope}%
\begin{pgfscope}%
\pgfpathrectangle{\pgfqpoint{0.150000in}{0.150000in}}{\pgfqpoint{4.700000in}{2.500000in}}%
\pgfusepath{clip}%
\pgfsetrectcap%
\pgfsetroundjoin%
\pgfsetlinewidth{2.007500pt}%
\definecolor{currentstroke}{rgb}{0.000000,0.000000,0.000000}%
\pgfsetstrokecolor{currentstroke}%
\pgfsetdash{}{0pt}%
\pgfpathmoveto{\pgfqpoint{0.363636in}{1.012500in}}%
\pgfpathlineto{\pgfqpoint{1.437786in}{1.012500in}}%
\pgfusepath{stroke}%
\end{pgfscope}%
\begin{pgfscope}%
\pgfpathrectangle{\pgfqpoint{0.150000in}{0.150000in}}{\pgfqpoint{4.700000in}{2.500000in}}%
\pgfusepath{clip}%
\pgfsetrectcap%
\pgfsetroundjoin%
\pgfsetlinewidth{2.007500pt}%
\definecolor{currentstroke}{rgb}{0.000000,0.000000,0.000000}%
\pgfsetstrokecolor{currentstroke}%
\pgfsetdash{}{0pt}%
\pgfpathmoveto{\pgfqpoint{2.016633in}{1.181250in}}%
\pgfpathlineto{\pgfqpoint{3.090782in}{1.181250in}}%
\pgfusepath{stroke}%
\end{pgfscope}%
\begin{pgfscope}%
\pgfpathrectangle{\pgfqpoint{0.150000in}{0.150000in}}{\pgfqpoint{4.700000in}{2.500000in}}%
\pgfusepath{clip}%
\pgfsetrectcap%
\pgfsetroundjoin%
\pgfsetlinewidth{2.007500pt}%
\definecolor{currentstroke}{rgb}{0.000000,0.000000,0.000000}%
\pgfsetstrokecolor{currentstroke}%
\pgfsetdash{}{0pt}%
\pgfpathmoveto{\pgfqpoint{3.562214in}{1.106250in}}%
\pgfpathlineto{\pgfqpoint{4.636364in}{1.106250in}}%
\pgfusepath{stroke}%
\end{pgfscope}%
\begin{pgfscope}%
\pgfpathrectangle{\pgfqpoint{0.150000in}{0.150000in}}{\pgfqpoint{4.700000in}{2.500000in}}%
\pgfusepath{clip}%
\pgfsetrectcap%
\pgfsetroundjoin%
\pgfsetlinewidth{2.007500pt}%
\definecolor{currentstroke}{rgb}{0.000000,0.000000,0.000000}%
\pgfsetstrokecolor{currentstroke}%
\pgfsetdash{}{0pt}%
\pgfpathmoveto{\pgfqpoint{0.572499in}{0.443750in}}%
\pgfpathlineto{\pgfqpoint{1.646648in}{0.443750in}}%
\pgfusepath{stroke}%
\end{pgfscope}%
\begin{pgfscope}%
\pgfpathrectangle{\pgfqpoint{0.150000in}{0.150000in}}{\pgfqpoint{4.700000in}{2.500000in}}%
\pgfusepath{clip}%
\pgfsetrectcap%
\pgfsetroundjoin%
\pgfsetlinewidth{2.007500pt}%
\definecolor{currentstroke}{rgb}{0.000000,0.000000,0.000000}%
\pgfsetstrokecolor{currentstroke}%
\pgfsetdash{}{0pt}%
\pgfpathmoveto{\pgfqpoint{1.885348in}{0.493750in}}%
\pgfpathlineto{\pgfqpoint{2.959497in}{0.493750in}}%
\pgfusepath{stroke}%
\end{pgfscope}%
\begin{pgfscope}%
\pgfpathrectangle{\pgfqpoint{0.150000in}{0.150000in}}{\pgfqpoint{4.700000in}{2.500000in}}%
\pgfusepath{clip}%
\pgfsetrectcap%
\pgfsetroundjoin%
\pgfsetlinewidth{2.007500pt}%
\definecolor{currentstroke}{rgb}{0.000000,0.000000,0.000000}%
\pgfsetstrokecolor{currentstroke}%
\pgfsetdash{}{0pt}%
\pgfpathmoveto{\pgfqpoint{3.269807in}{0.662500in}}%
\pgfpathlineto{\pgfqpoint{4.343956in}{0.662500in}}%
\pgfusepath{stroke}%
\end{pgfscope}%
\begin{pgfscope}%
\pgfpathrectangle{\pgfqpoint{0.150000in}{0.150000in}}{\pgfqpoint{4.700000in}{2.500000in}}%
\pgfusepath{clip}%
\pgfsetbuttcap%
\pgfsetroundjoin%
\pgfsetlinewidth{1.505625pt}%
\definecolor{currentstroke}{rgb}{0.000000,0.501961,0.000000}%
\pgfsetstrokecolor{currentstroke}%
\pgfsetdash{{5.550000pt}{2.400000pt}}{0.000000pt}%
\pgfpathmoveto{\pgfqpoint{4.302184in}{2.400000in}}%
\pgfpathlineto{\pgfqpoint{3.705434in}{2.337500in}}%
\pgfusepath{stroke}%
\end{pgfscope}%
\begin{pgfscope}%
\pgfpathrectangle{\pgfqpoint{0.150000in}{0.150000in}}{\pgfqpoint{4.700000in}{2.500000in}}%
\pgfusepath{clip}%
\pgfsetbuttcap%
\pgfsetroundjoin%
\definecolor{currentfill}{rgb}{0.274510,0.509804,0.705882}%
\pgfsetfillcolor{currentfill}%
\pgfsetlinewidth{1.505625pt}%
\definecolor{currentstroke}{rgb}{0.274510,0.509804,0.705882}%
\pgfsetstrokecolor{currentstroke}%
\pgfsetdash{}{0pt}%
\pgfsys@defobject{currentmarker}{\pgfqpoint{-0.069444in}{-0.069444in}}{\pgfqpoint{0.069444in}{0.069444in}}{%
\pgfpathmoveto{\pgfqpoint{-0.069444in}{-0.069444in}}%
\pgfpathlineto{\pgfqpoint{0.069444in}{0.069444in}}%
\pgfpathmoveto{\pgfqpoint{-0.069444in}{0.069444in}}%
\pgfpathlineto{\pgfqpoint{0.069444in}{-0.069444in}}%
\pgfusepath{stroke,fill}%
}%
\begin{pgfscope}%
\pgfsys@transformshift{4.302184in}{2.400000in}%
\pgfsys@useobject{currentmarker}{}%
\end{pgfscope}%
\end{pgfscope}%
\begin{pgfscope}%
\pgfpathrectangle{\pgfqpoint{0.150000in}{0.150000in}}{\pgfqpoint{4.700000in}{2.500000in}}%
\pgfusepath{clip}%
\pgfsetbuttcap%
\pgfsetroundjoin%
\definecolor{currentfill}{rgb}{0.980392,0.501961,0.447059}%
\pgfsetfillcolor{currentfill}%
\pgfsetlinewidth{1.505625pt}%
\definecolor{currentstroke}{rgb}{0.980392,0.501961,0.447059}%
\pgfsetstrokecolor{currentstroke}%
\pgfsetdash{}{0pt}%
\pgfsys@defobject{currentmarker}{\pgfqpoint{-0.069444in}{-0.069444in}}{\pgfqpoint{0.069444in}{0.069444in}}{%
\pgfpathmoveto{\pgfqpoint{-0.069444in}{-0.069444in}}%
\pgfpathlineto{\pgfqpoint{0.069444in}{0.069444in}}%
\pgfpathmoveto{\pgfqpoint{-0.069444in}{0.069444in}}%
\pgfpathlineto{\pgfqpoint{0.069444in}{-0.069444in}}%
\pgfusepath{stroke,fill}%
}%
\begin{pgfscope}%
\pgfsys@transformshift{3.705434in}{2.337500in}%
\pgfsys@useobject{currentmarker}{}%
\end{pgfscope}%
\end{pgfscope}%
\begin{pgfscope}%
\pgfpathrectangle{\pgfqpoint{0.150000in}{0.150000in}}{\pgfqpoint{4.700000in}{2.500000in}}%
\pgfusepath{clip}%
\pgfsetbuttcap%
\pgfsetroundjoin%
\definecolor{currentfill}{rgb}{0.274510,0.509804,0.705882}%
\pgfsetfillcolor{currentfill}%
\pgfsetlinewidth{1.505625pt}%
\definecolor{currentstroke}{rgb}{0.274510,0.509804,0.705882}%
\pgfsetstrokecolor{currentstroke}%
\pgfsetdash{}{0pt}%
\pgfsys@defobject{currentmarker}{\pgfqpoint{-0.069444in}{-0.069444in}}{\pgfqpoint{0.069444in}{0.069444in}}{%
\pgfpathmoveto{\pgfqpoint{-0.069444in}{-0.069444in}}%
\pgfpathlineto{\pgfqpoint{0.069444in}{0.069444in}}%
\pgfpathmoveto{\pgfqpoint{-0.069444in}{0.069444in}}%
\pgfpathlineto{\pgfqpoint{0.069444in}{-0.069444in}}%
\pgfusepath{stroke,fill}%
}%
\begin{pgfscope}%
\pgfsys@transformshift{0.453149in}{1.650000in}%
\pgfsys@useobject{currentmarker}{}%
\end{pgfscope}%
\end{pgfscope}%
\begin{pgfscope}%
\pgfpathrectangle{\pgfqpoint{0.150000in}{0.150000in}}{\pgfqpoint{4.700000in}{2.500000in}}%
\pgfusepath{clip}%
\pgfsetbuttcap%
\pgfsetroundjoin%
\definecolor{currentfill}{rgb}{0.980392,0.501961,0.447059}%
\pgfsetfillcolor{currentfill}%
\pgfsetlinewidth{1.505625pt}%
\definecolor{currentstroke}{rgb}{0.980392,0.501961,0.447059}%
\pgfsetstrokecolor{currentstroke}%
\pgfsetdash{}{0pt}%
\pgfsys@defobject{currentmarker}{\pgfqpoint{-0.069444in}{-0.069444in}}{\pgfqpoint{0.069444in}{0.069444in}}{%
\pgfpathmoveto{\pgfqpoint{-0.069444in}{-0.069444in}}%
\pgfpathlineto{\pgfqpoint{0.069444in}{0.069444in}}%
\pgfpathmoveto{\pgfqpoint{-0.069444in}{0.069444in}}%
\pgfpathlineto{\pgfqpoint{0.069444in}{-0.069444in}}%
\pgfusepath{stroke,fill}%
}%
\begin{pgfscope}%
\pgfsys@transformshift{1.527298in}{1.650000in}%
\pgfsys@useobject{currentmarker}{}%
\end{pgfscope}%
\end{pgfscope}%
\begin{pgfscope}%
\pgfpathrectangle{\pgfqpoint{0.150000in}{0.150000in}}{\pgfqpoint{4.700000in}{2.500000in}}%
\pgfusepath{clip}%
\pgfsetbuttcap%
\pgfsetroundjoin%
\definecolor{currentfill}{rgb}{0.274510,0.509804,0.705882}%
\pgfsetfillcolor{currentfill}%
\pgfsetlinewidth{1.505625pt}%
\definecolor{currentstroke}{rgb}{0.274510,0.509804,0.705882}%
\pgfsetstrokecolor{currentstroke}%
\pgfsetdash{}{0pt}%
\pgfsys@defobject{currentmarker}{\pgfqpoint{-0.069444in}{-0.069444in}}{\pgfqpoint{0.069444in}{0.069444in}}{%
\pgfpathmoveto{\pgfqpoint{-0.069444in}{-0.069444in}}%
\pgfpathlineto{\pgfqpoint{0.069444in}{0.069444in}}%
\pgfpathmoveto{\pgfqpoint{-0.069444in}{0.069444in}}%
\pgfpathlineto{\pgfqpoint{0.069444in}{-0.069444in}}%
\pgfusepath{stroke,fill}%
}%
\begin{pgfscope}%
\pgfsys@transformshift{2.004698in}{1.531250in}%
\pgfsys@useobject{currentmarker}{}%
\end{pgfscope}%
\end{pgfscope}%
\begin{pgfscope}%
\pgfpathrectangle{\pgfqpoint{0.150000in}{0.150000in}}{\pgfqpoint{4.700000in}{2.500000in}}%
\pgfusepath{clip}%
\pgfsetbuttcap%
\pgfsetroundjoin%
\definecolor{currentfill}{rgb}{0.980392,0.501961,0.447059}%
\pgfsetfillcolor{currentfill}%
\pgfsetlinewidth{1.505625pt}%
\definecolor{currentstroke}{rgb}{0.980392,0.501961,0.447059}%
\pgfsetstrokecolor{currentstroke}%
\pgfsetdash{}{0pt}%
\pgfsys@defobject{currentmarker}{\pgfqpoint{-0.069444in}{-0.069444in}}{\pgfqpoint{0.069444in}{0.069444in}}{%
\pgfpathmoveto{\pgfqpoint{-0.069444in}{-0.069444in}}%
\pgfpathlineto{\pgfqpoint{0.069444in}{0.069444in}}%
\pgfpathmoveto{\pgfqpoint{-0.069444in}{0.069444in}}%
\pgfpathlineto{\pgfqpoint{0.069444in}{-0.069444in}}%
\pgfusepath{stroke,fill}%
}%
\begin{pgfscope}%
\pgfsys@transformshift{3.078847in}{1.531250in}%
\pgfsys@useobject{currentmarker}{}%
\end{pgfscope}%
\end{pgfscope}%
\begin{pgfscope}%
\pgfpathrectangle{\pgfqpoint{0.150000in}{0.150000in}}{\pgfqpoint{4.700000in}{2.500000in}}%
\pgfusepath{clip}%
\pgfsetbuttcap%
\pgfsetroundjoin%
\definecolor{currentfill}{rgb}{0.274510,0.509804,0.705882}%
\pgfsetfillcolor{currentfill}%
\pgfsetlinewidth{1.505625pt}%
\definecolor{currentstroke}{rgb}{0.274510,0.509804,0.705882}%
\pgfsetstrokecolor{currentstroke}%
\pgfsetdash{}{0pt}%
\pgfsys@defobject{currentmarker}{\pgfqpoint{-0.069444in}{-0.069444in}}{\pgfqpoint{0.069444in}{0.069444in}}{%
\pgfpathmoveto{\pgfqpoint{-0.069444in}{-0.069444in}}%
\pgfpathlineto{\pgfqpoint{0.069444in}{0.069444in}}%
\pgfpathmoveto{\pgfqpoint{-0.069444in}{0.069444in}}%
\pgfpathlineto{\pgfqpoint{0.069444in}{-0.069444in}}%
\pgfusepath{stroke,fill}%
}%
\begin{pgfscope}%
\pgfsys@transformshift{3.245937in}{1.350000in}%
\pgfsys@useobject{currentmarker}{}%
\end{pgfscope}%
\end{pgfscope}%
\begin{pgfscope}%
\pgfpathrectangle{\pgfqpoint{0.150000in}{0.150000in}}{\pgfqpoint{4.700000in}{2.500000in}}%
\pgfusepath{clip}%
\pgfsetbuttcap%
\pgfsetroundjoin%
\definecolor{currentfill}{rgb}{0.980392,0.501961,0.447059}%
\pgfsetfillcolor{currentfill}%
\pgfsetlinewidth{1.505625pt}%
\definecolor{currentstroke}{rgb}{0.980392,0.501961,0.447059}%
\pgfsetstrokecolor{currentstroke}%
\pgfsetdash{}{0pt}%
\pgfsys@defobject{currentmarker}{\pgfqpoint{-0.069444in}{-0.069444in}}{\pgfqpoint{0.069444in}{0.069444in}}{%
\pgfpathmoveto{\pgfqpoint{-0.069444in}{-0.069444in}}%
\pgfpathlineto{\pgfqpoint{0.069444in}{0.069444in}}%
\pgfpathmoveto{\pgfqpoint{-0.069444in}{0.069444in}}%
\pgfpathlineto{\pgfqpoint{0.069444in}{-0.069444in}}%
\pgfusepath{stroke,fill}%
}%
\begin{pgfscope}%
\pgfsys@transformshift{4.320086in}{1.350000in}%
\pgfsys@useobject{currentmarker}{}%
\end{pgfscope}%
\end{pgfscope}%
\begin{pgfscope}%
\pgfpathrectangle{\pgfqpoint{0.150000in}{0.150000in}}{\pgfqpoint{4.700000in}{2.500000in}}%
\pgfusepath{clip}%
\pgfsetbuttcap%
\pgfsetroundjoin%
\definecolor{currentfill}{rgb}{0.274510,0.509804,0.705882}%
\pgfsetfillcolor{currentfill}%
\pgfsetlinewidth{1.505625pt}%
\definecolor{currentstroke}{rgb}{0.274510,0.509804,0.705882}%
\pgfsetstrokecolor{currentstroke}%
\pgfsetdash{}{0pt}%
\pgfsys@defobject{currentmarker}{\pgfqpoint{-0.069444in}{-0.069444in}}{\pgfqpoint{0.069444in}{0.069444in}}{%
\pgfpathmoveto{\pgfqpoint{-0.069444in}{-0.069444in}}%
\pgfpathlineto{\pgfqpoint{0.069444in}{0.069444in}}%
\pgfpathmoveto{\pgfqpoint{-0.069444in}{0.069444in}}%
\pgfpathlineto{\pgfqpoint{0.069444in}{-0.069444in}}%
\pgfusepath{stroke,fill}%
}%
\begin{pgfscope}%
\pgfsys@transformshift{0.363636in}{1.012500in}%
\pgfsys@useobject{currentmarker}{}%
\end{pgfscope}%
\end{pgfscope}%
\begin{pgfscope}%
\pgfpathrectangle{\pgfqpoint{0.150000in}{0.150000in}}{\pgfqpoint{4.700000in}{2.500000in}}%
\pgfusepath{clip}%
\pgfsetbuttcap%
\pgfsetroundjoin%
\definecolor{currentfill}{rgb}{0.980392,0.501961,0.447059}%
\pgfsetfillcolor{currentfill}%
\pgfsetlinewidth{1.505625pt}%
\definecolor{currentstroke}{rgb}{0.980392,0.501961,0.447059}%
\pgfsetstrokecolor{currentstroke}%
\pgfsetdash{}{0pt}%
\pgfsys@defobject{currentmarker}{\pgfqpoint{-0.069444in}{-0.069444in}}{\pgfqpoint{0.069444in}{0.069444in}}{%
\pgfpathmoveto{\pgfqpoint{-0.069444in}{-0.069444in}}%
\pgfpathlineto{\pgfqpoint{0.069444in}{0.069444in}}%
\pgfpathmoveto{\pgfqpoint{-0.069444in}{0.069444in}}%
\pgfpathlineto{\pgfqpoint{0.069444in}{-0.069444in}}%
\pgfusepath{stroke,fill}%
}%
\begin{pgfscope}%
\pgfsys@transformshift{1.437786in}{1.012500in}%
\pgfsys@useobject{currentmarker}{}%
\end{pgfscope}%
\end{pgfscope}%
\begin{pgfscope}%
\pgfpathrectangle{\pgfqpoint{0.150000in}{0.150000in}}{\pgfqpoint{4.700000in}{2.500000in}}%
\pgfusepath{clip}%
\pgfsetbuttcap%
\pgfsetroundjoin%
\definecolor{currentfill}{rgb}{0.274510,0.509804,0.705882}%
\pgfsetfillcolor{currentfill}%
\pgfsetlinewidth{1.505625pt}%
\definecolor{currentstroke}{rgb}{0.274510,0.509804,0.705882}%
\pgfsetstrokecolor{currentstroke}%
\pgfsetdash{}{0pt}%
\pgfsys@defobject{currentmarker}{\pgfqpoint{-0.069444in}{-0.069444in}}{\pgfqpoint{0.069444in}{0.069444in}}{%
\pgfpathmoveto{\pgfqpoint{-0.069444in}{-0.069444in}}%
\pgfpathlineto{\pgfqpoint{0.069444in}{0.069444in}}%
\pgfpathmoveto{\pgfqpoint{-0.069444in}{0.069444in}}%
\pgfpathlineto{\pgfqpoint{0.069444in}{-0.069444in}}%
\pgfusepath{stroke,fill}%
}%
\begin{pgfscope}%
\pgfsys@transformshift{2.016633in}{1.181250in}%
\pgfsys@useobject{currentmarker}{}%
\end{pgfscope}%
\end{pgfscope}%
\begin{pgfscope}%
\pgfpathrectangle{\pgfqpoint{0.150000in}{0.150000in}}{\pgfqpoint{4.700000in}{2.500000in}}%
\pgfusepath{clip}%
\pgfsetbuttcap%
\pgfsetroundjoin%
\definecolor{currentfill}{rgb}{0.980392,0.501961,0.447059}%
\pgfsetfillcolor{currentfill}%
\pgfsetlinewidth{1.505625pt}%
\definecolor{currentstroke}{rgb}{0.980392,0.501961,0.447059}%
\pgfsetstrokecolor{currentstroke}%
\pgfsetdash{}{0pt}%
\pgfsys@defobject{currentmarker}{\pgfqpoint{-0.069444in}{-0.069444in}}{\pgfqpoint{0.069444in}{0.069444in}}{%
\pgfpathmoveto{\pgfqpoint{-0.069444in}{-0.069444in}}%
\pgfpathlineto{\pgfqpoint{0.069444in}{0.069444in}}%
\pgfpathmoveto{\pgfqpoint{-0.069444in}{0.069444in}}%
\pgfpathlineto{\pgfqpoint{0.069444in}{-0.069444in}}%
\pgfusepath{stroke,fill}%
}%
\begin{pgfscope}%
\pgfsys@transformshift{3.090782in}{1.181250in}%
\pgfsys@useobject{currentmarker}{}%
\end{pgfscope}%
\end{pgfscope}%
\begin{pgfscope}%
\pgfpathrectangle{\pgfqpoint{0.150000in}{0.150000in}}{\pgfqpoint{4.700000in}{2.500000in}}%
\pgfusepath{clip}%
\pgfsetbuttcap%
\pgfsetroundjoin%
\definecolor{currentfill}{rgb}{0.274510,0.509804,0.705882}%
\pgfsetfillcolor{currentfill}%
\pgfsetlinewidth{1.505625pt}%
\definecolor{currentstroke}{rgb}{0.274510,0.509804,0.705882}%
\pgfsetstrokecolor{currentstroke}%
\pgfsetdash{}{0pt}%
\pgfsys@defobject{currentmarker}{\pgfqpoint{-0.069444in}{-0.069444in}}{\pgfqpoint{0.069444in}{0.069444in}}{%
\pgfpathmoveto{\pgfqpoint{-0.069444in}{-0.069444in}}%
\pgfpathlineto{\pgfqpoint{0.069444in}{0.069444in}}%
\pgfpathmoveto{\pgfqpoint{-0.069444in}{0.069444in}}%
\pgfpathlineto{\pgfqpoint{0.069444in}{-0.069444in}}%
\pgfusepath{stroke,fill}%
}%
\begin{pgfscope}%
\pgfsys@transformshift{3.562214in}{1.106250in}%
\pgfsys@useobject{currentmarker}{}%
\end{pgfscope}%
\end{pgfscope}%
\begin{pgfscope}%
\pgfpathrectangle{\pgfqpoint{0.150000in}{0.150000in}}{\pgfqpoint{4.700000in}{2.500000in}}%
\pgfusepath{clip}%
\pgfsetbuttcap%
\pgfsetroundjoin%
\definecolor{currentfill}{rgb}{0.980392,0.501961,0.447059}%
\pgfsetfillcolor{currentfill}%
\pgfsetlinewidth{1.505625pt}%
\definecolor{currentstroke}{rgb}{0.980392,0.501961,0.447059}%
\pgfsetstrokecolor{currentstroke}%
\pgfsetdash{}{0pt}%
\pgfsys@defobject{currentmarker}{\pgfqpoint{-0.069444in}{-0.069444in}}{\pgfqpoint{0.069444in}{0.069444in}}{%
\pgfpathmoveto{\pgfqpoint{-0.069444in}{-0.069444in}}%
\pgfpathlineto{\pgfqpoint{0.069444in}{0.069444in}}%
\pgfpathmoveto{\pgfqpoint{-0.069444in}{0.069444in}}%
\pgfpathlineto{\pgfqpoint{0.069444in}{-0.069444in}}%
\pgfusepath{stroke,fill}%
}%
\begin{pgfscope}%
\pgfsys@transformshift{4.636364in}{1.106250in}%
\pgfsys@useobject{currentmarker}{}%
\end{pgfscope}%
\end{pgfscope}%
\begin{pgfscope}%
\pgfpathrectangle{\pgfqpoint{0.150000in}{0.150000in}}{\pgfqpoint{4.700000in}{2.500000in}}%
\pgfusepath{clip}%
\pgfsetbuttcap%
\pgfsetroundjoin%
\definecolor{currentfill}{rgb}{0.274510,0.509804,0.705882}%
\pgfsetfillcolor{currentfill}%
\pgfsetlinewidth{1.505625pt}%
\definecolor{currentstroke}{rgb}{0.274510,0.509804,0.705882}%
\pgfsetstrokecolor{currentstroke}%
\pgfsetdash{}{0pt}%
\pgfsys@defobject{currentmarker}{\pgfqpoint{-0.069444in}{-0.069444in}}{\pgfqpoint{0.069444in}{0.069444in}}{%
\pgfpathmoveto{\pgfqpoint{-0.069444in}{-0.069444in}}%
\pgfpathlineto{\pgfqpoint{0.069444in}{0.069444in}}%
\pgfpathmoveto{\pgfqpoint{-0.069444in}{0.069444in}}%
\pgfpathlineto{\pgfqpoint{0.069444in}{-0.069444in}}%
\pgfusepath{stroke,fill}%
}%
\begin{pgfscope}%
\pgfsys@transformshift{0.572499in}{0.443750in}%
\pgfsys@useobject{currentmarker}{}%
\end{pgfscope}%
\end{pgfscope}%
\begin{pgfscope}%
\pgfpathrectangle{\pgfqpoint{0.150000in}{0.150000in}}{\pgfqpoint{4.700000in}{2.500000in}}%
\pgfusepath{clip}%
\pgfsetbuttcap%
\pgfsetroundjoin%
\definecolor{currentfill}{rgb}{0.980392,0.501961,0.447059}%
\pgfsetfillcolor{currentfill}%
\pgfsetlinewidth{1.505625pt}%
\definecolor{currentstroke}{rgb}{0.980392,0.501961,0.447059}%
\pgfsetstrokecolor{currentstroke}%
\pgfsetdash{}{0pt}%
\pgfsys@defobject{currentmarker}{\pgfqpoint{-0.069444in}{-0.069444in}}{\pgfqpoint{0.069444in}{0.069444in}}{%
\pgfpathmoveto{\pgfqpoint{-0.069444in}{-0.069444in}}%
\pgfpathlineto{\pgfqpoint{0.069444in}{0.069444in}}%
\pgfpathmoveto{\pgfqpoint{-0.069444in}{0.069444in}}%
\pgfpathlineto{\pgfqpoint{0.069444in}{-0.069444in}}%
\pgfusepath{stroke,fill}%
}%
\begin{pgfscope}%
\pgfsys@transformshift{1.646648in}{0.443750in}%
\pgfsys@useobject{currentmarker}{}%
\end{pgfscope}%
\end{pgfscope}%
\begin{pgfscope}%
\pgfpathrectangle{\pgfqpoint{0.150000in}{0.150000in}}{\pgfqpoint{4.700000in}{2.500000in}}%
\pgfusepath{clip}%
\pgfsetbuttcap%
\pgfsetroundjoin%
\definecolor{currentfill}{rgb}{0.274510,0.509804,0.705882}%
\pgfsetfillcolor{currentfill}%
\pgfsetlinewidth{1.505625pt}%
\definecolor{currentstroke}{rgb}{0.274510,0.509804,0.705882}%
\pgfsetstrokecolor{currentstroke}%
\pgfsetdash{}{0pt}%
\pgfsys@defobject{currentmarker}{\pgfqpoint{-0.069444in}{-0.069444in}}{\pgfqpoint{0.069444in}{0.069444in}}{%
\pgfpathmoveto{\pgfqpoint{-0.069444in}{-0.069444in}}%
\pgfpathlineto{\pgfqpoint{0.069444in}{0.069444in}}%
\pgfpathmoveto{\pgfqpoint{-0.069444in}{0.069444in}}%
\pgfpathlineto{\pgfqpoint{0.069444in}{-0.069444in}}%
\pgfusepath{stroke,fill}%
}%
\begin{pgfscope}%
\pgfsys@transformshift{1.885348in}{0.493750in}%
\pgfsys@useobject{currentmarker}{}%
\end{pgfscope}%
\end{pgfscope}%
\begin{pgfscope}%
\pgfpathrectangle{\pgfqpoint{0.150000in}{0.150000in}}{\pgfqpoint{4.700000in}{2.500000in}}%
\pgfusepath{clip}%
\pgfsetbuttcap%
\pgfsetroundjoin%
\definecolor{currentfill}{rgb}{0.980392,0.501961,0.447059}%
\pgfsetfillcolor{currentfill}%
\pgfsetlinewidth{1.505625pt}%
\definecolor{currentstroke}{rgb}{0.980392,0.501961,0.447059}%
\pgfsetstrokecolor{currentstroke}%
\pgfsetdash{}{0pt}%
\pgfsys@defobject{currentmarker}{\pgfqpoint{-0.069444in}{-0.069444in}}{\pgfqpoint{0.069444in}{0.069444in}}{%
\pgfpathmoveto{\pgfqpoint{-0.069444in}{-0.069444in}}%
\pgfpathlineto{\pgfqpoint{0.069444in}{0.069444in}}%
\pgfpathmoveto{\pgfqpoint{-0.069444in}{0.069444in}}%
\pgfpathlineto{\pgfqpoint{0.069444in}{-0.069444in}}%
\pgfusepath{stroke,fill}%
}%
\begin{pgfscope}%
\pgfsys@transformshift{2.959497in}{0.493750in}%
\pgfsys@useobject{currentmarker}{}%
\end{pgfscope}%
\end{pgfscope}%
\begin{pgfscope}%
\pgfpathrectangle{\pgfqpoint{0.150000in}{0.150000in}}{\pgfqpoint{4.700000in}{2.500000in}}%
\pgfusepath{clip}%
\pgfsetbuttcap%
\pgfsetroundjoin%
\definecolor{currentfill}{rgb}{0.274510,0.509804,0.705882}%
\pgfsetfillcolor{currentfill}%
\pgfsetlinewidth{1.505625pt}%
\definecolor{currentstroke}{rgb}{0.274510,0.509804,0.705882}%
\pgfsetstrokecolor{currentstroke}%
\pgfsetdash{}{0pt}%
\pgfsys@defobject{currentmarker}{\pgfqpoint{-0.069444in}{-0.069444in}}{\pgfqpoint{0.069444in}{0.069444in}}{%
\pgfpathmoveto{\pgfqpoint{-0.069444in}{-0.069444in}}%
\pgfpathlineto{\pgfqpoint{0.069444in}{0.069444in}}%
\pgfpathmoveto{\pgfqpoint{-0.069444in}{0.069444in}}%
\pgfpathlineto{\pgfqpoint{0.069444in}{-0.069444in}}%
\pgfusepath{stroke,fill}%
}%
\begin{pgfscope}%
\pgfsys@transformshift{3.269807in}{0.662500in}%
\pgfsys@useobject{currentmarker}{}%
\end{pgfscope}%
\end{pgfscope}%
\begin{pgfscope}%
\pgfpathrectangle{\pgfqpoint{0.150000in}{0.150000in}}{\pgfqpoint{4.700000in}{2.500000in}}%
\pgfusepath{clip}%
\pgfsetbuttcap%
\pgfsetroundjoin%
\definecolor{currentfill}{rgb}{0.980392,0.501961,0.447059}%
\pgfsetfillcolor{currentfill}%
\pgfsetlinewidth{1.505625pt}%
\definecolor{currentstroke}{rgb}{0.980392,0.501961,0.447059}%
\pgfsetstrokecolor{currentstroke}%
\pgfsetdash{}{0pt}%
\pgfsys@defobject{currentmarker}{\pgfqpoint{-0.069444in}{-0.069444in}}{\pgfqpoint{0.069444in}{0.069444in}}{%
\pgfpathmoveto{\pgfqpoint{-0.069444in}{-0.069444in}}%
\pgfpathlineto{\pgfqpoint{0.069444in}{0.069444in}}%
\pgfpathmoveto{\pgfqpoint{-0.069444in}{0.069444in}}%
\pgfpathlineto{\pgfqpoint{0.069444in}{-0.069444in}}%
\pgfusepath{stroke,fill}%
}%
\begin{pgfscope}%
\pgfsys@transformshift{4.343956in}{0.662500in}%
\pgfsys@useobject{currentmarker}{}%
\end{pgfscope}%
\end{pgfscope}%
\begin{pgfscope}%
\pgfpathrectangle{\pgfqpoint{0.150000in}{0.150000in}}{\pgfqpoint{4.700000in}{2.500000in}}%
\pgfusepath{clip}%
\pgfsetbuttcap%
\pgfsetroundjoin%
\pgfsetlinewidth{1.505625pt}%
\definecolor{currentstroke}{rgb}{0.000000,0.501961,0.000000}%
\pgfsetstrokecolor{currentstroke}%
\pgfsetdash{{5.550000pt}{2.400000pt}}{0.000000pt}%
\pgfpathmoveto{\pgfqpoint{2.004698in}{1.531250in}}%
\pgfpathlineto{\pgfqpoint{1.527298in}{1.650000in}}%
\pgfusepath{stroke}%
\end{pgfscope}%
\begin{pgfscope}%
\pgfpathrectangle{\pgfqpoint{0.150000in}{0.150000in}}{\pgfqpoint{4.700000in}{2.500000in}}%
\pgfusepath{clip}%
\pgfsetbuttcap%
\pgfsetroundjoin%
\pgfsetlinewidth{1.505625pt}%
\definecolor{currentstroke}{rgb}{0.000000,0.501961,0.000000}%
\pgfsetstrokecolor{currentstroke}%
\pgfsetdash{{5.550000pt}{2.400000pt}}{0.000000pt}%
\pgfpathmoveto{\pgfqpoint{3.245937in}{1.350000in}}%
\pgfpathlineto{\pgfqpoint{3.078847in}{1.531250in}}%
\pgfusepath{stroke}%
\end{pgfscope}%
\begin{pgfscope}%
\pgfpathrectangle{\pgfqpoint{0.150000in}{0.150000in}}{\pgfqpoint{4.700000in}{2.500000in}}%
\pgfusepath{clip}%
\pgfsetbuttcap%
\pgfsetroundjoin%
\pgfsetlinewidth{1.505625pt}%
\definecolor{currentstroke}{rgb}{0.000000,0.501961,0.000000}%
\pgfsetstrokecolor{currentstroke}%
\pgfsetdash{{5.550000pt}{2.400000pt}}{0.000000pt}%
\pgfpathmoveto{\pgfqpoint{2.016633in}{1.181250in}}%
\pgfpathlineto{\pgfqpoint{1.437786in}{1.012500in}}%
\pgfusepath{stroke}%
\end{pgfscope}%
\begin{pgfscope}%
\pgfpathrectangle{\pgfqpoint{0.150000in}{0.150000in}}{\pgfqpoint{4.700000in}{2.500000in}}%
\pgfusepath{clip}%
\pgfsetbuttcap%
\pgfsetroundjoin%
\pgfsetlinewidth{1.505625pt}%
\definecolor{currentstroke}{rgb}{0.000000,0.501961,0.000000}%
\pgfsetstrokecolor{currentstroke}%
\pgfsetdash{{5.550000pt}{2.400000pt}}{0.000000pt}%
\pgfpathmoveto{\pgfqpoint{3.562214in}{1.106250in}}%
\pgfpathlineto{\pgfqpoint{3.090782in}{1.181250in}}%
\pgfusepath{stroke}%
\end{pgfscope}%
\begin{pgfscope}%
\pgfpathrectangle{\pgfqpoint{0.150000in}{0.150000in}}{\pgfqpoint{4.700000in}{2.500000in}}%
\pgfusepath{clip}%
\pgfsetbuttcap%
\pgfsetroundjoin%
\pgfsetlinewidth{1.505625pt}%
\definecolor{currentstroke}{rgb}{0.000000,0.501961,0.000000}%
\pgfsetstrokecolor{currentstroke}%
\pgfsetdash{{5.550000pt}{2.400000pt}}{0.000000pt}%
\pgfpathmoveto{\pgfqpoint{1.885348in}{0.493750in}}%
\pgfpathlineto{\pgfqpoint{1.646648in}{0.443750in}}%
\pgfusepath{stroke}%
\end{pgfscope}%
\begin{pgfscope}%
\pgfpathrectangle{\pgfqpoint{0.150000in}{0.150000in}}{\pgfqpoint{4.700000in}{2.500000in}}%
\pgfusepath{clip}%
\pgfsetbuttcap%
\pgfsetroundjoin%
\pgfsetlinewidth{1.505625pt}%
\definecolor{currentstroke}{rgb}{0.000000,0.501961,0.000000}%
\pgfsetstrokecolor{currentstroke}%
\pgfsetdash{{5.550000pt}{2.400000pt}}{0.000000pt}%
\pgfpathmoveto{\pgfqpoint{3.269807in}{0.662500in}}%
\pgfpathlineto{\pgfqpoint{2.959497in}{0.493750in}}%
\pgfusepath{stroke}%
\end{pgfscope}%
\begin{pgfscope}%
\pgfsetrectcap%
\pgfsetmiterjoin%
\pgfsetlinewidth{0.803000pt}%
\definecolor{currentstroke}{rgb}{0.501961,0.501961,0.501961}%
\pgfsetstrokecolor{currentstroke}%
\pgfsetdash{}{0pt}%
\pgfpathmoveto{\pgfqpoint{0.150000in}{0.150000in}}%
\pgfpathlineto{\pgfqpoint{0.150000in}{2.650000in}}%
\pgfusepath{stroke}%
\end{pgfscope}%
\begin{pgfscope}%
\pgfsetrectcap%
\pgfsetmiterjoin%
\pgfsetlinewidth{0.803000pt}%
\definecolor{currentstroke}{rgb}{0.501961,0.501961,0.501961}%
\pgfsetstrokecolor{currentstroke}%
\pgfsetdash{}{0pt}%
\pgfpathmoveto{\pgfqpoint{4.850000in}{0.150000in}}%
\pgfpathlineto{\pgfqpoint{4.850000in}{2.650000in}}%
\pgfusepath{stroke}%
\end{pgfscope}%
\begin{pgfscope}%
\pgfsetrectcap%
\pgfsetmiterjoin%
\pgfsetlinewidth{0.803000pt}%
\definecolor{currentstroke}{rgb}{0.501961,0.501961,0.501961}%
\pgfsetstrokecolor{currentstroke}%
\pgfsetdash{}{0pt}%
\pgfpathmoveto{\pgfqpoint{0.150000in}{0.150000in}}%
\pgfpathlineto{\pgfqpoint{4.850000in}{0.150000in}}%
\pgfusepath{stroke}%
\end{pgfscope}%
\begin{pgfscope}%
\pgfsetrectcap%
\pgfsetmiterjoin%
\pgfsetlinewidth{0.803000pt}%
\definecolor{currentstroke}{rgb}{0.501961,0.501961,0.501961}%
\pgfsetstrokecolor{currentstroke}%
\pgfsetdash{}{0pt}%
\pgfpathmoveto{\pgfqpoint{0.150000in}{2.650000in}}%
\pgfpathlineto{\pgfqpoint{4.850000in}{2.650000in}}%
\pgfusepath{stroke}%
\end{pgfscope}%
\begin{pgfscope}%
\pgfpathrectangle{\pgfqpoint{0.150000in}{0.150000in}}{\pgfqpoint{4.700000in}{2.500000in}}%
\pgfusepath{clip}%
\pgfsetbuttcap%
\pgfsetroundjoin%
\definecolor{currentfill}{rgb}{0.274510,0.509804,0.705882}%
\pgfsetfillcolor{currentfill}%
\pgfsetlinewidth{1.505625pt}%
\definecolor{currentstroke}{rgb}{0.274510,0.509804,0.705882}%
\pgfsetstrokecolor{currentstroke}%
\pgfsetdash{}{0pt}%
\pgfsys@defobject{currentmarker}{\pgfqpoint{-0.069444in}{-0.069444in}}{\pgfqpoint{0.069444in}{0.069444in}}{%
\pgfpathmoveto{\pgfqpoint{-0.069444in}{-0.069444in}}%
\pgfpathlineto{\pgfqpoint{0.069444in}{0.069444in}}%
\pgfpathmoveto{\pgfqpoint{-0.069444in}{0.069444in}}%
\pgfpathlineto{\pgfqpoint{0.069444in}{-0.069444in}}%
\pgfusepath{stroke,fill}%
}%
\begin{pgfscope}%
\pgfsys@transformshift{0.453149in}{1.650000in}%
\pgfsys@useobject{currentmarker}{}%
\end{pgfscope}%
\begin{pgfscope}%
\pgfsys@transformshift{2.004698in}{1.531250in}%
\pgfsys@useobject{currentmarker}{}%
\end{pgfscope}%
\begin{pgfscope}%
\pgfsys@transformshift{3.245937in}{1.350000in}%
\pgfsys@useobject{currentmarker}{}%
\end{pgfscope}%
\begin{pgfscope}%
\pgfsys@transformshift{0.363636in}{1.012500in}%
\pgfsys@useobject{currentmarker}{}%
\end{pgfscope}%
\begin{pgfscope}%
\pgfsys@transformshift{2.016633in}{1.181250in}%
\pgfsys@useobject{currentmarker}{}%
\end{pgfscope}%
\begin{pgfscope}%
\pgfsys@transformshift{3.562214in}{1.106250in}%
\pgfsys@useobject{currentmarker}{}%
\end{pgfscope}%
\begin{pgfscope}%
\pgfsys@transformshift{0.572499in}{0.443750in}%
\pgfsys@useobject{currentmarker}{}%
\end{pgfscope}%
\begin{pgfscope}%
\pgfsys@transformshift{1.885348in}{0.493750in}%
\pgfsys@useobject{currentmarker}{}%
\end{pgfscope}%
\begin{pgfscope}%
\pgfsys@transformshift{3.269807in}{0.662500in}%
\pgfsys@useobject{currentmarker}{}%
\end{pgfscope}%
\end{pgfscope}%
\begin{pgfscope}%
\pgfpathrectangle{\pgfqpoint{0.150000in}{0.150000in}}{\pgfqpoint{4.700000in}{2.500000in}}%
\pgfusepath{clip}%
\pgfsetbuttcap%
\pgfsetroundjoin%
\definecolor{currentfill}{rgb}{0.980392,0.501961,0.447059}%
\pgfsetfillcolor{currentfill}%
\pgfsetlinewidth{1.505625pt}%
\definecolor{currentstroke}{rgb}{0.980392,0.501961,0.447059}%
\pgfsetstrokecolor{currentstroke}%
\pgfsetdash{}{0pt}%
\pgfsys@defobject{currentmarker}{\pgfqpoint{-0.069444in}{-0.069444in}}{\pgfqpoint{0.069444in}{0.069444in}}{%
\pgfpathmoveto{\pgfqpoint{-0.069444in}{-0.069444in}}%
\pgfpathlineto{\pgfqpoint{0.069444in}{0.069444in}}%
\pgfpathmoveto{\pgfqpoint{-0.069444in}{0.069444in}}%
\pgfpathlineto{\pgfqpoint{0.069444in}{-0.069444in}}%
\pgfusepath{stroke,fill}%
}%
\begin{pgfscope}%
\pgfsys@transformshift{1.527298in}{1.650000in}%
\pgfsys@useobject{currentmarker}{}%
\end{pgfscope}%
\begin{pgfscope}%
\pgfsys@transformshift{3.078847in}{1.531250in}%
\pgfsys@useobject{currentmarker}{}%
\end{pgfscope}%
\begin{pgfscope}%
\pgfsys@transformshift{4.320086in}{1.350000in}%
\pgfsys@useobject{currentmarker}{}%
\end{pgfscope}%
\begin{pgfscope}%
\pgfsys@transformshift{1.437786in}{1.012500in}%
\pgfsys@useobject{currentmarker}{}%
\end{pgfscope}%
\begin{pgfscope}%
\pgfsys@transformshift{3.090782in}{1.181250in}%
\pgfsys@useobject{currentmarker}{}%
\end{pgfscope}%
\begin{pgfscope}%
\pgfsys@transformshift{4.636364in}{1.106250in}%
\pgfsys@useobject{currentmarker}{}%
\end{pgfscope}%
\begin{pgfscope}%
\pgfsys@transformshift{1.646648in}{0.443750in}%
\pgfsys@useobject{currentmarker}{}%
\end{pgfscope}%
\begin{pgfscope}%
\pgfsys@transformshift{2.959497in}{0.493750in}%
\pgfsys@useobject{currentmarker}{}%
\end{pgfscope}%
\begin{pgfscope}%
\pgfsys@transformshift{4.343956in}{0.662500in}%
\pgfsys@useobject{currentmarker}{}%
\end{pgfscope}%
\end{pgfscope}%
\begin{pgfscope}%
\pgfsetbuttcap%
\pgfsetroundjoin%
\definecolor{currentfill}{rgb}{0.274510,0.509804,0.705882}%
\pgfsetfillcolor{currentfill}%
\pgfsetlinewidth{1.505625pt}%
\definecolor{currentstroke}{rgb}{0.274510,0.509804,0.705882}%
\pgfsetstrokecolor{currentstroke}%
\pgfsetdash{}{0pt}%
\pgfsys@defobject{currentmarker}{\pgfqpoint{-0.069444in}{-0.069444in}}{\pgfqpoint{0.069444in}{0.069444in}}{%
\pgfpathmoveto{\pgfqpoint{-0.069444in}{-0.069444in}}%
\pgfpathlineto{\pgfqpoint{0.069444in}{0.069444in}}%
\pgfpathmoveto{\pgfqpoint{-0.069444in}{0.069444in}}%
\pgfpathlineto{\pgfqpoint{0.069444in}{-0.069444in}}%
\pgfusepath{stroke,fill}%
}%
\begin{pgfscope}%
\pgfsys@transformshift{0.413889in}{2.455935in}%
\pgfsys@useobject{currentmarker}{}%
\end{pgfscope}%
\end{pgfscope}%
\begin{pgfscope}%
\definecolor{textcolor}{rgb}{0.000000,0.000000,0.000000}%
\pgfsetstrokecolor{textcolor}%
\pgfsetfillcolor{textcolor}%
\pgftext[x=0.663889in,y=2.419477in,left,base]{\color{textcolor}{\sffamily\fontsize{10.000000}{12.000000}\selectfont\catcode`\^=\active\def^{\ifmmode\sp\else\^{}\fi}\catcode`\%=\active\def
\end{pgfscope}%
\begin{pgfscope}%
\pgfsetbuttcap%
\pgfsetroundjoin%
\definecolor{currentfill}{rgb}{0.980392,0.501961,0.447059}%
\pgfsetfillcolor{currentfill}%
\pgfsetlinewidth{1.505625pt}%
\definecolor{currentstroke}{rgb}{0.980392,0.501961,0.447059}%
\pgfsetstrokecolor{currentstroke}%
\pgfsetdash{}{0pt}%
\pgfsys@defobject{currentmarker}{\pgfqpoint{-0.069444in}{-0.069444in}}{\pgfqpoint{0.069444in}{0.069444in}}{%
\pgfpathmoveto{\pgfqpoint{-0.069444in}{-0.069444in}}%
\pgfpathlineto{\pgfqpoint{0.069444in}{0.069444in}}%
\pgfpathmoveto{\pgfqpoint{-0.069444in}{0.069444in}}%
\pgfpathlineto{\pgfqpoint{0.069444in}{-0.069444in}}%
\pgfusepath{stroke,fill}%
}%
\begin{pgfscope}%
\pgfsys@transformshift{0.413889in}{2.252078in}%
\pgfsys@useobject{currentmarker}{}%
\end{pgfscope}%
\end{pgfscope}%
\begin{pgfscope}%
\definecolor{textcolor}{rgb}{0.000000,0.000000,0.000000}%
\pgfsetstrokecolor{textcolor}%
\pgfsetfillcolor{textcolor}%
\pgftext[x=0.663889in,y=2.215620in,left,base]{\color{textcolor}{\sffamily\fontsize{10.000000}{12.000000}\selectfont\catcode`\^=\active\def^{\ifmmode\sp\else\^{}\fi}\catcode`\%=\active\def
\end{pgfscope}%
\begin{pgfscope}%
\pgfsetbuttcap%
\pgfsetroundjoin%
\pgfsetlinewidth{1.505625pt}%
\definecolor{currentstroke}{rgb}{0.000000,0.501961,0.000000}%
\pgfsetstrokecolor{currentstroke}%
\pgfsetdash{{5.550000pt}{2.400000pt}}{0.000000pt}%
\pgfpathmoveto{\pgfqpoint{0.275000in}{2.060374in}}%
\pgfpathlineto{\pgfqpoint{0.413889in}{2.060374in}}%
\pgfpathlineto{\pgfqpoint{0.552778in}{2.060374in}}%
\pgfusepath{stroke}%
\end{pgfscope}%
\begin{pgfscope}%
\definecolor{textcolor}{rgb}{0.000000,0.000000,0.000000}%
\pgfsetstrokecolor{textcolor}%
\pgfsetfillcolor{textcolor}%
\pgftext[x=0.663889in,y=2.011762in,left,base]{\color{textcolor}{\sffamily\fontsize{10.000000}{12.000000}\selectfont\catcode`\^=\active\def^{\ifmmode\sp\else\^{}\fi}\catcode`\%=\active\def
\end{pgfscope}%
\begin{pgfscope}%
\pgfsetrectcap%
\pgfsetroundjoin%
\pgfsetlinewidth{2.007500pt}%
\definecolor{currentstroke}{rgb}{0.000000,0.000000,0.000000}%
\pgfsetstrokecolor{currentstroke}%
\pgfsetdash{}{0pt}%
\pgfpathmoveto{\pgfqpoint{0.275000in}{1.856516in}}%
\pgfpathlineto{\pgfqpoint{0.413889in}{1.856516in}}%
\pgfpathlineto{\pgfqpoint{0.552778in}{1.856516in}}%
\pgfusepath{stroke}%
\end{pgfscope}%
\begin{pgfscope}%
\definecolor{textcolor}{rgb}{0.000000,0.000000,0.000000}%
\pgfsetstrokecolor{textcolor}%
\pgfsetfillcolor{textcolor}%
\pgftext[x=0.663889in,y=1.807905in,left,base]{\color{textcolor}{\sffamily\fontsize{10.000000}{12.000000}\selectfont\catcode`\^=\active\def^{\ifmmode\sp\else\^{}\fi}\catcode`\%=\active\def
\end{pgfscope}%
\end{pgfpicture}%
\makeatother%
\endgroup%